%% file: main.tex
\documentclass{article}

\usepackage[english]{babel}
\usepackage{graphicx} 
\usepackage{defs,fonts}
\usepackage[margin=1in]{geometry} 
\usepackage{algpseudocode}
\usepackage{afterpage}
\usepackage{amsbsy}
\usepackage{amsmath}
\usepackage{amssymb}
\usepackage{amsthm}
\usepackage{bm}
\usepackage{bbm}
\usepackage[
backend=biber,
style=alphabetic,
sorting=ynt
]{biblatex}
\usepackage{caption}
\usepackage{cases}
\usepackage{color}
\usepackage{dcolumn}
\usepackage{dsfont }
\usepackage{fonts}
\usepackage{graphicx} 
\usepackage{microtype}
\usepackage{subcaption}
\usepackage{tabularx}
\usepackage{tikz} 
\usepackage{tikz-3dplot}
\usepackage{makecell}
\tikzset{>=latex}
\usetikzlibrary{calc}
\usepackage[colorlinks=true, allcolors=blue]{hyperref}
\usepackage{cleveref}
\usepackage{booktabs}      
\usepackage[ruled,vlined,linesnumbered,noend]{algorithm2e}
\usepackage{amsfonts}
\usepackage{comment}
\usepackage{tikz}
\usepackage{tikz-3dplot}
\usepackage{multirow} 

\usepackage{wrapfig}
\usepackage{booktabs}    
\usepackage{adjustbox}
\usepackage{graphicx}
\usepackage{array}
\usepackage{url} 
\usepackage{authblk}
\usepackage{amsmath}
\usepackage{graphicx}

\title{Efficient Neural Surrogates for Linear Radiation Transport on the Lattice and Hohlraum benchmarks \thanks{This manuscript has been authored, in part, by UT-Battelle, LLC under Contract No.~DE-AC05-00OR22725 with the U.S.~Department of Energy. The United States Government retains and the publisher, by accepting the article for publication, acknowledges that the United States Government retains a non-exclusive, paid-up, irrevocable, world-wide license to publish or reproduce the published form of this manuscript, or allow others to do so, for United States Government purposes. The Department of Energy will provide public access to these results of federally sponsored research in accordance with the DOE Public Access Plan(\url{http://energy.gov/downloads/doe-public-access-plan}).\\
S.~Schotth\"ofer and C.D. Hauck were supported by the Office of Science, Advanced Scientific Computing Research program of the U.S. Department of Energy.
}}

\author[1]{Carmelo Gonzales}
\author[2]{Steffen Schotth\"ofer}
\author[3]{Cory D. Hauck}

\affil[1]{NVIDIA, Santa Clara, US }
\affil[2]{Seamless Labs, Mountain View, US }
\affil[3]{
Oak Ridge National Laboratory}

\begin{document}
\input{macros_steffen}

\definecolor{ssblue}{rgb}{0.0,0.35,0.75}
\newcommand{\ssfix}[1]{{\color{ssblue}\textbf{[SS:]}~#1}}
\providecommand{\ssbold}[1]{\mathbf{#1}}
\providecommand{\bp}{\ssbold{p}}
\providecommand{\bff}{\ssbold{f}}
\providecommand{\bh}{\ssbold{h}}
\providecommand{\bv}{\ssbold{v}}
\providecommand{\bfe}{\ssbold{e}}
\providecommand{\bc}{\ssbold{c}}
\providecommand{\bzero}{\ssbold{0}}

\maketitle

\begin{abstract}
Linear radiation transport equations (RTEs) form the simulation foundations underpinning design and analysis tasks in nuclear engineering, inertial confinement fusion, medical imaging, and astrophysics, but resolving the high-dimensional phase space at engineering fidelity remains expensive enough that outer-loop workflows, such as design optimization, uncertainty quantification, and parameter sweeps, are routinely budget-bound on traditional solvers. Neural surrogates promise to relax this bottleneck by amortizing simulation cost across thousands of downstream queries, but the architectural choices and engineered inductive biases that make a surrogate accurate on one transport problem do not transfer straightforwardly across model families. We benchmark two parameter-matched neural surrogate architectures, the physics-attention Transolver \cite{wu2024transolver} and the multi-scale graph network Bi-Stride Multi-Scale MeshGraphNet (BSMS-MGN) \cite{cao2022bi}, as end-to-end approximations of the final-time particle concentration for the two-dimensional linear RTE on the canonical Lattice and Hohlraum benchmarks of \cite{reference_paper}. Both surrogates are trained on data generated with the KiT-RT solver \cite{kusch2023kit} using a discrete-ordinates / finite-volume solver and deliver inference speedups ranging from roughly $90\times$ to $4400\times$ over the high-resolution reference solver, depending on benchmark, architecture, and evaluation mesh. An ablation across Fourier features and region-weighted training loss exposes strongly architecture-dependent inductive-bias preferences, indicating that design choices common to physics-informed surrogate workflows must be revisited per architecture rather than imported across model families, and that downstream utility depends on per-QoI sensitivity rather than a single field-level score. The model training recipe, training data, and evaluation pipeline are released alongside this paper to support reproduction, transfer to related transport problems, and evaluation as amortized forward-model components in larger outer-loop simulation workflows.  
\end{abstract}

\input{sections/intro}
\input{sections/rte}

\section{Benchmark problems}\label{sec_problems}

In this section, we restate the two benchmark problems.  For each benchmark, we provide (i) a detailed mathematical description of the domain and problem inputs, (ii) a list of the quantities of interest to be computed, and (iii) a list of design parameters.  The presentation here follows \cite{reference_paper} closely.

\subsection{Lattice benchmark}\label{sec_lattice}
\begin{figure}

\begin{subfigure}[t!]{0.32\linewidth}
    \centering
    \includegraphics[width=0.9\textwidth,trim={0 0 6cm 0},clip]{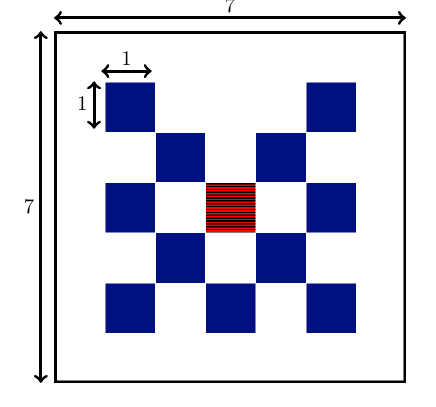}
    \vfill
    \caption{Geometric layout}
      \label{fig_lattice_layout}
\end{subfigure}
    \begin{subfigure}[t!]{0.32\linewidth}
     \vfill
    \centering
      \resizebox{\textwidth}{!}{%
\begin{tabular}{l | c ccc}
subdomain &  $\sigmat$ &  $\sigmas$ &$\sigmaa$ & $Q$ \\
\midrule 
white & $\sigmas+\sigmaa$ & $[0.1,10.1]$ & $0$ & $0$\\
blue & $\sigmas+\sigmaa$ & $0$ & $[5,105]$ & $0$ \\
red  & $\sigmas+\sigmaa$ & $1$ &$0$ & $1$  \\
\bottomrule
        \end{tabular}   
        }
    \caption{Material parameters}
    \label{fig_lattice_table}
\end{subfigure}
\begin{subfigure}[t!]{0.32\linewidth}
    \centering
 \includegraphics[width=\textwidth]{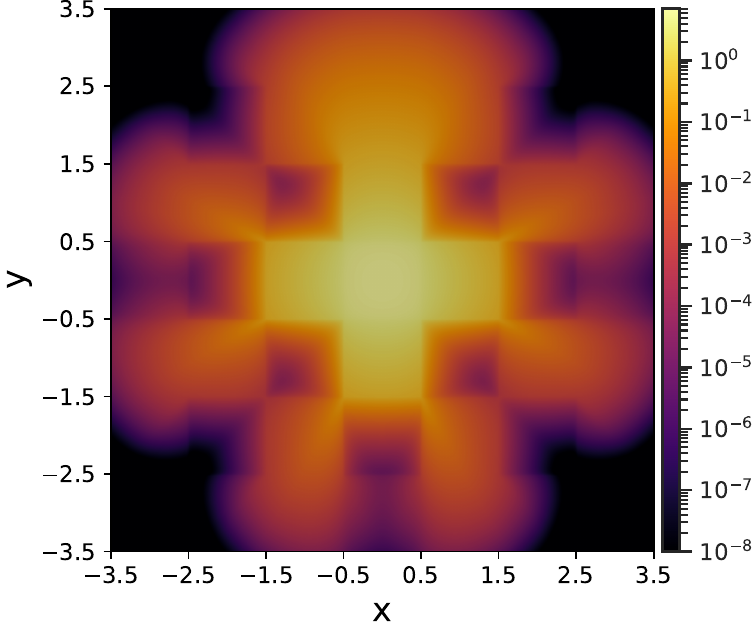}
    \vfill
    \caption{Reference solution}
\end{subfigure}
    \caption{The Lattice benchmark: (a) geometric layout; (b) material parameters; (c) particle concentration $\Phi$ from a high-resolution simulation, using the method described in \Cref{sec_numerics}.  The default values are $\sigmas^W=1$ for the scattering cross section in the white material and $\sigmaa^B=10$ for the absorption cross section in the blue material.}
    \label{fig_lattice}
\end{figure}

The Lattice benchmark, as described in \cite[Section 3.1]{reference_paper}, was first proposed in \cite{brunner2002forms}.  
It is characterized by a source that emits particles into a surrounding heterogeneous material environment.  
The layout, material parameters, and a plot of the particle concentration from a high-resolution simulation are provided in \Cref{fig_lattice}.
The two-dimensional domain $X$ is a collection of three types of materials arranged in a checkerboard pattern, i.e., the domain is partitioned into $X=B\cup R \cup W$, where $B$ is the collection of blue regions filled with highly absorbing material, $R$ is the red region containing the radiation source, and $W$ are the white void quadrants. 
The total, scattering, and absorption cross sections $\sigmat,\sigmas,\sigmaa$ of the respective subdomains are denoted by a corresponding superscript, e.g., $\sigmas^W$ for the scattering cross section in the white domain.

The domain is initially devoid of particles, but over time they enter via the source located in $R$; that is,
\begin{equation}
    Q = Q^R \chi_{R},
\end{equation}
where $Q^R = 1$. The final simulation time is $T = 3.2$.

\subsubsection{Quantities of interest}
\label{subsubsec_lattice_QoI}

Several different quantities of interest for the Lattice benchmark were defined in \cite{reference_paper}.  For the purposes of the current paper, we consider only  \textbf{QoI-3} from \cite[Section 3.1]{reference_paper}, which gives radiation absorption at final time $T$:
    \begin{align}\label{eq_abs_qoi_lattice}
         {\textup{QoI}}_{\textup{Lattice}} = \textup{QoI}_{B}[\Phi],
         \qquad
         \textup{QoI}_{S}[\Phi] := \int_{S} \sigmaa(\bx) \Phi(\bx,T)  d\bx,
    \end{align}
{where $\textup{QoI}_{S}[\Phi]$ denotes the absorption in a subdomain $S\subset X$ at the final time. The same functional with different choices of $S$ defines the Hohlraum QoIs in \Cref{subsubsec_hohlraum_QOI}, the QoI loss \eqref{eq_qoi_loss}, and the error metric \eqref{eqn:relative_error_pct}.}
In this work, we consider only neural-network-based surrogate models that map the material/source configuration and benchmark-specific boundary data to the final-time particle concentration. The remaining quantities of interest defined in \cite[Section 3.1]{reference_paper}, which require information of the particle concentration at all time steps or kinetic-level information from $\Psi$, are out of scope and deferred to future work. 

\subsubsection{Design Parameters}\label{sec:lattice_dp}
The Lattice benchmark in \cite[Section 3.1]{reference_paper} is equipped with design parameters for outer-loop applications such as design optimization or uncertainty quantification. We consider the design parameter set:
\begin{itemize}
 \item\textbf{DP-1}: the absorption cross-section $\sigmaa^B$ and the scattering cross-section $\sigmas^W$
\end{itemize}
The parameter space for $\textbf{DP-1}$ is two-dimensional. The full Cartesian product of uniformly spaced discrete values 
\begin{equation}
    \sigmaa^B \in \{5, 7.5, \ldots, 102.5\}
    \qquand 
    \sigmas^W \in \{0.1, 0.6, \ldots, 9.6\}
\end{equation}
with equal spacings $2.5$ and $0.5$ produces 800 parameter configurations. Each configuration is simulated once and corresponds to one sample used in the surrogate modeling pipelines, with the runs partitioned into train, test, and validation splits defined in \Cref{tab:dataset-distribution}.

\subsection{Hohlraum benchmark}\label{sec_hohlraum}
The Hohlraum benchmark, as described in \cite[Section 3.2]{reference_paper}, is a linearized and symmetrized version of the nonlinear thermal radiative transfer problem introduced in \cite{brunner2002forms}. The linearizations were originally introduced in \cite{hauck2013collision,osti_1550352,CROCKATT2019455,CROCKATT2020109765}, and are based on the assumption that there is no coupling between the radiation and the material.

\begin{figure}[t]
\begin{subfigure}[t!]{0.38\linewidth}
    \centering
    \includegraphics[width=\linewidth]{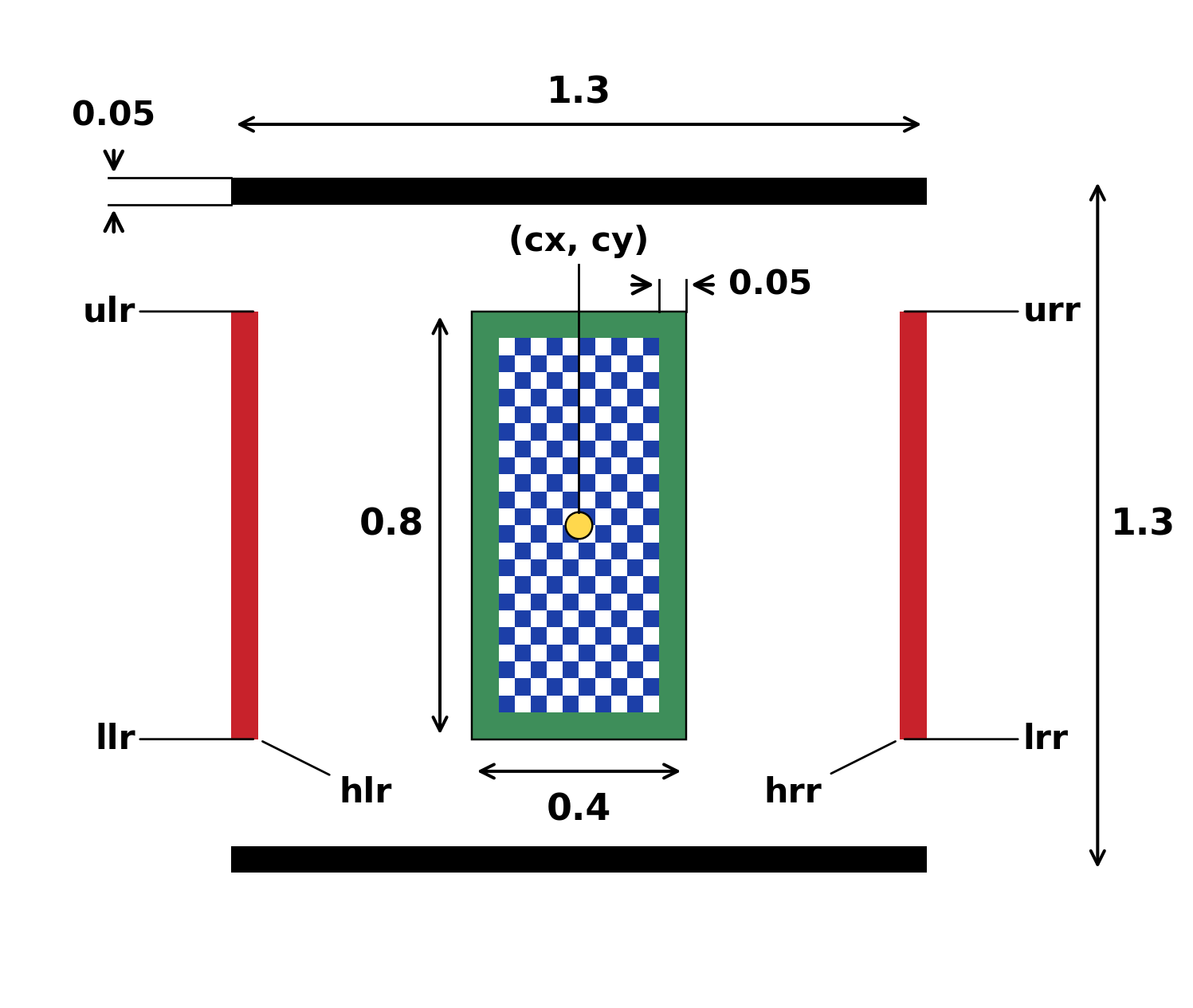}
    \vfill
    \caption{Geometric layout}
    \label{fig_hohlraum_schematic}
\end{subfigure}%
\hfill%
\begin{subfigure}[t!]{0.24\linewidth}
    \centering
    \vfill
    \resizebox{\linewidth}{!}{%
        \begin{tabular}{l | ccc}
            subdomain & $\sigmat$ & $\sigmas$ & $\sigmaa$ \\
            \midrule
            white          & $0.1$ & $0.1$ & $0$   \\
            blue-checkered & $100$ & $0$   & $100$ \\
            red            & $100$ & $95$  & $5$   \\
            green          & $100$ & $90$  & $10$  \\
            black          & $100$ & $50$  & $50$  \\
            \bottomrule
        \end{tabular}%
    }
    \caption{Material parameters}
    \label{fig_hohlraum_table}
\end{subfigure}%
\hfill%
\begin{subfigure}[t!]{0.34\linewidth}
    \centering
    \includegraphics[width=\linewidth]{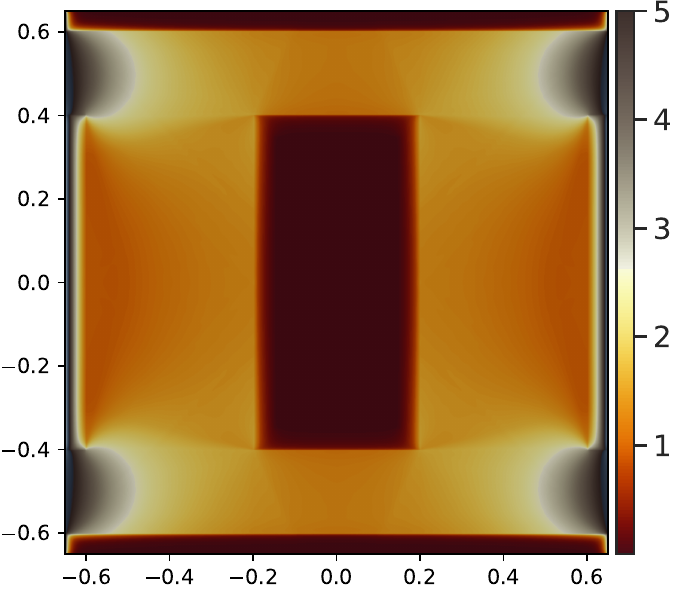}
    \vfill
    \caption{Reference solution}
    \label{fig_hohlraum_referece}
\end{subfigure}
\caption{The Hohlraum benchmark: (a) geometric layout The labels in (a) are the eight design parameters of \Cref{sec:hohlraum_dp}: $\mathrm{ulr},\mathrm{llr}$ and $\mathrm{urr},\mathrm{lrr}$ are the $y$ coordinates of the upper and lower edges of the left and right red strips (\textbf{DP-1}); $\mathrm{hlr}$ and $\mathrm{hrr}$ are the distances of the interior edges of the left and right red strips from the vertical symmetry axis, i.e., the interior edges lie at $x=-\mathrm{hlr}$ and $x=\mathrm{hrr}$ (\textbf{DP-2}); and $(c_x,c_y)$ is the center of the capsule (\textbf{DP-3}). The default configuration is $\mathrm{ulr}=\mathrm{urr}=0.4$, $\mathrm{llr}=\mathrm{lrr}=-0.4$, $\mathrm{hlr}=\mathrm{hrr}=0.6$, and $(c_x,c_y)=(0,0)$.; (b) material parameters; (c) particle concentration $\Phi$ from a high-resolution simulation, using the method described in \Cref{sec_numerics}. }
\label{fig_hohlraum}
\end{figure}

The Hohlraum geometry is defined on a square domain with shielding material along the top and bottom boundaries. At the center lies a rectangular capsule, whose outer layer is irradiated from the left and right sides of the domain. Additional shielding on these sides reduces the amount of radiation that directly reaches the capsule. The configuration, material properties, and a plot an approximate particle concentration from a high-resolution simulation are shown in \Cref{fig_hohlraum}.
We denote the red shielding material region at the left and right boundaries by $R$, the black shielding at the top and bottom of the domain by $K$, the green hull of the capsule by $G$, the  blue checkerboard interior of the capsule by $B$, and the white void region by $W$. Thus, the computational domain $X$ is partitioned into $X=R\cup K\cup G\cup B\cup W$. 
The total, scattering, and absorption cross sections $\sigmat,\sigmas,\sigmaa$ of the respective subdomains are denoted by a corresponding superscript, e.g., $\sigmas^W$ for the scattering cross section in the white domain.

There are no initial particles and no source in the Hohlraum benchmark. Thus $Q = 0$ and $\Psi_0 = 0$ in \eqref{eq_RTE}.
The inflow data in \eqref{eq_RTE} is given by
\begin{align}
\Psi^-(x,y,\mu,\varphi,t) =
\begin{cases}
1.0, \quad & x=-0.65, \;|y| < 0.65,\; \xi > 0 ,\\
1.0, \quad & x=0.65,\quad |y|< 0.65,\; \xi < 0 ,\\
0, & \text{otherwise},
\end{cases}
\end{align}
where $\xi = \sqrt{1-\mu^2} \cos \varphi$ with {{$(\mu,\varphi) \in [-1,1] \times [0,2 \pi)$}} is the horizontal component of the angular velocity in polar coordinates.  The final simulation time is $T =2.6$.

\subsubsection{Quantities of interest}
\label{subsubsec_hohlraum_QOI}
Of the seven quantities of interest for the Hohlraum test problem in \cite[Section 3.2]{reference_paper}, we consider a variation of \textbf{QoI-2}: the particle absorption in the various materials at the final time, i.e.
   \begin{align}\label{eq_abs_qoi_hohlraum}
    {\textup{QoI}}_{\textup{Hohlraum}}^S  = \textup{QoI}_{S}[\Phi] =\int_{S}  \sigmaa(\bx) \Phi(\bx,T)  d\bx
   \end{align}
where $S \in \{G \cup B, R, K\}${ and $\textup{QoI}_{S}[\cdot]$ is the absorption functional introduced in \eqref{eq_abs_qoi_lattice}}.

\subsubsection{Design parameters}\label{sec:hohlraum_dp}
Following \cite{reference_paper}, we consider the following design parameters that change the benchmark geometry, namely
\begin{itemize}
   \item \textbf{DP-1}: the location of the top and bottom edges of the red strips on each side,
   \item \textbf{DP-2}: the thickness of the red strips, via location of the interior edge,
    \item \textbf{DP-3}: the displacement of the capsule center.
\end{itemize}
{In \Cref{fig_hohlraum_schematic}, these parameters are labeled $\mathrm{ulr},\mathrm{llr},\mathrm{urr},\mathrm{lrr}$ (\textbf{DP-1}), $\mathrm{hlr},\mathrm{hrr}$ (\textbf{DP-2}), and $c_x,c_y$ (\textbf{DP-3}). They are treated as random variables when an outer loop workflow places a probability distribution on them.}
The scenario in \textbf{DP-1} involves as many as four parameters,  \textbf{DP-2} and \textbf{DP-3} as many as two parameters, summing to a total of $8$ design parameters.  
For \textbf{DP-1}, the $y$ coordinates of the top edges take values in $\{0.3,0.4,0.5\}$, and the $y$ coordinates of the bottom edges take values in $\{-0.5,-0.4,-0.3\}$, with default values $\pm 0.4$, respectively. 
For \textbf{DP-2}, the $x$ coordinates of the interior edges take values in $\{0.5,0.6,0.63\}$ and $\{-0.63,-0.6,-0.5\}$, with default values $\pm 0.6$. 
For \textbf{DP-3}, the $(x,y)$ coordinates of the capsule center point take values from $\{-0.1,0,0.1\}\times \{-0.075,0,0.075\}$, with default value $(0,0)$. Each configuration is simulated once and corresponds to one sample used in the surrogate modeling pipelines, with the runs partitioned into train, test, and validation splits defined in \Cref{tab:dataset-distribution}.

\section{Numerical discretization and training data generation}\label{sec_numerics}
 To create high-resolution reference solutions and training data for the neural network-based surrogates, we use the KiT-RT solver package \cite{kusch2023kit}\footnote{Software and source code are available at \url{https://github.com/KiT-RT/kitrt_code}. 
 We use  commit \texttt{d257b1a} in this paper. We recommend using the python wrapper CharmKit \url{https://github.com/ScSteffen/CharmKiT}.}. 
  KiT-RT uses a discrete-ordinates method \cite{carlson1968transport,azmy2010advances}, commonly referred to as the \sn method, which discretizes the angular domain using  quadratures to evaluate integrals over $\bbS^2$.  
  For simulations, we choose a tessellation quadrature \cite{JarrellAdams2011} of order $N_q$, which in general requires $8N_q^2$ quadrature points. However due to the symmetry of the problem, the quadrature need only be evaluated over the top half of the sphere, $\bbS_+^2$, so that the number of ordinates can be reduced to $4N_q^2$.  It is this reduced number that will be reported in all results.
  
  The system \eqref{eq_RTE} is discretized in space using a second-order finite-volume scheme on an unstructured triangular mesh and in time using and explicit second-order Runge Kutta method (SSP-RK2 or Heun) scheme in time.
For details of the numerical method, meshing strategy and parallelization, we refer to \cite[Section 4 and 5]{reference_paper}. 
Further, the time step is chosen as {$\Delta t = \mathrm{CFL}\cdot \min_{i}\sqrt{|K_i|}$, where $|K_i|$ is the area of cell $K_i$, the minimum is taken over all cells of the mesh, and the particle speed is one (this is the definition implemented in KiT-RT)}. The CFL number  $0.7$ for the Lattice benchmark and $0.8$ for the Hohlraum benchmark, following \cite{reference_paper}.

\begin{table}[t]
  \centering
  
  \caption{Mesh resolution across datasets. The choices for feasible low-resolution discretizations are inferred from \cite{reference_paper}. {The characteristic length is the target cell edge length passed to the mesh generator; the realized cell size follows from the cell count and the domain area.}}
  \label{tab:mesh-resolution}
  \resizebox{\textwidth}{!}{%
  \begin{tabular}{llccccccc}
    \toprule
    Resolution & Geometry & char. length & Min Cells & Max Cells & quad order & quadrature points & time steps & max. degrees of freedom \\
    \midrule
    High-res   & Lattice  & 0.015  & 79{,}862 & 79{,}862 & 6 & 144 & 978  & 11{,}500{,}128 \\
    High-res   & Hohlraum & 0.0075 & 69{,}897 & 81{,}775 & 6 & 144 & 1728 & 11{,}775{,}600 \\
    \midrule
    Low-res    & Lattice  & 0.05   & 7{,}364  & 7{,}364  & 4 & 64  & 176  & 471{,}296 \\
    Low-res    & Hohlraum & 0.02   & 10{,}196 & 11{,}958 & 6 & 144 & 662  & 1{,}721{,}952 \\
    \bottomrule
  \end{tabular}%
  }
\end{table}
\subsection{Data generation for surrogate model training}
We generate training and validation data for the Lattice and Hohlraum benchmarks using the KiT-RT solver. For each benchmark problem, we specify a ``high-resolution" and ``low-resolution" discretization in angle, space, and time for KiT-RT, described in 
\Cref{tab:mesh-resolution}. In  the Lattice benchmark, the computational grid is independent of the design parameters . However, in the Hohlraum benchmark, the design parameters prescribe the benchmark geometry and thus each set of design parameters requires re-meshing. Depending on the geometry, the generated mesh has a varying number of cells. We report the minimum and maximum number of grid cells in 
\Cref{tab:mesh-resolution}, since the runtime of KiT-RT scales with grid size. In both benchmarks, the angular mesh depends only on the quadrature order. Because each grid cell carries the information of all quadrature points, the total number of KiT-RT degrees of freedom is equal to the number of spatial cells times the number of quadrature points, .

For each benchmark problem, high-resolution samples are generated from the discrete Cartesian-product design sets over the parameter values specified in \Cref{sec:lattice_dp} and \Cref{sec:hohlraum_dp}. The Lattice design set contains $800$ unique parameter configurations, of which $707$ yielded complete and valid simulations. 
The full Hohlraum design set contains $6561$ unique parameter configurations. From this set, we sampled $1000$ configurations uniformly at random without replacement, of which $846$ yielded complete and valid simulations. 
Incomplete runs were due to job scheduling, timeout, or other infrastructure failures rather than known failures of the simulation code. No runs failed due to the KiT-RT solver itself.

Additionally, a set of low-resolution test samples is generated using the same parameter values as the high-resolution test samples, with the exception of the mesh generation parameters. The low-resolution simulations serve as a baseline for comparing the neural network-based surrogate models against a cheaper conventional solver in terms of compute cost and accuracy. 

We store the discrete representation of $\Phi$ for each finite-volume cell at final time $T$.  We also store the cell-centroid coordinates and cell areas. 
\begin{table}[t]
  \centering
  \caption{Dataset distribution for training and evaluation.}
  \label{tab:dataset-distribution}
  \begin{tabular}{lcc}
    \toprule
                & Lattice & Hohlraum \\
    \midrule
    Train       & 494     & 592      \\
    Validation  & 106     & 126      \\
    Test        & 107     & 128      \\
    \bottomrule
  \end{tabular}
\end{table}

\section{Neural network-based surrogates}\label{sec_ai}

\subsection{Surrogate Model Formulation}
We train surrogates that predict the spatial field $\Phi$ at time $T$ from the static material configuration and source field $Q$.  Let the spatial domain $X$ be discretized into $N_h$ mesh cells with centroids $\{\bx_i\}_{i=1}^{N_h} \subset X$ and a maximum diameter $h$. The cell coordinates are normalized to the unit box $[-1,1]^2$ before being input to the surrogate,
\begin{equation}\label{eq_coord_norm}
  \bar{\bx}_i = \frac{\bx_i - \bc}{r},
\end{equation}
where $\bc$ is the domain center and $r$ the domain half-extent; both benchmark domains are square and centered at the origin, so $\bc=\bzero$ and $\bar{\bx}_i = \bx_i / L$, with $L=3.5$ for the Lattice and $L=0.65$ for the Hohlraum benchmark. At each cell $i$, the material configuration is described by a vector of physical properties $\bp_i = (\sigma_{\rm{a},{i}}, \sigma_{\rm{s},{i}}, \sigma_{\rm{t},i}, Q_i)$ that contains the absorption, scattering, and total cross-sections and the source term.
Let $\bff_i$ denote an auxiliary feature vector at cell $i$ that encodes additional problem-specific information, in this case, Fourier feature defined in \Cref{sec_shared_surrogate_setup}. The surrogate model $\mathcal{F}_\theta$, parameterized by learnable weights $\theta$, is given by

\begin{align}\label{eq_surrogate_model}
    \mathcal{F}_\theta : \{(\bx_{i}, \bp_i, \bff_i)\}_{i=1}^{N}
    \quad \longmapsto \quad
    \{\hat{\Phi}_i\}_{i=1}^{N},
\end{align}
where $\hat{\Phi}_i \approx \Phi(\bx_{i}, T)$ is the predicted particle concentration in cell $i$ at the final simulation time $T$. The model parameters $\theta$ are optimized by minimizing a loss of the form

\begin{align}\label{eq:total-loss}
    \mathcal{L}(\theta) =  \mathcal{L}_{\text{region}}(\theta) + \lambda \, \mathcal{L}_{\text{QoI}}(\theta),
\end{align}
where $\lambda>0$ is a parameter. 

The first term in \eqref{eq:total-loss} is a region-weighted mean squared error (MSE) that assigns per-cell weights based on material type, which is a simplified version of learned region weights such as \cite{MCCLENNY2023111722} and cell area weighted loss in \cite{lam2023learning}:
\begin{align}\label{eq_region_loss}
    \mathcal{L}_{\text{region}}(\theta) = \frac{1}{W} \sum_{i=1}^{N_h} w_i\left( \hat{\Phi}_i - \Phi_i \right)^2, \quad W = \sum_{i=1}^{N_h} w_i,
\end{align}
where the weight $w_i$ is determined by the material label of cell $i$, taken from the benchmark layout; that is
\begin{equation}
w_i = \begin{cases} w_{\text{void}}, & \bx_i \in W,\\ w_{\text{mat}}, & \text{otherwise.}\end{cases}
\end{equation}
with $w_{\text{void}}=10$ and $w_{\text{mat}}=1$. We ablate the choice of $w_i$ in \Cref{sec_ablations}, with the baseline case setting $w_{\text{void}} = w_{\text{mat}}=1$.

The second term  in \eqref{eq:total-loss} is a quantity of interest alignment loss defined as,
\begin{align}\label{eq_qoi_loss}
    \mathcal{L}_{\text{QoI}}(\theta) = \left( \frac{ \textup{QoI}_{S}[\hat{\Phi}] - \textup{QoI}_{S}[\Phi]}{|\textup{QoI}_{S}[\Phi]| + \epsilon} \right)^2, 
\end{align}
{where $\textup{QoI}_{S}[\cdot]$ is the absorption functional of \eqref{eq_abs_qoi_lattice}, evaluated on the particle concentration $\hat{\Phi}$ computed with the surrogate model and on the high-resolution finite-volume solution $\Phi$, and $\epsilon$ is a small constant for numerical stability. For the Lattice benchmark \eqref{eq_abs_qoi_lattice} $S=B$ f and for the Hohlraum benchmark \eqref{eq_abs_qoi_hohlraum}, $S=G\cup B$.}

\subsection{Input feature design}\label{sec_input_feature_design}
The surrogate model families use spatial coordinates, material properties, and architecture-specific auxiliary features as inputs. The final hyperparameters and feature choices are described in \Cref{app:hyperparams}.

\paragraph{Fourier features.} \label{sec_shared_surrogate_setup}
Fourier features map low-dimensional coordinates into a higher-dimensional sinusoidal basis before they are passed to the neural network, making it easier for the model to represent high-frequency spatial variation \cite{tancik2020fourier}. For cell $i$ with normalized centroid $\bar{\bx}_i=(\bar x_i,\bar y_i)$ (\Cref{eq_coord_norm}), the deterministic multi-resolution encoding is
\begin{align}\label{eq_fourier_feature}
  \bff_i = \bff(\bar{\bx}_i) =
    \left[
        \left\{
            \sin(2\pi f_k \bar x_i),\,
            \cos(2\pi f_k \bar x_i),\,
            \sin(2\pi f_k \bar y_i),\,
            \cos(2\pi f_k \bar y_i)
        \right\}_{k=0,\ldots,K_{\mathrm{ff}}-1}
    \right],
    \quad f_k = f_0\, 2^k,
\end{align}
with $K_{\mathrm{ff}}=3 $ the number of Fourier frequencies and $f_0=2$ (so $f_k\in\{2,4,8\}$). These encoded coordinates are appended to the coordinate inputs for both surrogate models. An ablation of this optional input feature is provided as the FF factor of the Hohlraum ablation in \Cref{app:ablation_visualizations}, with per-cell field-prediction visualizations in \Cref{app:ablation_visualizations}.

\subsection{Transolver}

\subsubsection{Overview}

Transformers \cite{vaswani2017attention} are effective for sequence and field data. Their token-based communication is useful for radiation transport because it provides global coupling across distant mesh regions while remaining linear in the number of mesh cells for fixed token count. Because the slice assignments are learned from cell features rather than tied to a fixed grid, the model can be applied to the unstructured meshes and varying geometries considered here.

However, the underlying mechanism  of a transformer provides weighted communication between points, which scales quadratically with the number of input points.   Transolver \cite{wu2024transolver} reduces this cost with Physics-Attention. In Physics-Attention, a learned projection from input features to physics slices is used to softly assign mesh points under similar physical state to the same slice. The slice assignment is soft - instead learning a one-to-one mapping of input point to physics slice, each distinct spatial point receives a normalized weighted distribution over the $M$ physics slices. After slice weights are assigned, a spatially weighted slice aggregation encodes each physics slice into a single physics-aware token. Self-attention is applied over these tokens, and the updated token information is dispatched back to the cells {, i.e., each cell receives the combination of the updated tokens weighted by its own slice weights}.

\subsubsection{Transolver surrogate instantiation}
\label{sec_transolver_recap}

We use the Transolver architecture of \cite{wu2024transolver} without modifying its core Physics-Attention mechanism; this subsection summarizes the architecture to specify its task-specific inputs, outputs, and computational scaling for the present finite-volume transport meshes. The full model definition can be found in \Cref{sec_transolver}.

For each mesh cell $i$, the raw input is assembled from the normalized cell-centroid coordinates $\bar{\bx}_i$, optional Fourier features $\bff_i$, and per-cell material/source quantities
\[
    \bp_i=(\sigma_{\rm{a},i},\sigma_{\rm{s},i},\sigma_{\rm{t},i},Q_i).
\]
The source term $Q_i$ is omitted for the Hohlraum, since $Q = 0$ in that case. A pointwise input embedding $E_{\mathrm{in}}$ maps these quantities to a latent cell representation,
\begin{align}
    \bh_i^0
    =
    E_{\mathrm{in}}(\bar{\bx}_i,\bp_i,\bff_i)
    \in \mathbb{R}^{C},
    \qquad
    i=1,\ldots,N_h .
\end{align}

Each Transolver layer updates the cell states through Physics-Attention. Instead of applying all-pairs attention over the $N_h$ mesh cells, the layer learns a soft assignment of cells to $M$ physics-aware slices, aggregates each slice into a latent token, applies multi-head self-attention over the resulting $M$ tokens, and dispatches the updated token information back to the cells using the same slice assignments. This gives global communication across the mesh through $M\ll N_h$ learned tokens rather than through direct pointwise attention between all cell pairs. After $L$ Transolver layers, a pointwise output embedding maps the final hidden state at each cell to the predicted particle concentration,
\begin{align}
    \hat{\Phi}_i
    =
    E_{\mathrm{out}}(\bh_i^L),
    \qquad
    i=1,\ldots,N_h .
\end{align}

Overall, the Transolver surrogate can be written as
\begin{align}
    \mathcal{F}_\theta
    =
    E_{\mathrm{out}}
    \circ
    \mathcal{T}_L
    \circ
    \cdots
    \circ
    \mathcal{T}_1
    \circ
    E_{\mathrm{in}},
\end{align}
where $\mathcal{T}_\ell$ denotes the $\ell$-th Physics-Attention Transolver layer. The trainable parameters consist of the input and output embeddings, the slice-assignment projections, token self-attention weights, dispatch projections, and pointwise MLPs.

\subsubsection{Computational cost}

For cost accounting, one Physics-Attention layer first assigns $N_h$ cell states to $M$ learned physics tokens, then applies self-attention over those $M$ tokens, and finally dispatches the token information back to the cells. The cell-to-token aggregation and token-to-cell dispatch scale as
$
    \mathcal{O}(N_h M C),
$
where $C$ is the latent channel dimension. The token self-attention step scales as
$
    \mathcal{O}(M^2 C).
$
Thus, the attention-specific cost of one Transolver layer is
$
    \mathcal{O}(N_h M C + M^2 C).
$
Including the fixed-depth pointwise projections and MLPs with $\mathcal{O}(N_hMC)$ gives a per-layer cost of
\begin{equation}
    \mathcal{O}\!\left(N_hC^2+M^2C+N_hMC\right),
\end{equation}
if the latent width $C$ is treated as variable. 
In the experiments, $(L,M,C)$ are fixed architecture parameters, so the forward-pass cost scales linearly with the number of mesh cells $N_h$. This replaces the $\mathcal{O}(N_h^2C)$ attention cost of standard pointwise self-attention over all mesh cells by attention over $M\ll N_h$ learned physics tokens.

\subsubsection{Architecture choices}
When Fourier features are enabled, the coordinate input is augmented by $K_{\mathrm{ff}}=3$ Fourier frequencies, giving $12$ additional positional channels on top of the two raw coordinates. All other Transolver architectural parameters are fixed across benchmark problems and ablation configurations: $L=8$ (number of Transolver layers), $C=256$ (latent channel width), $H_{\mathrm{attn}}=16$ (attention heads), $d_h=C/H_{\mathrm{attn}}=16$ (per-head dimension), $M=128$ (physics slices/tokens), MLP ratio $4$ (pointwise-MLP hidden expansion, hidden width $4C$), dropout $0$. The resulting trainable parameter counts are reported in \Cref{tab:training-cost}. We ablate the inclusion of Fourier features and region-weighted MSE loss.

\subsubsection{Training Transolver}\label{sec_transolver_training}
The Transolver surrogates for both Lattice and Hohlraum are trained as an operator from the static problem specification to the particle concentration at time $t=T$, see \eqref{eq:total-loss}, on the high-resolution training data split of \Cref{tab:dataset-distribution}. The composite objective in \Cref{eq:total-loss} is used during training, with QoI loss weight $\lambda=5\times 10^{-3}$ for the Lattice benchmark and $\lambda=10^{-2}$ for the Hohlraum benchmark. For the Lattice benchmark, the QoI in the loss term \eqref{eq_qoi_loss} is defined in \eqref{eq_abs_qoi_lattice}. For the Hohlraum benchmark, we use the QoI defined in \eqref{eq_abs_qoi_hohlraum} for $S=G\cup B$, and reserve $S\in \{R,K\}$ to measure extrapolation across QoIs not explicitly included in the loss.

Training uses distributed data parallelism (DDP) on a single DGX H100 node with 8 H100 80\,GB GPUs. Each GPU processes one sample per optimization step, giving an effective batch size of 8. We use a hybrid optimizer where matrix-valued parameters are optimized with Muon \cite{jordan2024muon} (momentum $\beta=0.95$) and the remaining one-dimensional parameters with Adam \cite{kingma2015adam}, both at a shared learning rate $3\times 10^{-5}$. The learning rate follows a cosine schedule with a linear 10 epoch warm-up from $10^{-6}$ to the peak and then annealing back to the minimum at the final epoch. A global parameter gradient $\ell^2$-norm is computed and used to rescale the gradients such that the norm does not exceed $10$ before the optimizer update to aid in stabilizing model training. BF16 automatic mixed precision is enabled throughout. Each model is trained for a fixed budget of 500 epochs, with the best model selected by lowest total validation loss. Training wall time and GPU-hours are reported in \Cref{tab:training-cost}, along with complete optimizer and schedule settings.

\begin{figure}
    \centering
    \begin{subfigure}[t]{0.48\linewidth}
        \centering
        \includegraphics[width=\linewidth]{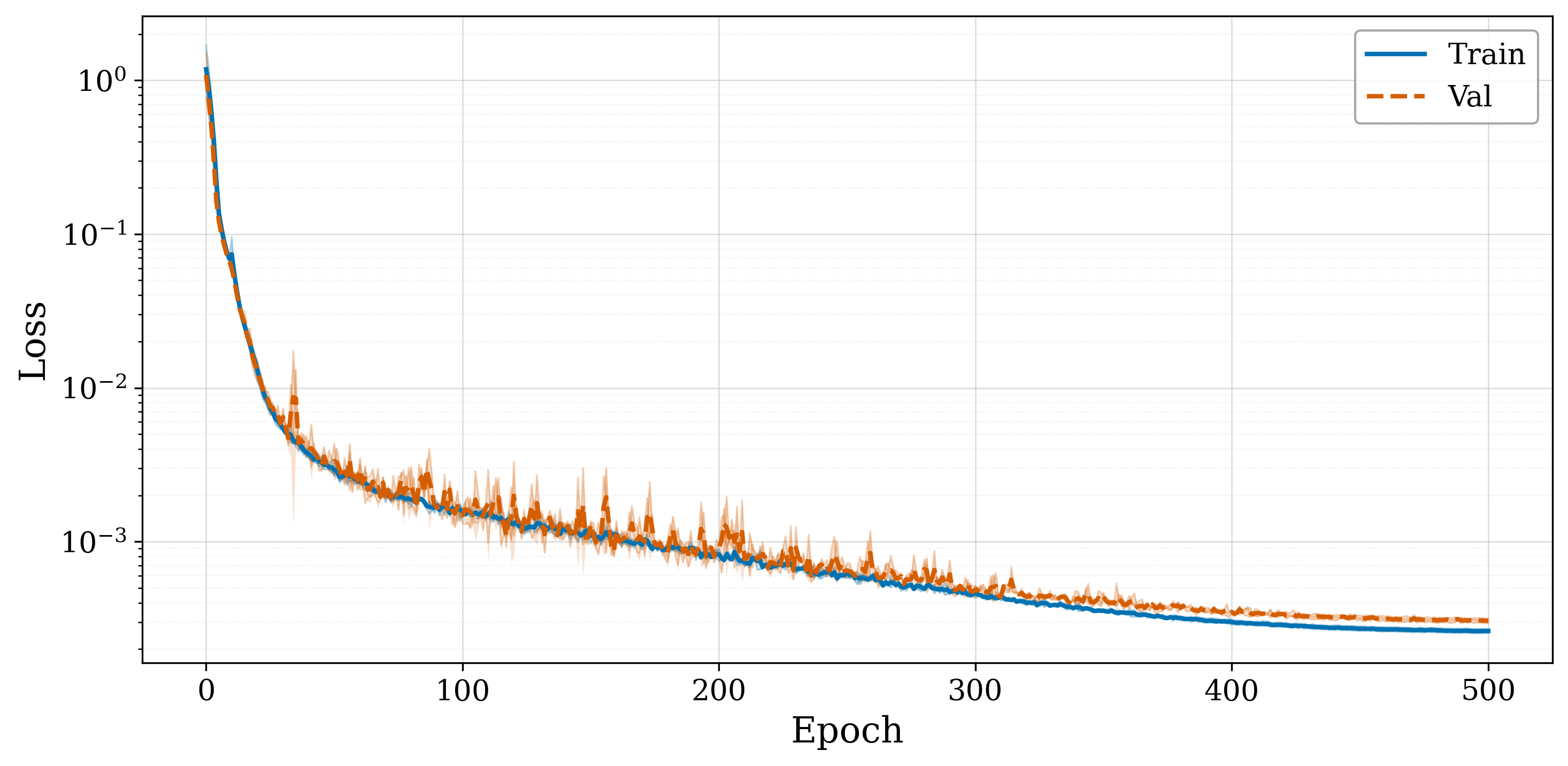}
        \caption{Transolver, Lattice training and validation.}
        \label{fig_transolver_lattice_loss}
    \end{subfigure}
    \hfill
    \begin{subfigure}[t]{0.48\linewidth}
        \centering
        \includegraphics[width=\linewidth]{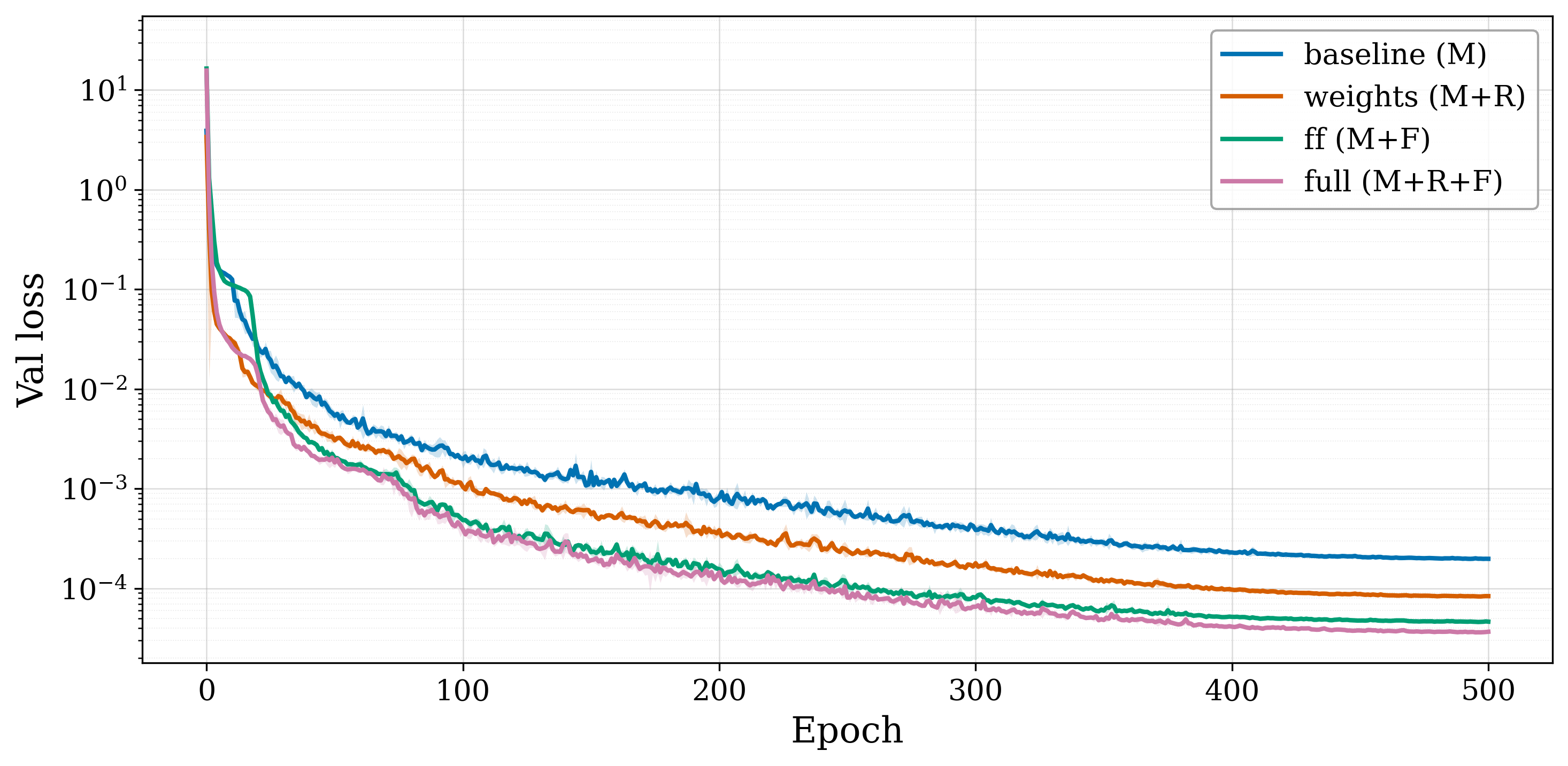}
        \caption{Transolver, Hohlraum validation loss.}
        \label{fig_transolver_hohlraum_loss}
    \end{subfigure}
    \caption{Transolver training and validation loss curves (see \Cref{eq:total-loss}) on the training and validation split of \Cref{tab:dataset-distribution}. 
    Solid lines show the median across three random seeds and shaded bands show the seed-wise min-max envelope; we select the lowest-validation-loss checkpoint per seed for evaluation. 
    The Lattice panel shows the baseline configuration, the only configuration trained on Lattice. 
    The Hohlraum panel overlays all four configurations of the Hohlraum ablation validation loss curves (\Cref{sec_ablations}): Base, +RW, +FF, and +FF+RW.}
    \label{fig_transolver_loss}
\end{figure}

\subsection{Bi-Stride Multi-Scale MeshGraphNet (BSMS-MGN)}

\subsubsection{Overview}

Graph neural networks (GNNs)\cite{xu2018powerful} operate directly on irregular and unstructured meshes, where convolutional architectures designed for regular grids do not apply \cite{bronstein2017geometric, battaglia2018relational, sanchez2020learning}. MeshGraphNet \cite{pfaff2020learning} is an encode-process-decode architecture whose processor is a stack of message-passing (MP) blocks that exchange information along the edges of the mesh graph.  GNNs are a natural match for the KiT-RT meshes. Nodes and edges correspond to finite-volume cells and their faces, and since the model does not depend on a fixed grid, the same trained network can be applied to the varying Hohlraum meshes.

Standard message passing is typically local. Thus if the graph is based on the cell-dual graph of the mesh, then $K$ message passing blocks couple only cells that are at most $K$ edges apart. This type of locality is a limitation for streaming-dominated transport, where particles emitted at the source or at the inflow boundary must reach absorbers on the far side of the domain within a single forward pass. Bi-Stride Multi-Scale MeshGraphNet (BSMS-MGN) \cite{cao2022bi} address this type of challenge with a U-Net-style \cite{ronneberger2015unetconvolutionalnetworksbiomedical} multiscale processor that runs message passing on a hierarchy of coarsened graphs, so that one hop on a coarse level spans many cells of the mesh. The coarse graphs are generated by bi-stride pooling, a parameter-free construction described below.

\subsubsection{BSMS-MGN surrogate instantiation}\label{sec_bsms_mgn_recap}

We use the BSMS-MGN architecture of \cite{cao2022bi} as implemented in PhysicsNeMo \cite{PhysicsNeMo_Contributors_NVIDIA_PhysicsNeMo_An_2023} without modification; this subsection specifies its graph construction, task-specific features, and computational scaling. The complete definition is given in \Cref{sec_bsms_mgn_definition}.

\paragraph{Graph.}
The finest graph $\mathcal{G}^{(0)}=(\mathcal{V}^{(0)},\mathcal{E}^{(0)})$ is the graph corresponding to the finite-volume mesh. It has one node per cell, placed at the cell centroid $\bx_i$, and a pair of directed edges $(i,j)$ and $(j,i)$ for every two cells that share a face. On a triangular mesh every interior node has three neighbors, so $|\mathcal{E}^{(0)}|\le 3N_h$. Node $i$ carries the same input as in the Transolver case, i.e., the normalized centroid $\bar{\bx}_i$ of \eqref{eq_coord_norm}, the material and source vector $\bp_i$ (without $Q_i$ for the Hohlraum, where $Q=0$), and the optional Fourier features $\bff_i$. Edge\footnote{In this section, "edge" describes the edge of a graph consumed by the BSMS-MGN, and "face" the interface (or edge) between to grid cells of the original computational grid of the radiation transport equation.} $(i,j)$ carries the geometric offset between the two centroids, 
\begin{align}
\bfe_{ij}^{0}
=
[\Delta x_{ij},\,\Delta y_{ij},\,\|\Delta\bx_{ij}\|],
\qquad
\Delta\bx_{ij}=\bx_{i}-\bx_j .
\end{align}

\paragraph{Graph hierarchy.}
The coarser graphs $\mathcal{G}^{(1)},\ldots,\mathcal{G}^{(L_g)}$ are built by bi-stride pooling \cite{cao2022bi}. Starting from a seed node, the nodes of $\mathcal{G}^{(l)}$ are ordered by a breadth-first search, every second breadth-first front is kept, and two kept nodes are connected in $\mathcal{G}^{(l+1)}$ if they are at most two edges apart in $\mathcal{G}^{(l)}$. Each level therefore has roughly half as many nodes as the previous one, and one hop on level $l$ corresponds to $2^{l}$ hops on the mesh graph. Node states are transferred between levels by fixed averaging operators (pooling) and their transposes (unpooling), which play the role of restriction and prolongation in a multigrid method. The hierarchy and the transfer operators contain no trainable parameters; they are computed once per mesh and stored with the dataset.

\paragraph{Network.}
An encoder $\mathrm{Enc}=(\mathrm{Enc}_v,\mathrm{Enc}_e)$ maps the node and edge inputs to latent states $\ssbold{n}_i^{0},\,\ssbold{g}_{ij}^{0}\in\mathbb{R}^{C_g}$. The processor first applies $K_{\mathrm{mp}}$ message-passing blocks $\mathcal{P}_0,\ldots,\mathcal{P}_{K_{\mathrm{mp}}-1}$ on $\mathcal{G}^{(0)}$. Block $k$ updates every edge from the states of its two end nodes and then every node from the sum of its incoming edge states,
\begin{align}\label{eq_mp_block}
\ssbold{g}_{ji}^{k+1} &= \ssbold{g}_{ji}^{k} + \mathrm{MLP}_e^{k}\big([\ssbold{g}_{ji}^{k},\,\ssbold{n}_j^{k},\,\ssbold{n}_i^{k}]\big),
& (j,i)&\in\mathcal{E}^{(0)}, \\
\ssbold{n}_i^{k+1} &= \ssbold{n}_i^{k} + \mathrm{MLP}_v^{k}\Big(\big[\ssbold{n}_i^{k},\,\textstyle\sum_{j:(j,i)\in\mathcal{E}^{(0)}}\ssbold{g}_{ji}^{k+1}\big]\Big),
& i&\in\mathcal{V}^{(0)},
\end{align}
where $\mathrm{MLP}_e^{k}$ and $\mathrm{MLP}_v^{k}$ are the edge and node networks of block $k$ and $[\cdot]$ denotes concatenation. After $K_{\mathrm{mp}}$ blocks, each cell has exchanged information with all cells within $K_{\mathrm{mp}}$ faces  of it. The processor then performs one U-Net pass $\mathcal{U}_{L_g}$ through the hierarchy: on the way down it applies one message-passing block on each level and pools the node states to the next coarser level, at level $L_g$ it applies one block, and on the way up it unpools the states, adds the states saved on the way down (skip connections), and applies one block per level. The blocks of the U-Net pass have the form \eqref{eq_mp_block} but recompute their edge inputs from the positional offsets and the node states at every call instead of carrying an edge state, and every block has its own weights. Writing $\ssboldnKmp=(\ssbold{n}_1^{K_{\mathrm{mp}}},\ldots,\ssbold{n}_{N_h}^{K_{\mathrm{mp}}})$ for the finest-level states after the $K_{\mathrm{mp}}$ blocks, a decoder maps the output of the U-Net pass at cell $i$ to the prediction,
\begin{align}
    \hat{\Phi}_i = \mathrm{Dec}\Big(\big[\mathcal{U}_{L_g}(\ssboldnKmp)\big]_i\Big), \qquad i=1,\ldots,N_h ,
\end{align}
so that the surrogate reads
\begin{align}
\mathcal{F}_\theta
=
\mathrm{Dec}
\circ
\mathcal{U}_{L_g}
\circ
\mathcal{P}_{K_{\mathrm{mp}}-1}
\circ
\cdots
\circ
\mathcal{P}_{0}
\circ
\mathrm{Enc}.
\end{align}
The trainable parameters are the encoder and decoder MLPs and the edge and node MLPs of the $K_{\mathrm{mp}}$ finest-level blocks and of the $2L_g+1$ blocks of the U-Net pass. The receptive field of one forward pass is $\mathcal{O}(K_{\mathrm{mp}}+2^{L_g})$ cells: the finest-level blocks reach $K_{\mathrm{mp}}$ cells and the U-Net pass about $5\cdot 2^{L_g}$ cells (\Cref{sec_bsms_mgn_definition}).

\subsubsection{Computational cost}

One message-passing block on level $l$ evaluates the edge MLP once per directed edge and the node MLP once per node. With MLPs of fixed depth and width $C_g$, each evaluation costs $\mathcal{O}(C_g^2)$, and since the node degree is bounded, one block costs $\mathcal{O}(|\mathcal{E}^{(l)}|\,C_g^2)$. The finest-level processor therefore costs $\mathcal{O}(K_{\mathrm{mp}}\,|\mathcal{E}^{(0)}|\,C_g^2)$. The U-Net pass visits each level at most twice and the edge counts decrease geometrically with the level, so its cost is $\mathcal{O}(|\mathcal{E}^{(0)}|\,C_g^2)$. Including the encoder and decoder, the full forward pass costs
\begin{equation}
    \mathcal{O}\!\left((K_{\mathrm{mp}}+1)\,|\mathcal{E}^{(0)}|\,C_g^2\right)
    =
    \mathcal{O}\!\left(N_h\,C_g^2\right),
\end{equation}
which is linear in the number of cells for fixed $(K_{\mathrm{mp}},L_g,C_g)$, as is the Transolver cost of \Cref{sec_transolver_recap} summed over its $L$ layers. The measured inference times of BSMS-MGN are nevertheless two to three times longer (\Cref{tab:Lattice-eval,tab:Hohlraum-eval}), partly because the constant in front of $N_h$ is larger and partly because message passing requires irregular gather and scatter memory access, whereas Transolver operates on dense token matrices.

\subsubsection{Architecture choices}\label{bsms-arch-choices}

We use $K_{\mathrm{mp}}=16$ finest-level blocks, $L_g=5$ coarsening levels, and latent width $C_g=156$, chosen so that the trainable parameter count matches the Transolver baseline (\Cref{tab:training-cost}). All encoder, message-passing, and decoder MLPs have two hidden layers of width $C_g$ with ReLU activations; all but the decoder apply LayerNorm to their output. The remaining settings are the defaults of the PhysicsNeMo implementation \cite{PhysicsNeMo_Contributors_NVIDIA_PhysicsNeMo_An_2023} and are listed in \Cref{tab:mgn-hparams}. As for Transolver, we ablate the Fourier features and the region-weighted loss.

\subsubsection{Training BSMS-MGN}

\begin{figure}
    \centering
    \begin{subfigure}[t]{0.48\linewidth}
        \centering
        \includegraphics[width=\linewidth]{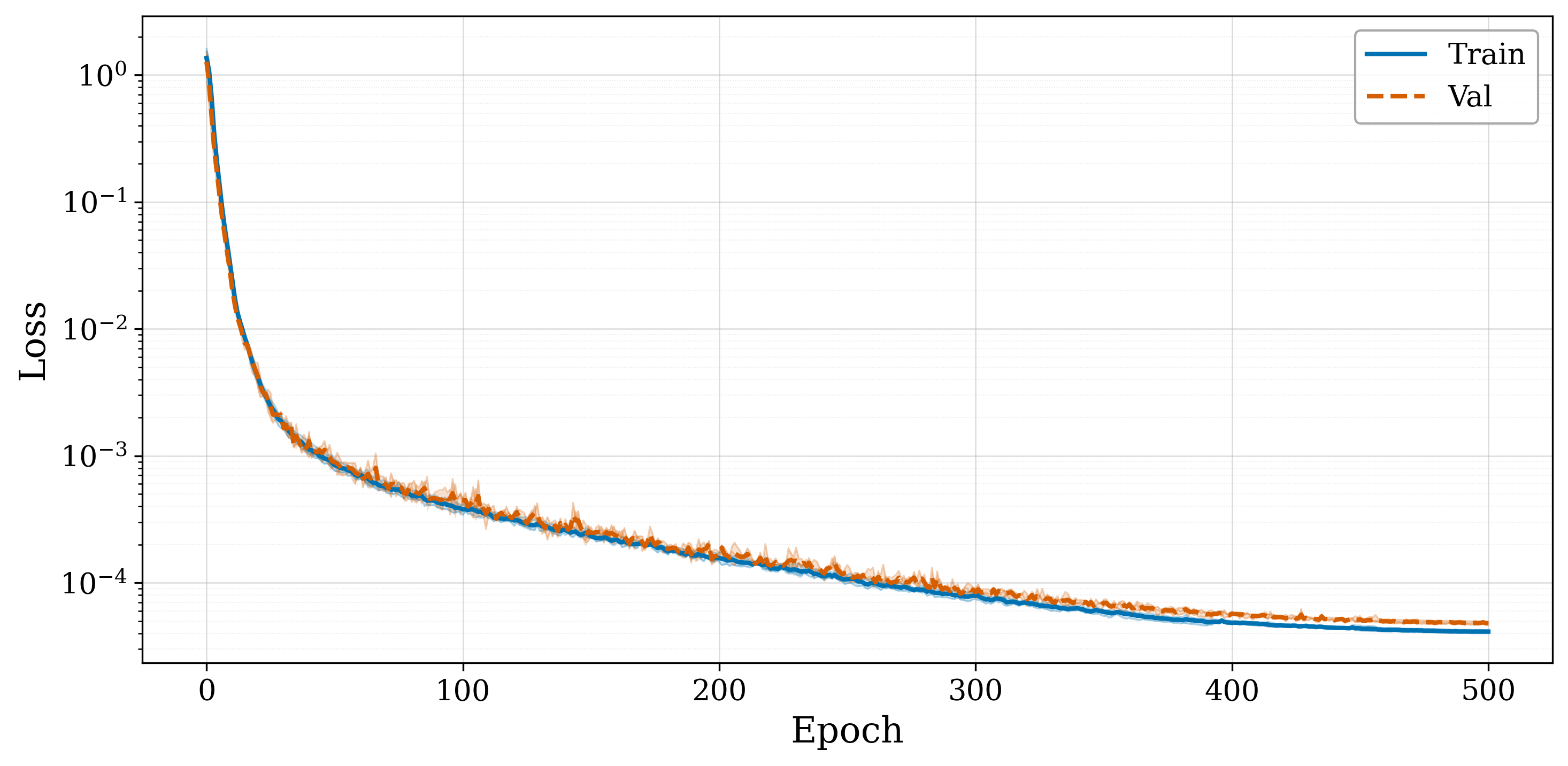}
        \caption{BSMS-MGN, Lattice training and validation.}
        \label{fig_bsms_mgn_lattice_loss}
    \end{subfigure}
    \hfill
    \begin{subfigure}[t]{0.48\linewidth}
        \centering
        \includegraphics[width=\linewidth]{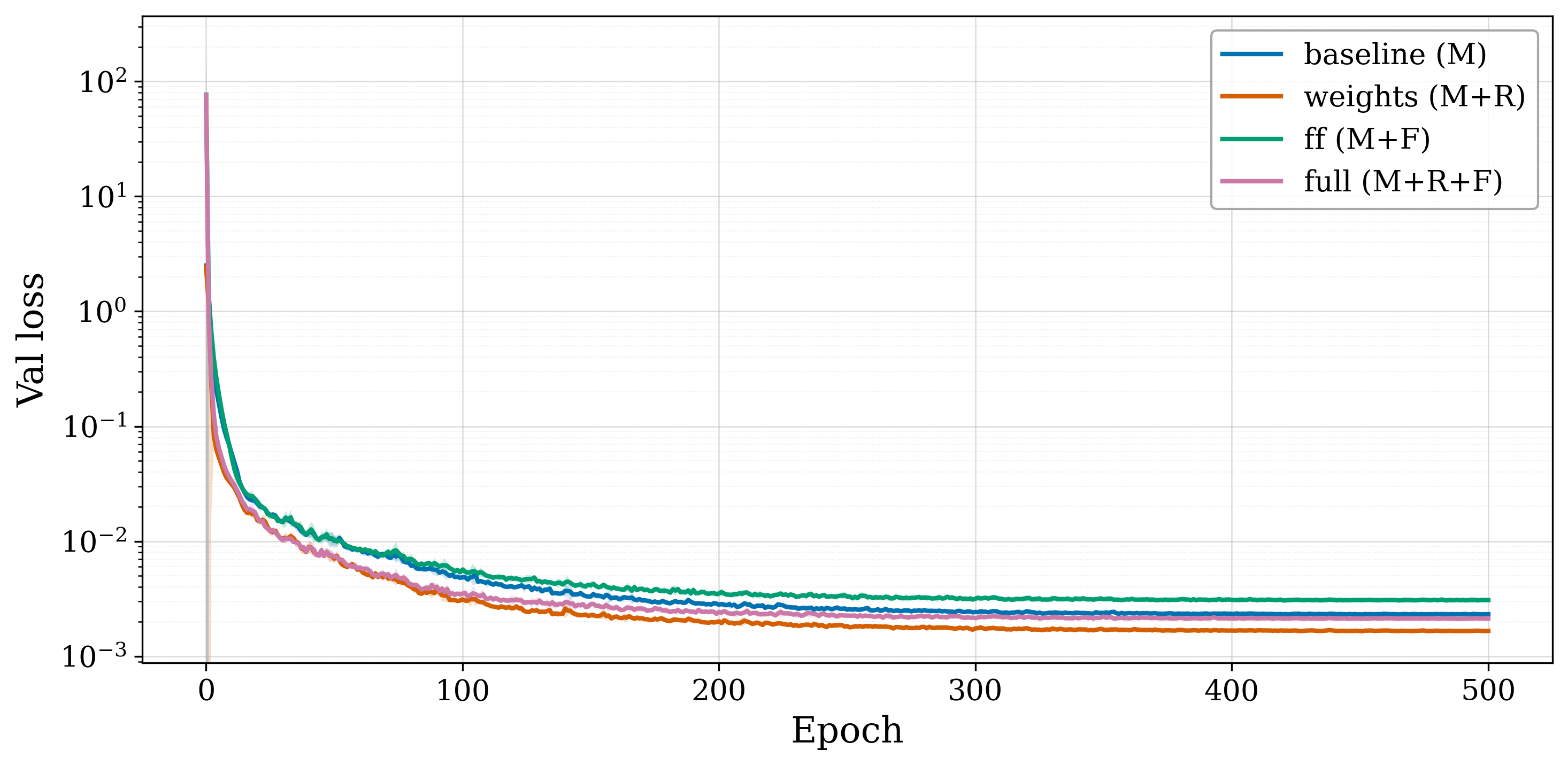}
        \caption{BSMS-MGN, Hohlraum validation.}
        \label{fig_bsms_mgn_hohlraum_loss}
    \end{subfigure}
    \caption{BSMS-MGN training and validation loss curves (see \Cref{eq:total-loss}) on the training and validation split of \Cref{tab:dataset-distribution}. Solid lines show the median across three random seeds and shaded bands show the seed-wise min-max envelope; we select the lowest-validation-loss checkpoint per seed for evaluation. The Lattice panel shows the baseline configuration, the only one trained on Lattice. The Hohlraum panel overlays all four configurations of the Hohlraum ablation validation loss: Base, +RW, +FF, and +FF+RW.}
    \label{fig_bsms_mgn_loss}
\end{figure}

The BSMS-MGN surrogates are trained on the same final-time task as Transolver with the composite loss \eqref{eq:total-loss} and the same QoI loss weights, $\lambda=5\times 10^{-3}$ for the Lattice and $\lambda=10^{-2}$ for the Hohlraum benchmark. The QoI in \eqref{eq_qoi_loss} is \eqref{eq_abs_qoi_lattice} for the Lattice benchmark and \eqref{eq_abs_qoi_hohlraum} with $S=G\cup B$ for the Hohlraum benchmark; $S\in\{R,K\}$ is reserved to measure extrapolation to QoIs not included in the loss. Hardware, optimizer, learning-rate schedule, gradient clipping, mixed precision, training budget, and checkpoint selection are identical to \Cref{sec_transolver_training}; each GPU processes one graph per optimizer step, giving an effective batch size of $8$. Training wall time and GPU-hours are reported in \Cref{tab:training-cost}, the loss curves in \Cref{fig_bsms_mgn_loss}.

\section{Numerical results}\label{sec_results}
In this section, we present the numerical results of the Transolver and BSMS-MGN surrogate models for the Lattice and Hohlraum benchmarks. The surrogate models are trained on the high-resolution mesh and evaluated on both the high and low-resolution spatial mesh. The training uses only the particle concentration $\Phi$; thus the surrogates have no knowledge of the angular discretization of the \sn finite-volume solver. The trained surrogates are mesh-agnostic, per-cell operators (\Cref{eq_surrogate_model}) - evaluation on the low-resolution mesh reuses the same trained models with the identical input construction of \Cref{sec_ai}, namely normalized cell-centroid coordinates, material properties, and Fourier features, now assembled at the low-resolution cell centroids. In particular, the Fourier features \eqref{eq_fourier_feature} are recomputed at the new spatial points, so no retraining or interpolation is required.

We benchmark both surrogates against the KiT-RT solver on two axes, namely, accuracy and wall-clock cost, both measured relative to the high-resolution KiT-RT reference solution. The low-resolution KiT-RT solve serves as the cheap baseline that a useful surrogate must beat on the accuracy vs. cost trade-off, while the high-resolution solver outputs are the reference used to generate the training data with an unseen, heldout test set reserved for evaluation. Throughout, high- and low-resolution refer to the discretizations of \Cref{tab:mesh-resolution}. \Cref{sec_high_res} presents the high-resolution surrogate evaluation (the mesh the surrogates were trained on), Lattice first and then Hohlraum, and \Cref{sec_ablations} reports the high-resolution Hohlraum ablation. \Cref{sec_low_res} summarizes cross-mesh transfer to the coarser low-resolution mesh, with the full tables, field visualizations, and discussion in \Cref{app:low_res}. \Cref{sec_efficiency_amortization} reports the inference efficiency and amortization analysis.

For each surrogate we report two accuracy measures against the high-resolution KiT-RT reference, both in physical particle-concentration units and both computed per test sample and then averaged over the hold-out test set (\Cref{tab:dataset-distribution}); the tables report the mean and standard deviation of these test-set averages across the three random seeds. The first is the field root-mean-square error (RMSE) of the final-time particle concentration, computed over the $N_h$ mesh cells of a sample,
\begin{align}\label{eqn:rmse}
  \mathrm{RMSE}(\hat\Phi,\Phi) = \sqrt{\frac{1}{N_h}\sum_{i=1}^{N_h}\bigl(\hat\Phi_i - \Phi_i\bigr)^2}.
\end{align}
The second is the relative error in the absorption QoI, denoted by
\begin{align}\label{eqn:relative_error_pct}
  E_{\textup{QoI},\textup{rel}} = 100   \,
    \frac{
        \left|
        \textup{QoI}_{S}[\Phi_{\textup{method}}]
        -
        \textup{QoI}_{S}[\Phi_{\textup{HR}}]
        \right|
    }{
        \left|
        \textup{QoI}_{S}[\Phi_{\textup{HR}}]
        \right|
        + \epsilon
    },
\end{align}
where $\Phi_{\textup{HR}}$ denotes the high-resolution KiT-RT particle concentration, and $\Phi_{\textup{method}}$. For the Lattice table, $S=B$ as in \eqref{eq_abs_qoi_lattice}. For the Hohlraum table, $S\in\{G\cup B, R, K\}$ as in \eqref{eq_abs_qoi_hohlraum}, evaluated separately.

\subsection{High-resolution evaluation}\label{sec_high_res}

The high-resolution evaluation reports each surrogate on the same mesh resolution used to generate its training data (\Cref{tab:mesh-resolution}). The Lattice benchmark is reported in its single baseline configuration, and the Hohlraum benchmark is reported across all four ablation configurations of region weighting and Fourier feature usage.  

\subsubsection{Lattice}\label{sec_high_res_lattice}

\begin{figure}[t]
    \centering

    \begin{subfigure}{1\linewidth}
        \centering
        \includegraphics[width=1\linewidth]{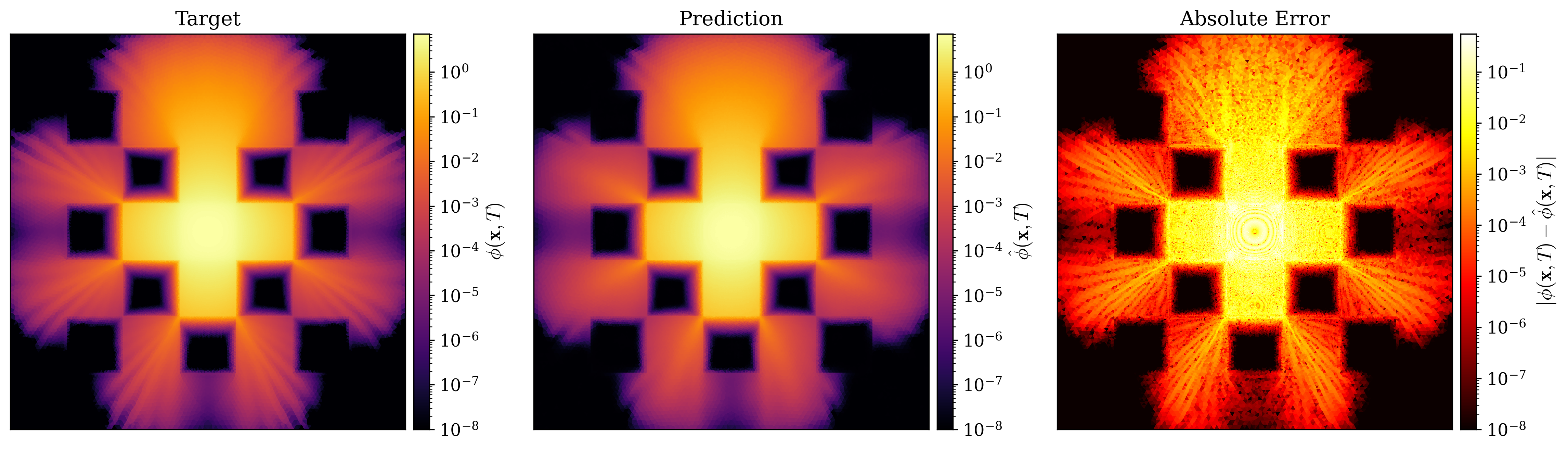}
        \caption{Transolver.}
        \label{fig:transolver_lattice}
    \end{subfigure}

    \vspace{0.75em}

    \begin{subfigure}{1\linewidth}
        \centering
        \includegraphics[width=1\linewidth]{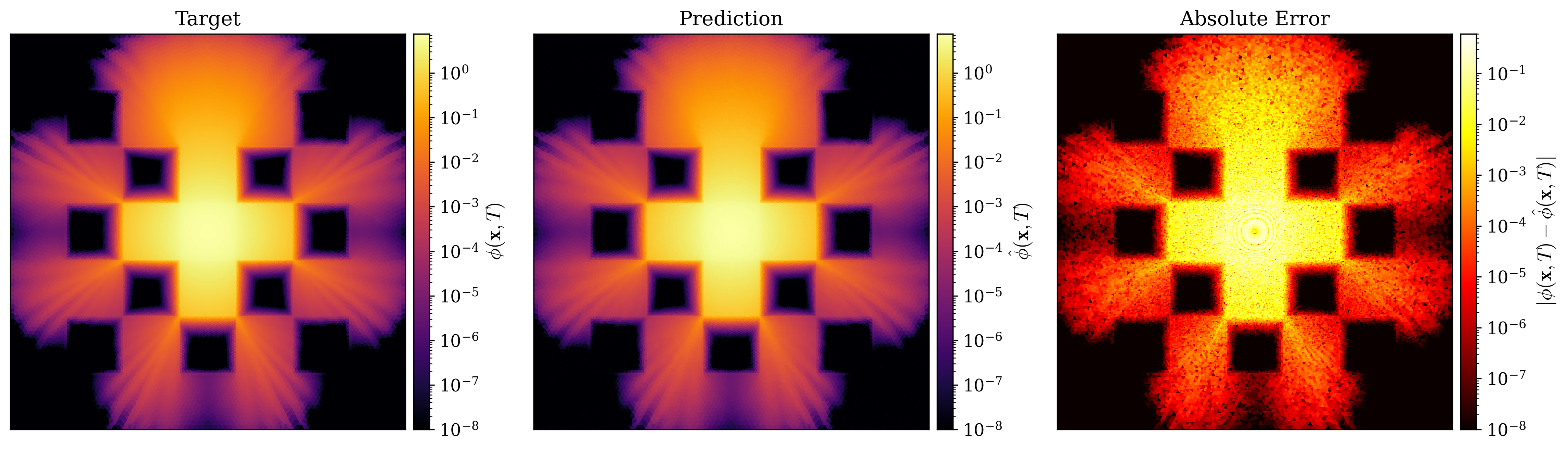}
        \caption{BSMS-MGN.}
        \label{fig:mgn_lattice}
    \end{subfigure}

    \caption{Lattice benchmark, high-resolution evaluation, baseline configuration for both architectures (no Fourier features, no region-weighted loss), representative test sample with design parameters $\sigmaa^B=40.0$, $\sigmas^W=1.1$. Per row: KiT-RT reference solution (left), surrogate prediction (middle), and absolute error (right). The particle-concentration field is shown in log scale. We observe that both surrogates reproduce the qualitative
features of the benchmark, with absolute errors that are visually concentrated along the strongest gradients
between absorbing and streaming subdomains. {The reference is computed with the high-resolution discretization of \Cref{tab:mesh-resolution}: characteristic mesh length $0.015$ ($79{,}862$ cells), $144$ ordinates, and $978$ time steps. The streak-like pattern that radiates from the source through the scattering medium is a ray effect of the discrete-ordinates angular discretization \cite{reference_paper}.}}
    \label{fig:lattice_high_comparison}
\end{figure}

{In \Cref{fig:lattice_high_comparison}, we present the final-time ($T=3.2$) particle concentration $\Phi$ for a representative Lattice test sample (design parameters $\sigmaa^B=40.0$, $\sigmas^W=1.1$), computed at high resolution with KiT-RT (the reference $\Phi_{\textup{HR}}$), Transolver, and BSMS-MGN (the surrogate predictions $\hat{\Phi}$).} Both surrogates reproduce the qualitative features of the benchmark, specifically, source-driven streaming through the white scattering medium and absorption in the blue regions, with absolute errors that are visually concentrated along the strongest gradients between absorbing and streaming subdomains.

\begin{table}[t]
  \centering
  \caption{Lattice benchmark, high-resolution surrogate evaluation (baseline configuration - no Fourier features, no region-weighted loss). Columns report the speedup over the high-resolution KiT-RT solver, per-sample inference time, field RMSE (\Cref{eqn:rmse}), and the relative error in the absorption QoI ($E_{\textup{QoI},\textup{rel}}$, \Cref{eqn:relative_error_pct}) at $t=T$. The high-resolution KiT-RT reference and the cheap low-resolution KiT-RT baseline are shown for comparison. Surrogate rows report mean $\pm$ 1 standard deviation across $n=3$ random seeds. The best surrogate mean per column is shown in bold. Full high- and low-resolution results are given in \Cref{tab:Lattice-eval-full}.}
  \label{tab:Lattice-eval}
  \setlength{\tabcolsep}{4.5pt}
  \begin{tabular}{l|rrrr}
    \toprule
    Method & \makecell{Speedup vs\\high-res \\ KiT-RT}  & Time (s) & RMSE $\Phi$ & $E_{\textup{QoI},\textup{rel}}$ (\%) \\
    \midrule\midrule
    KiT-RT (high-res) & $1.00\times$ & $13.62\pm0.455$ & $-$ & $0.00$ \\
    KiT-RT (low-res) & $59.2\times$ & $0.23\pm0.007$ & $-$ & $6.60$ \\
    \midrule
    Transolver & $197\times$ & $0.069\pm0.0007$ & $\bm{0.0828{\pm}5.20{\times}10^{-3}}$ & $1.06{\pm}0.0620$ \\
    BSMS-MGN & $90.4{\times}$ & $0.1506\pm0.0045$ & $0.0835{\pm}2.90{\times}10^{-5}$ & $\bm{0.276{\pm}0.210}$ \\
    \bottomrule
  \end{tabular}
\end{table}

{In \Cref{tab:Lattice-eval}, we report the field RMSE, the relative absorption-QoI error, and the per-sample timings, averaged over the held-out test samples (\Cref{tab:dataset-distribution}).} The Lattice results use the baseline configuration of each architecture (no Fourier features, no region-weighted loss).

At high-resolution, both surrogates are comparable on field RMSE and surpass the low-resolution KiT-RT finite-volume baseline (\Cref{tab:Lattice-eval}) on the absorption QoI error. BSMS-MGN is the more accurate on the absorption QoI by roughly a factor of four, while Transolver is the more accurate on the field RMSE by a small margin.

\begin{figure}[t]
    \centering

    \begin{subfigure}{1\linewidth}
        \centering
        \includegraphics[width=1\linewidth]{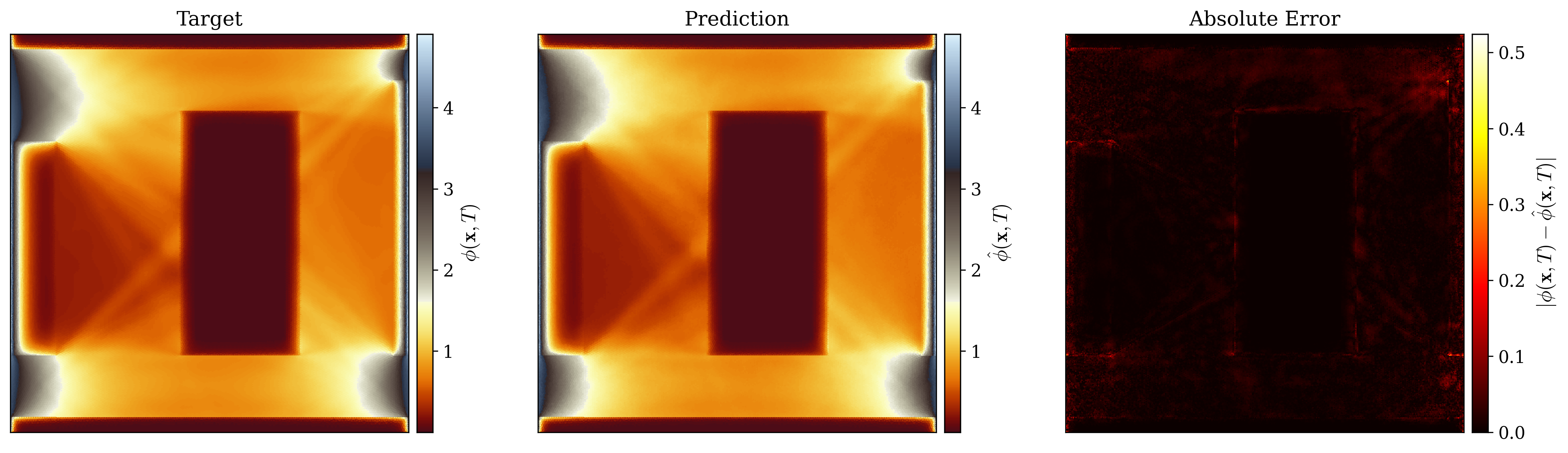}
        \caption{Transolver.}
        \label{fig:transolver_hohlraum}
    \end{subfigure}

    \vspace{0.75em}

    \begin{subfigure}{1\linewidth}
        \centering
        \includegraphics[width=1\linewidth]{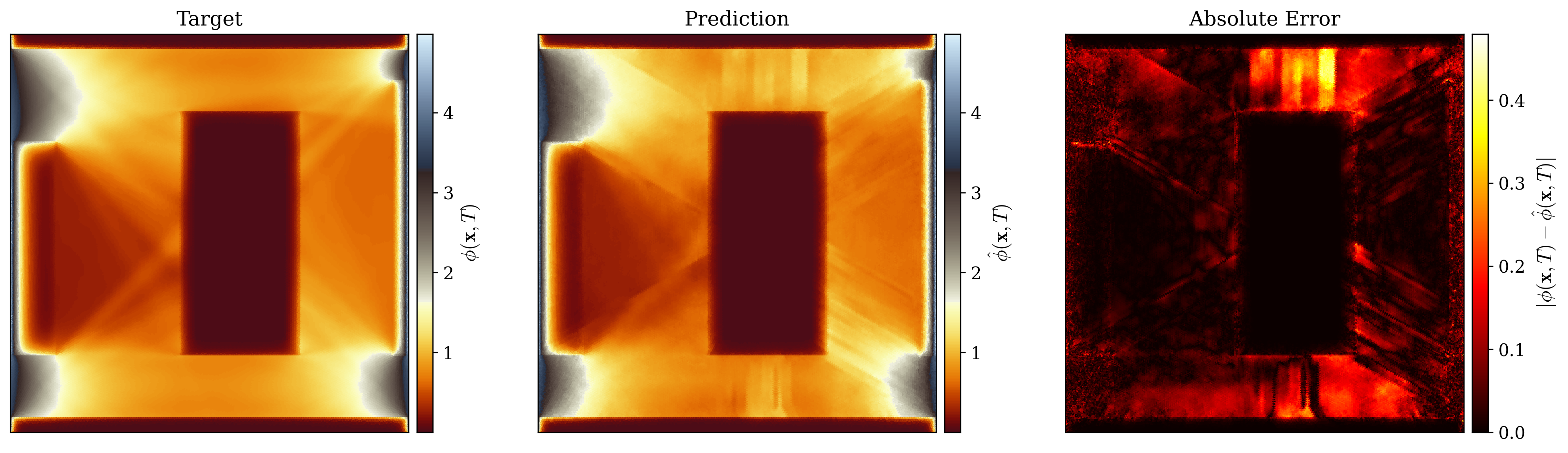}
        \caption{BSMS-MGN.}
        \label{fig:mgn_hohlraum}
    \end{subfigure}

    \caption{Hohlraum benchmark with high-resolution evaluation, full (+FF+RW) configuration for both architectures (Fourier features and region-weighted MSE loss enabled), sample with design parameters $\mathrm{ulr}=0.3$, $\mathrm{llr}=-0.4$, $\mathrm{urr}=0.5$, $\mathrm{lrr}=-0.4$, $\mathrm{hlr}=0.5$, $\mathrm{hrr}=0.6$, $c_x=0.1$, $c_y=0.0$. Per row: KiT-RT reference solution (left), surrogate prediction (middle), and absolute error (right). Both neural surrogates reproduce the main qualitative features of the high-resolution KiT-RT solution,
including the inflow-driven radiation field, attenuation through the red side shields, and absorption around the
central capsule. The Transolver full-configuration prediction is visibly closer to the high-resolution reference,
while the BSMS-MGN full-configuration prediction captures the streaming pattern correctly but exhibits
noticeably larger field errors, especially above and below the central absorbing capsule.}
    \label{fig:hohlraum_high_comparison}
\end{figure}

\subsubsection{Hohlraum}\label{sec_high_res_hohlraum}

{In \Cref{fig:hohlraum_high_comparison}, we present the final-time ($T=2.6$) particle concentration $\Phi$ for a representative Hohlraum test sample (design parameters listed in the caption), computed with KiT-RT (the reference $\Phi_{\textup{HR}}$), Transolver, and BSMS-MGN (the surrogate predictions $\hat{\Phi}$).}
The Hohlraum geometry varies from sample to sample, unlike the fixed Lattice geometry: each design-parameter draw changes the shielding layout and capsule position and therefore requires a separate mesh. Because the surrogates are mesh-agnostic per-cell operators (\Cref{eq_surrogate_model}), each sample is evaluated directly on its own unstructured mesh.

Both neural surrogates reproduce the main qualitative features of the high-resolution KiT-RT solution, including the inflow-driven radiation field, attenuation through the red side shields, and absorption around the central capsule. Of the two surrogates, the Transolver prediction in the full (+FF+RW) configuration is visibly closer to the high-resolution reference, while the BSMS-MGN prediction in the same configuration captures the streaming pattern correctly but exhibits noticeably larger field errors concentrated around the central absorbing capsule.

To localize the source of this accuracy gap between the two architectures on the central capsule, we ablate two engineered design choices per architecture, namely Fourier features (FF) and region-weighted MSE loss (RW), training each configuration from scratch with three random seeds. {In \Cref{tab:Hohlraum-eval}, we report the evaluation of all architecture and feature configuration combinations.}

\paragraph{Transolver.} The high-resolution Transolver achieves the best field RMSE (the RMSE of the full $\Phi$ field, \Cref{eqn:rmse}) among all non-reference methods in the +FF configuration, with relative absorption errors below 1\% on every Hohlraum QoI. Adding Fourier features to the baseline roughly halves the field RMSE and improves every QoI by a factor of roughly $1.4$ to $2.8$. Region weighting alone (+RW) gives only a marginal field-RMSE improvement over the baseline and slightly degrades every QoI. Combining both (+FF+RW) is the second-best configuration on every metric and is uniformly dominated by the +FF configuration. The inference cost is insensitive to whether Fourier features or region weighting are enabled at training time.

\paragraph{BSMS-MGN.} BSMS-MGN behaves differently. The Base and +RW configurations are essentially tied as the best BSMS-MGN configurations on the field RMSE and on the center and vertical QoIs.

Adding Fourier features (+FF or +FF+RW) raises the center-capsule QoI relative error by roughly $3$ to $5$ percentage points. 
On the vertical-shield QoI ($R$), the four configurations are an order of magnitude better than the low-resolution KiT-RT baseline and comparable to Transolver on the same QoI. On the horizontal-shield QoI ($K$), all four configurations have modestly higher error than the low-resolution KiT-RT baseline ($4.66$-$5.12\%$ vs.\ $4.00\%$). On the center-capsule QoI ($G \cup B$), the Base and +RW configurations essentially tie the low-resolution KiT-RT baseline ($10.6\%$/$10.5\%$ vs.\ $10.6\%$), while the +FF-equipped configurations degrade to $13.8$-$15.6\%$. The center QoI is the only target included in the training loss, yet it is the region where BSMS-MGN gains the least over the low-resolution baseline: the cell-dual graph and bi-stride hierarchy capture the vertical streaming and shielding patterns well but resolve the small-scale absorption signature of the central capsule less accurately than Transolver's physics-attention tokens.
The contrasting per-architecture configuration preferences motivate the analysis in \Cref{sec_ablations}.  

Finally, the Hohlraum benchmark provides a test of whether a surrogate trained with one absorption QoI can generalize to other absorption regions. The trained surrogates use only the center absorption QoI as QoI loss \Cref{eq_qoi_loss}, corresponding to $S=G\cup B$ in \Cref{eq_abs_qoi_hohlraum}. The vertical-strip and horizontal-strip shielding QoIs, corresponding to $S=R$ (red, left/right boundaries) and $S=K$ (black, top/bottom boundaries) respectively, are not explicitly included in the loss and therefore measure transfer to physically related but un-optimized integral quantities. We observe (\Cref{tab:Hohlraum-eval}) that these un-optimized QoIs are well below the low-resolution KiT-RT values for high-resolution Transolver (vertical $0.28$-$0.68\%$, horizontal $0.42$-$1.37\%$, vs.\ $8.05\%$ and $4.00\%$), and for high-resolution BSMS-MGN are well below on the vertical QoI ($0.51$-$0.60\%$) and comparable on the horizontal QoI ($4.66$-$5.12\%$ vs.\ $4.00\%$).

 \begin{table}[t]
  \centering
  \caption{Hohlraum benchmark, high-resolution surrogate evaluation across the four ablation configurations (Base, +RW, +FF, +FF+RW; defined in \Cref{tab:ablation-legend}). Columns report the speedup over the high-resolution KiT-RT solver, per-sample inference time, field RMSE (\Cref{eqn:rmse}), and the relative errors in the three per-region absorption QoIs (\Cref{eqn:relative_error_pct}): center $S=G\cup B$, vertical-strip shield $S=R$, and horizontal-strip shield $S=K$. The high-resolution KiT-RT reference and the cheap low-resolution KiT-RT baseline are shown for comparison; surrogate rows report mean and standard deviation across $n=3$ random seeds. The best surrogate mean per column within each architecture is shown in bold. Full high- and low-resolution results are given in \Cref{tab:Hohlraum-eval-full}.}
  \label{tab:Hohlraum-eval}
  \setlength{\tabcolsep}{4pt}
  \resizebox{\linewidth}{!}{%
  \begin{tabular}{l|rrrrrr}
    \toprule
    Method & \makecell{Speedup vs\\high-res \\ KiT-RT} & Time (s) & RMSE $\Phi$ & $E_{\textup{QoI},\textup{rel}}$  Center (\%) & $E_{\textup{QoI},\textup{rel}}$  Vert. (\%) & $E_{\textup{QoI},\textup{rel}}$  Horiz. (\%) \\
    \midrule\midrule
    KiT-RT (high-res) & $1{\times}$ & $35.95 \pm 0.495$ & $-$ & $0.00$ & $0.00$ & $0.00$ \\
    KiT-RT (low-res) & $18.5{\times}$ & $1.94 \pm 0.043$ & $-$ & $10.6$ & $8.05$ & $4.00$ \\
    \midrule
    Transolver Base & $637{\times}$  & $0.05645 \pm 0.0067$ &  $0.0559{\pm}1.20{\times}10^{-3}$ & $0.860{\pm}0.0660$ & $0.551{\pm}0.0420$ & $1.17{\pm}0.190$ \\
    Transolver +RW & $618{\times}$  & $0.05815 \pm 0.0055$ & $0.0517{\pm}2.20{\times}10^{-4}$ & $0.959{\pm}0.0570$ & $0.684{\pm}0.0720$ & $1.37{\pm}0.130$ \\
    Transolver +FF & $639{\times}$  & $0.05625 \pm 0.0085$ & $\bm{0.0244{\pm}2.80{\times}10^{-4}}$ & $\bm{0.624{\pm}0.0380}$ & $\bm{0.280{\pm}0.0130}$ & $\bm{0.422{\pm}5.60{\times}10^{-3}}$ \\
    Transolver +FF+RW & $637{\times}$  & $0.05645 \pm 0.0058$& $0.0252{\pm}3.70{\times}10^{-4}$ & $0.649{\pm}0.0660$ & $0.335{\pm}9.10{\times}10^{-3}$ & $0.672{\pm}0.0450$ \\
    \midrule
    BSMS-MGN Base & $264{\times}$  & $0.13605 \pm 0.0936$ & $0.107{\pm}1.90{\times}10^{-3}$ & $10.6{\pm}0.120$ & $0.602{\pm}0.0960$ & $\bm{4.66{\pm}0.0520}$ \\
    BSMS-MGN +RW & $296{\times}$  & $0.12135 \pm 0.0623$ & $\bm{0.106{\pm}9.20{\times}10^{-4}}$ & $\bm{10.5{\pm}0.0960}$ & $\bm{0.508{\pm}0.0640}$ & $4.83{\pm}0.180$ \\
    BSMS-MGN +FF & $260{\times}$  & $0.13855 \pm 0.1033$ & $0.115{\pm}1.50{\times}10^{-3}$ & $15.6{\pm}0.850$ & $0.549{\pm}0.190$ & $4.81{\pm}0.0550$ \\
    BSMS-MGN +FF+RW &  $270{\times}$  & $0.13295 \pm 0.0959$ & $0.111{\pm}1.80{\times}10^{-3}$ & $13.8{\pm}0.710$ & $0.585{\pm}0.0990$ & $5.12{\pm}0.150$ \\
    \bottomrule
  \end{tabular}
  }
\end{table}

\subsection{Cross-mesh transfer}\label{sec_low_res}

Reusing the same trained checkpoints, we also evaluate each surrogate on the coarser low-resolution mesh (\Cref{tab:mesh-resolution}), a pure cross-mesh transfer with no retraining. Transfer degrades accuracy in both surrogate models. Transolver largely retains its field RMSE but loses accuracy on the absorption QoIs, whereas BSMS-MGN degrades much more severely and produces over-diffused fields. 
The low-resolution surrogate evaluation is therefore competitive only on inference time, not accuracy, so we recommend matching the inference mesh resolution to the training resolution, especially for BSMS-MGN. Transfer learning to a different mesh resolution remains a viable option to increase performance on the low-resolution mesh, however that is not explored in this work. The full high- and low-resolution tables, the corresponding field visualizations, and a detailed discussion are given in \Cref{app:low_res}.

\subsection{Ablations}\label{sec_ablations}
To localize the source of the architecture specific Hohlraum behavior reported in \Cref{tab:Hohlraum-eval}, we ablate two engineered design choices per architecture while holding all other settings of \Cref{tab:transolver-hparams} and \Cref{tab:mgn-hparams} fixed. The first factor is Fourier features (FF). When FF is included, cell-centroid coordinates are lifted to $4K_{\mathrm{ff}}=12$ sinusoidal channels via \Cref{eq_fourier_feature}. When FF is omitted, those channels are removed and the cell-centroid $(x, y)$ coordinates are fed directly into the Transolver or BSMS-MGN node feature embedding. The second factor is the region-weighted MSE loss (RW). When RW is included, we set $w_{\mathrm{void}}=10$ and $w_{\mathrm{mat}}=1$ in \Cref{eq_region_loss}. When RW is omitted, $w_{\mathrm{void}}=w_{\mathrm{mat}}=1$ recovers an unweighted MSE. The QoI loss weight $\lambda$ and all optimizer and schedule settings are left at the values reported in \Cref{tab:mgn-hparams,tab:transolver-hparams}. Toggling the two factors gives the four ablation configurations Base (neither factor), +RW (region weighting only), +FF (Fourier features only), and +FF+RW (both), and each configuration is trained from scratch with three random seeds for both architectures. The resulting multiseed means for each configuration are reported in \Cref{tab:Hohlraum-eval}, and the corresponding field-prediction visualizations are collected in \Cref{app:ablation_visualizations}. The four configurations and their factor settings are summarized in \Cref{tab:ablation-legend}.

\begin{table}[t]
  \centering
  \caption{The four Hohlraum ablation configurations, obtained by toggling two engineered factors: Fourier features (FF) and region-weighted MSE loss (RW). A checkmark denotes that the factor is enabled.}
  \label{tab:ablation-legend}
  \begin{tabular}{lcc}
    \toprule
    Configuration & Fourier features (FF) & Region-weighted loss (RW) \\
    \midrule
    Base   & $-$        & $-$        \\
    +FF    & \checkmark & $-$        \\
    +RW    & $-$        & \checkmark \\
    +FF+RW & \checkmark & \checkmark \\
    \bottomrule
  \end{tabular}
\end{table}

\paragraph{Per-architecture optima are different.}
The ablation reverses the optimal configuration across the two architectures. For Transolver, the +FF configuration wins on every metric (see \Cref{tab:Hohlraum-eval}). Relative to the baseline, the field RMSE more than halves and every per-region QoI drops below one percent. Region weighting alone (+RW) is unhelpful, with each per-region QoI slightly above the baseline. Combining both (+FF+RW) is the second-best Transolver configuration and is uniformly dominated by +FF. This suggests that the physics-attention slice projection benefits from an explicit positional lift to disambiguate cells that share similar material features but lie in spatially distinct streaming regimes, and region weighting is redundant on top of that.

For BSMS-MGN the picture flips. The +RW configuration is the best ablation on every metric besides the horizontal-shield QoI, where the baseline is statistically tied within one standard deviation. Adding Fourier features (+FF or +FF+RW) increases both the field RMSE and the center-capsule QoI values relative to the baseline. The cell-dual graph topology already encodes the spatial structure that Transolver needs Fourier features to recover, so the Fourier lift instead acts as a distractor near the strong absorption signature of the central capsule.

\paragraph{Contrasting inductive-bias preferences.}
Taken across the two architectures, the same two engineered features helps one model and hurts the other. Fourier features are important for Transolver but a net negative for BSMS-MGN on the center-capsule QoI, and region weighting is the mirror image, a mild distractor for Transolver and the only consistently helpful engineered feature for BSMS-MGN. We do not retune the baselines in this paper, but the ablation suggests that the architectural inductive bias (token-based attention versus cell-dual graph message passing with hierarchical pooling) meaningfully changes which auxiliary inputs and loss-weight choices a practitioner should default to. The corresponding field-prediction visualizations for every panel of the ablation are in \Cref{app:ablation_visualizations}.

\begin{table}[t]
  \centering
  \caption{Training cost for the Base configuration of each benchmark and surrogate model on a single DGX H100 node with 8 H100 80\,GB GPUs. Transolver performs global attention over a much smaller latent representation (physics tokens), while BSMS-MGN performs explicit message passing on the cell-dual graph at the finest level and an additional bi-stride U-Net pass over the multi-level hierarchy, leading to longer per-epoch wall times. GPU-hours equal wall time multiplied by the 8 GPUs used per task. Parameter counts and wall times are reported for the Base configuration. The other Hohlraum ablation configurations differ in trainable parameter count by less than $0.1\%$ and in wall time by less than $1\%$ (see \Cref{tab:transolver-hparams,tab:mgn-hparams}).}
  \label{tab:training-cost}
 \begin{tabular}{llrrrr}
    \toprule
    Model & Benchmark & Parameters & Wall time (h) & GPU-hours & s/epoch \\
    \midrule
    BSMS-MGN     & Lattice  & 5,842,669 & $2.06$ & $16.5$ & $14.8$ \\
    BSMS-MGN     & Hohlraum & 5,842,513 & $3.30$ & $26.4$ & $23.7$ \\
    Transolver   & Lattice  & 5,951,105 & $1.57$ & $12.5$ & $11.3$ \\
    Transolver   & Hohlraum & 5,950,593 & $1.86$ & $14.9$ & $13.4$ \\
    \bottomrule
  \end{tabular}
\end{table}

\subsection{Inference efficiency and amortization}\label{sec_efficiency_amortization}
{An AI-based surrogate like transsolver and BSMS-MGN pays off in outer loop applications, e.g., uncertainty quantification, sensitivity analysis, design optimization, and inverse analysis. These applications require thousands to millions of forward evaluations of the same parametrized problem, so that the one-time cost of data generation and training is spread over many queries. We quantify this trade-off below with an amortization model. A second class of applications does not depend on amortization at all: in latency-bound settings such as real-time control, in-line monitoring, or interactive design, what matters is the time to obtain one answer of sufficient accuracy when it is needed, and the up-front cost of building the surrogate is paid off-line, independent of how many queries follow.}

\paragraph{Inference cost.}
Transolver and BSMS-MGN are configured at comparable parameter counts (within roughly two percent of each other (see \Cref{tab:training-cost}), so the per-epoch compute differences between the two architectures are due primarily to differences in architectural structure rather than by capacity. 

BSMS-MGN performs explicit message passing over the cell-dual graph at the finest level for $K_{\mathrm{mp}}=16$ blocks, followed by a bi-stride U-Net pass that visits one GMP block per coarsening level on the down and up legs plus one block at the coarsest level. 
The per-level edge counts form a geometric series, so the U-Net cost is bounded by a constant multiple of the finest-level edge work. Measured per-sample inference wall times are reported in \Cref{tab:Lattice-eval} for Lattice and \Cref{tab:Hohlraum-eval} for Hohlraum, while training wall times and GPU-hours are reported in \Cref{tab:training-cost}.

\begin{figure}
    \centering
    \includegraphics[width=\linewidth]{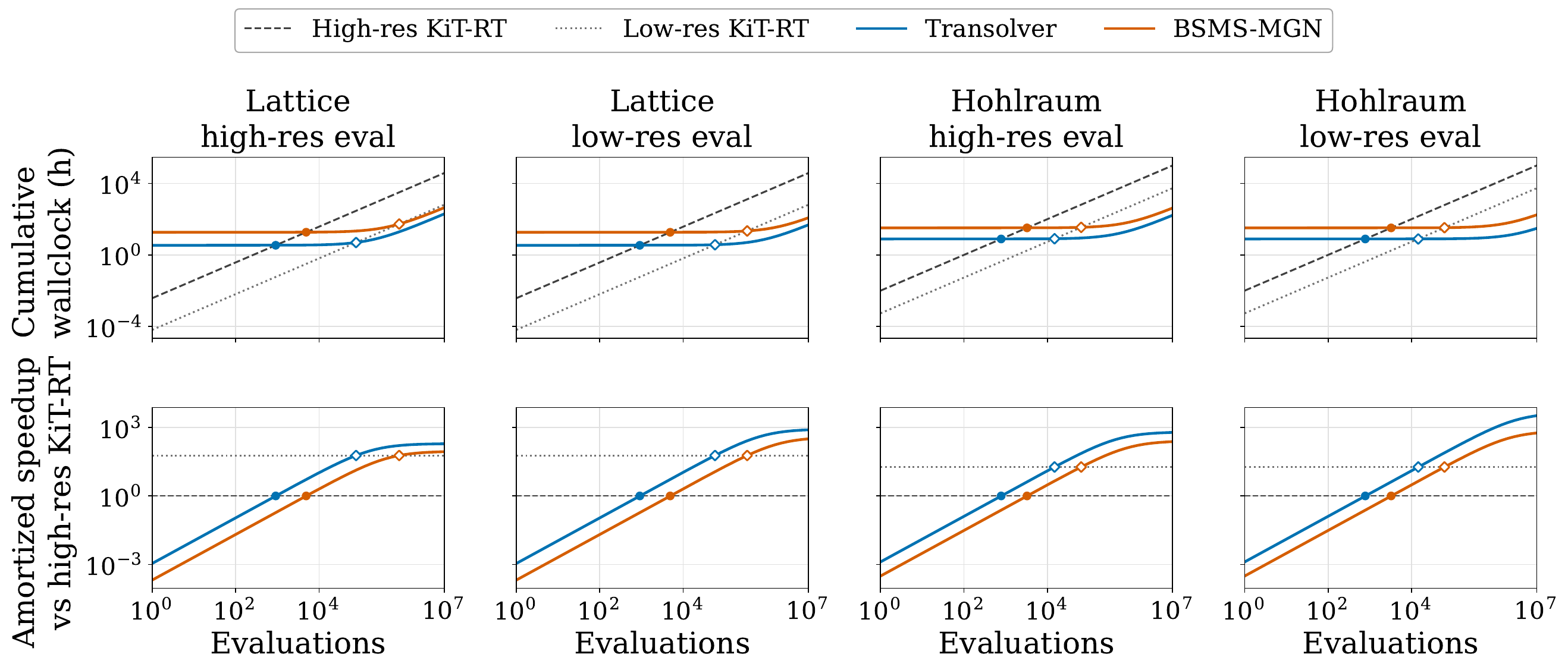}
    \caption{
    Amortization time and amortized speedup of the surrogate models against the high-resolution KiT-RT solver. The Transolver model requires 917 evaluations on Lattice and 781 evaluations on Hohlraum to amortize training and data-generation cost. BSMS-MGN requires approximately 1,050 and 930 evaluations, respectively; the difference to Transolver reflects its longer training and inference times.}
    \label{fig:amortization}
\end{figure}

\paragraph{Amortization model.}
For the high-resolution KiT-RT solver, the total cost for $N$ forward evaluations is
\begin{align}\label{eq_amort_kitrt}
    T_{\mathrm{KiT\mbox{-}RT}}(N) = N\,t_{\mathrm{KiT\mbox{-}RT}},
\end{align}
whereas for a trained surrogate it is
\begin{align}\label{eq_amort_surr}
    T_{\mathrm{surr}}(N) = T_{\mathrm{data}} + T_{\mathrm{train}} + N\,t_{\mathrm{surr}}.
\end{align}
Here $T_{\mathrm{data}}$ is the wall time required to generate the training set, $T_{\mathrm{train}}$ is the {wall-clock} model training {time} (\Cref{tab:training-cost}), and $t_{\mathrm{surr}}$ is the per-sample surrogate inference time (\Cref{tab:Lattice-eval,tab:Hohlraum-eval}). The amortization break-even point is the smallest $N$ at which $T_{\mathrm{surr}}(N) \le T_{\mathrm{KiT\mbox{-}RT}}(N)$. 
We display the amortization time, measured in forward evaluations, and the amortized speedup as functions of the number of forward evaluations in \Cref{fig:amortization}.

\paragraph{Lattice.}
 On the Lattice benchmark, the Transolver surrogate evaluated on the high-resolution mesh amortizes the combined training and data-generation cost after approximately 917 high-resolution KiT-RT evaluations. BSMS-MGN amortizes after approximately {1,050} high-resolution KiT-RT evaluations.

\paragraph{Hohlraum.}

On the Hohlraum benchmark, the Transolver surrogate amortizes the combined training and data-generation cost after approximately 781 high-resolution KiT-RT solves and roughly 15,322 low-resolution KiT-RT solves when evaluated on a high-resolution mesh. BSMS-MGN amortizes after approximately {930} high-resolution KiT-RT solves and {18,400} low-resolution KiT-RT solves. {The amortization counts on Hohlraum are lower than on Lattice because the KiT-RT solves are more expensive per sample ($35.95$ s versus $13.62$ s at high resolution and $1.94$ s versus $0.23$ s at low resolution, \Cref{tab:Lattice-eval,tab:Hohlraum-eval}), mainly because more time steps are needed to reach the final time ($1728$ versus $978$ at high resolution and $662$ versus $176$ at low resolution, \Cref{tab:mesh-resolution}), while the surrogate inference times are similar on both benchmarks.}

\section{Conclusion and future work}\label{sec_conc}
The hypothesis behind this work is that a neural surrogate trained on a few hundred high-resolution $S_N$ solutions can replace the transport solver as the forward model in outer-loop workflows, that is, it reaches the accuracy of the high-resolution solver on engineering quantities of interest at a per-query cost below that of a coarse conventional solve. 

To address this hypothesis, we benchmarked two efficient neural surrogate architectures, the physics-attention Transolver and the multi-scale graph network BSMS-MGN, as end-to-end approximations of the final-time particle concentration for the linear radiation transport equation on the Lattice and Hohlraum benchmarks. At matched parameter counts, Transolver consistently outperforms BSMS-MGN on the Hohlraum center-capsule QoI when Fourier features are enabled, whereas BSMS-MGN benefits from region-weighted training and is competitive on the field-RMSE of the Lattice baseline. The ablation across Fourier features and region weighting reveals strongly architecture-dependent inductive-bias preferences, suggesting that surrogate-model design choices should be revisited per architecture rather than imported across model families. Future work includes extending the surrogate inputs to include the inflow boundary data as part of the per-cell input vector $\bp$, broadening the QoI catalog to include time-resolved and angular quantities of interest, and pushing the same architectures to three-dimensional radiation transport benchmarks.

The prediction performance of the surrogates relative to their wall-time cost---including training and data-generation cost---indicates their usefulness for downstream workflows such as multi-fidelity optimization, uncertainty quantification, design-space exploration, and inverse analysis. These settings often require thousands to millions of forward evaluations, making repeated direct high-resolution transport simulation computationally prohibitive. The results therefore support the use of trained neural surrogates as amortized forward models in outer-loop transport workflows, provided that architecture-specific feature choices and mesh-transfer limitations are accounted for.

\section{Acknowledgements}
This material is based upon work supported by the U.S. Department of Energy, Office of Science, Office of Advanced Scientific Computing Research, as part of their Applied Mathematics Research Program. The work was performed at the Oak Ridge National Laboratory, which is managed by UT-Battelle, LLC under Contract No. De-AC05-00OR22725. The United States Government retains and the publisher, by accepting the article for publication, acknowledges that the United States Government retains a non-exclusive, paid-up, irrevocable, world-wide license to publish or reproduce the published form of this manuscript, or allow others to do so, for the United States Government purposes. The Department of Energy will provide public access to these results of federally sponsored research in accordance with the DOE Public Access Plan (http://energy.gov/downloads/doe-public-access-plan).

\clearpage
\printbibliography
\clearpage 
\input{appendix.tex}

\end{document}

%% file: macros_steffen.tex

\newcommand{\R}{\mathbb{R}}
\newcommand{\N}{\mathbb{N}}
\newcommand{\C}{\mathbb{C}}

\newcommand{\T}{\mathcal{T}}
\newcommand{\Lagr}{\mathcal{L}}

\newcommand{\dimx}{d} 
\newcommand{\dimp}{n_p}

\newcommand{\ds}{\mathrm{d}s}
\newcommand{\dsig}{\mathrm{d}\sigma}
\newcommand{\dsigD}{\mathrm{d}\sigma_D}
\newcommand{\Stwo}{{\mathbf{S}^2}}

\newcommand{\norm}[1]{\left\lVert#1\right\rVert}
\newcommand{\abs}[1]{\left|#1\right|}
\newcommand{\set}[1]{\left\lbrace#1\right\rbrace}
\newcommand{\limes}[2]{\lim\limits_{{#1}\rightarrow {#2}}}
\newcommand{\intD}{\;\mathrm{d}}
\newcommand{\rom}[1]{\uppercase\expandafter{\romannumeral #1\relax}}
\newcommand{\landauO}{\mathcal{O}}

\newcommand{\inner}[1]{\left< #1 \right>}
\newcommand{\red}[1]{\textcolor{red}{#1}}
\newcommand*\interior[1]{#1^{\mathsf{o}}}
\newcommand{\fu}{\mathbf{u}}
\newcommand{\ofu}{\overline{{\mathbf{u}}}}
\newcommand{\ofus}{{\overline{\mathbf{u}}_{\#}}}
\newcommand{\fv}{\mathbf{v}}
\newcommand{\fw}{\mathbf{w}}
\newcommand{\fy}{\mathbf{y}}
\newcommand{\fz}{\mathbf{z}}
\newcommand{\fx}{\mathbf{x}}
\newcommand{\fm}{\mathbf{m}}
\newcommand{\fg}{\mathbf{g}}
\newcommand{\fa}{\mathbf{a}}
\newcommand{\fb}{\mathbf{b}}
\newcommand{\innerInt}[1]{\int_{\mathbb{S}^2} #1 \intD\fv}

\newcommand{\balpha}{{\boldsymbol{\alpha}}}
\newcommand{\bbeta}{{\boldsymbol{\beta}}}
\newcommand{\blue}[1]{\textcolor{blue}{#1}}
\newcommand{\green}[1]{\textcolor{green}{#1}}
\newcommand{\cdh}[1]{\textcolor{red}{#1}}

\newtheorem{theorem}{Theorem}
\newtheorem{lemma}{Lemma}
\newtheorem{corollary}{Corollary}
\newtheorem{definition}{Definition}
\newtheorem{assumption}{Assumption}
\newcommand{\lrW}{W_r}
\newcommand{\W}{W}
\newcommand{\augUc}{\widetilde{U}_c}
\newcommand{\augSc}{\widetilde{S}_c}
\newcommand{\augVc}{\widetilde{V}_c}
\newcommand{\augWrc}{\widetilde{W}_{r,c}}
\newcommand{\augU}{{\widetilde{U}}}
\newcommand{\augS}{{\widetilde{S}}}
\newcommand{\hatS}{{\widehat{S}}}
\newcommand{\hatSc}{{\widehat{S}_c}}
\newcommand{\augV}{{\widetilde{V}}}
\newcommand{\augWr}{{\widetilde{W}_{r}}}
\newcommand{\Wr}{{{W}_{r}}}
\newcommand{\augF}{{\widetilde{F}}}

\def\intstep{{\small \textsf{one-step-integrate}}}
\def\timestep{\eta}

\newcommand{\cgnote}[1]{\textbf{\textcolor[RGB]{50,205,50}{[CG] $\bigstar$ #1 $\bigstar$}}}
\newcommand{\ssboldnKmp}{\ssbold{n}^{K_{\mathrm{mp}}}}   


\newcommand{\scalarFlux}{{\widetilde{F}}}

%% file: sections/intro.tex
\section{Introduction}
The radiation transport equation (RTE) describes the propagation of particles through space as they interact with a surrounding material medium.  The RTE and its variant have applications in in medical imaging \cite{suetens2017fundamentals}, cancer therapies \cite{baskar2012cancer}, and astrophysical flows \cite{mihalis2013foundations}, fission reactors \cite{davison1957neutron,lewis2008fundamentals}, and inertial confinement fusion \cite{hurricane2023physics}.  In all of these examples, the RTE provides a high fidelity kinetic description of particle transport.  However, this fidelity comes at significant cost.  Indeed, the RTE solution may be a function of (as many as) six phase-space dimensions (three space and three momentum) and may also depend on time. 
Thus, standard grid-based discretization methods require a great deal of memory and a large number of floating point operations to obtain an accurate solution, while Monte Carlo approaches are known to suffer from noise \cite{dupree2012monte,lewis1984computational}.  
Second, the interactions of radiation with the material can be extremely complex \cite{mihalas1984radiation}, and variations in collision rates, due to multiple materials or strong dependencies on energy, lead to solution that are supported across a wide range of spatial and temporal scales \cite{bensoussan1979boundary, larsen2023asymptotic}.  Resolving all of these scales increases the computing requirements even further.

There are two basic approaches to reduce the computational expense of simulating RTEs. The first  approach is to use surrogate models, including diffusion equations \cite{levermore1981flux, habetler1975uniform,larsen1980diffusion}, various type of moments models \cite{olson2000diffusion,hauck2011high, frank2006partial, dubroca1999etude,brunner2005two} and reduced spectral approximations \cite{mcclarren2010theoretical,hamilton2015efficient, gelbard1960application}.  The second approach is to make high-resolution phase space simulations more efficient.  
Two of the major advances in this area include  (i) asymptotic preserving methods \cite{ larsen1989asymptotic,larsen1987asymptotic,jin2010asymptotic} that can compute accurate solutions to collision-dominated flows with under-resolved meshes and (ii) sophisticated iterative schemes that enable implicit time integration with large time steps \cite{warsa2004krylov,adams2002fast,chacon2017multiscale,gol1964quasi,yamamoto2005generalized}.  

In the current work, we consider surrogates that are trained using data generated from high-resolution simulations of a linear RTE. Our approach is motivated by the recent advances in operator learning methods \cite{kovachki2023neural,kovachki2024operator, boulle2024mathematical,subedi2025operator} that learns use neural-network architectures to generate map between functions spaces, in order to approximation solutions of partial differential equations.  We consider two such surrogates.  The first, called the Transolver \cite{wu2024transolver}, is based on a transformer architecture.  The second, called Bi-Stride Multi-Scale MeshGraphNet (BSMS-MGN) \cite{cao2022bi}, uses a hierarchical multi-scale graph neural network architecture. These models are trained and tested using the NVIDIA's PhysicsNeMo, an open-source PyTorch framework for scientific machine learning \cite{PhysicsNeMo_Contributors_NVIDIA_PhysicsNeMo_An_2023}.

We benchmark the approximation quality and compare the performance of both surrogate models against a discrete-ordinates/finite-volume method on the well-established benchmark problems \cite{reference_paper}.  The first benchmark problem, referred to as the Lattice benchmark, is a toy mock-up of a nuclear reactor assembly and is taken directly from \cite{brunner2002forms}.  The second problem is a linearized, symmetric Hohlraum benchmark and is a variation on the original nonlinear Hohlraum problem first presented in \cite{brunner2002forms}.  Both test problems are equipped with design parameters and quantities of interest.  

We construct training data for both benchmarks using the highly parallel  \texttt{C++}-based, GPU-accelerated solver KiT-RT~\cite{kusch2023kit} using a discrete-ordinates ($S_N$) discretization in angular space, a finite-volume method on an unstructured grid for physical space, and an explicit SSP Runge-Kutta time stepping scheme.  
Using high-resolution training data, we generate reproducible surrogate models for each benchmark that map the physical properties of the background material to the particle concentration at a fixed final time.  We investigate the ability of each surrogate to recover high-resolution solutions at previously unseen material inputs, using a course-solution of the KiT-RT solver as a baseline for understanding accuracy vs. cost tradeoffs.  We find that in almost all cases the surrogates outperform the course simulation results. We also perform an amortization analysis to quantify the number of input samples are needed in to recover surrogate training and data-generation costs in outer-loop workflows.

The remainder of this paper is organized as follows.
In \Cref{sec_rad_transport}, we present the radiation transport equation. In \Cref{sec_problems},  we provide a brief recap of the two benchmark problems:  Lattice and Hohlraum. In \Cref{sec_numerics}, we present the discretization scheme used to compute training data for these problems, together with the solver settings.
In \Cref{sec_ai} we describe the details of both surrogate models. 
In \Cref{sec_results}, we present the numerical results. In \Cref{sec_conc}, we draw conclusions and provide discussion.  Several appendices are included to fix notation \Cref{app:notation} , provide extended details of the surrogate training and implementation \Cref{app:hyperparams}, to show visualizations of some of the ablation studies for the Hohlruam \Cref{app:ablation_visualizations}, and to show performance of the surrogates when evaluated directly on a coarse mesh \Cref{app:low_res}. 

All code and data are used to generate results are publicly available. In particular, the model training and inference code can be found at \url{https://github.com/NVIDIA/physicsnemo/tree/main/examples/nuclear_engineering/radiation_transport}, the code for the reference simulation and data generation can be found at \url{https://github.com/KiT-RT} and the pre-generated dataset can be found at \url{https://huggingface.co/datasets/nvidia/Linear-Radiation-Transport}.

%% file: sections/rte.tex
\section{The radiation transport equation}\label{sec_rad_transport}
We briefly present the  radiation transport equation (RTE). To reduce simulation costs, the benchmarks below are assumed to be extrusions of two-dimensional geometries into a third direction.  This allows the spatial domain of the usual RTE to be reduced from three dimensions to two.  Details of this reduction can be found in \cite{reference_paper}; see also \cite{lewis1984computational}.

Let $X\subset \mathbb{R}^2$ be a domain with Lipschitz boundary $\p X$, and let $\bbS^2$ be the unit sphere in $\bbR^3$. Let 
\begin{align}\label{eq_3d_boundary}
    \Gamma^{\pm} = \{ (\bx,\bsOmega) \in \p X \times \bbS^2 \colon \pm \bn(\bx) \cdot \bsOmega > 0 \}
\end{align}
be the inflow (-) and outflow boundaries (+), where $\bn(\bx) \in  \bbR^2$ is the outward unit normal vector at $\bx \in X$, defined for almost every $\bx \in \p X$. The spatial variable is parametrized in Cartesian coordinates, i.e., $\bx=(x,y)$,  and the angular variable in polar coordinates, i.e., $\bsOmega=(\xi,\eta)$ with  $\xi = \sqrt{1-\mu^2} \cos \varphi$  and  $\eta = \sqrt{1-\mu^2} \sin \varphi$  where {{$(\mu,\varphi) \in [-1,1] \times [0,2 \pi)$}}. 

Let $\Psi(\bx,\bsOmega,t)$ be the density of particles with respect to the phase space measure $d \bsOmega d \bx$ at position $\bx\in X$ and traveling in the direction  $\bsOmega$ at time $t \geq 0$.  We consider particles of unit speed that interact with a static material via scattering and absorption, at rates characterized by cross-sections $\sigmas=\sigmas(\bx)$ and $\sigmaa = \sigmaa(\bx)$, respectively. In particular, the scattering is assumed to be isotropic, i.e., independent of $\bsOmega$. The material also generates a known isotropic particle source $Q=Q(\bx,t)$. Under these assumptions, $\Psi$ is governed by an RTE of the form
\begin{align}\label{eq_RTE}
\begin{cases}
  	\p_t \Psi+\bsOmega \cdot \nabla_{\bx}\Psi +\sigmat \Psi 
        =  \frac{\sigmas}{4\pi} \Phi +Q,  
        &\quad (\bx , \bsOmega) \in X \times {\mathbb{S}^2}, \;  t >0,  
        \\
    \Psi(\bx,\bsOmega,t) =  \Psi^-(\bx,\bsOmega,t) 
    & \quad   (\bx , \bsOmega) \in \Gamma^- , \; t>0, \\
    \Psi(\bx,\bsOmega,0) =  \Psi_0(\bx,\bsOmega) 
    & \quad (\bx , \bsOmega) \in X \times {\mathbb{S}^2},
 \end{cases}
\end{align}
where $\sigmat = \sigmas + \sigmaa$ and 
  \begin{align}     \label{eq_scalar_flux}
    \Phi(\bx,t) =\int_{\bbS^2}\Psi(\bx,\bsOmega,t) d \bsOmega.
    \end{align}
is the particle concentration, i.e., the density of particles at position $\bx$, with respect to the measure $d \bx$, at $t$.  The initial condition $\Psi_0(\bx,\bsOmega)$ and inflow boundary data $\Psi^{-}  = \Psi^-(\bx,\bsOmega,t)$ are given. 

%% file: appendix.tex
\appendix

\section{Notation}\label{app:notation}

\begin{table}[htbp]
\centering
\caption{ Notation for the continuous radiation transport equation.}
\label{tab:notation-rte}
\renewcommand{\arraystretch}{1.15}
\begin{tabularx}{\textwidth}{p{0.22\textwidth} X}
\toprule
Symbol & Meaning \\
\midrule
$X$ & Physical material domain in two spatial dimensions. \\
$\p X$ & Boundary of the physical domain. \\
$\bx=(x,y)$ & Spatial coordinate. \\
$\bn(\bx)$ & Outward unit normal on $\p X$. \\
$\mathbb{S}^2$ & Unit sphere in angular space. \\
$\mathbb{S}_+^2$ & Upper half of the unit sphere in angular space. \\
$\bsOmega$ & Angular direction. \\
$\Gamma^\pm$ & Inflow and outflow phase-space boundaries. \\
$\Psi(\bx,\bsOmega,t)$ & Angular particle density in the continuous RTE. \\
$\Psi^-(\bx,\bsOmega,t)$ & Prescribed inflow boundary data on $\Gamma^-$. \\
$\Psi_0(\bx,\bsOmega)$ & Initial angular particle density. \\
$\sigmas(\bx)$ & Scattering cross-section. \\
$\sigmaa(\bx)$ & Absorption cross-section. \\
$\sigmat(\bx)$ & Total cross-section, $\sigmat=\sigmas+\sigmaa$. \\
$\sigmas^{W},\,\sigmaa^{B},\ldots$ & Cross section restricted to a subdomain, indicated by the superscript ($W$ white, $B$ blue, $R$ red, $K$ black, $G$ green). \\
$Q(\bx,t)$ & Isotropic material source (time-independent in both benchmarks). \\
$\Phi(\bx,t)$ & Particle concentration, or scalar flux, obtained by angular integration \eqref{eq_scalar_flux}. \\
$T$ & Final simulation time. \\
\bottomrule
\end{tabularx}
\end{table}

\begin{table}[htbp]
\centering
\caption{Notation for the finite-volume and discrete-ordinates reference solver.}
\label{tab:notation-fv-sn}
\renewcommand{\arraystretch}{1.15}
\begin{tabularx}{\textwidth}{p{0.22\textwidth} X}
\toprule
Symbol & Meaning \\
\midrule
$h$ & Maximum mesh-cell diameter. \\
$N_h$ & Number of finite-volume mesh cells. \\
$i$ & Mesh-cell index. \\
$\bx_i$ & Centroid of mesh cell $i$. \\
$\Phi_i$ & Reference particle concentration at cell $i$ and final time. \\
$N_q$ & Tessellation quadrature order. \\
$4N_q^2$ & Number of angular quadrature points for tessellation quadrature order $N_q$. \\
$\mu,\varphi$ & Polar angular variables. \\
$\xi,\eta$ & Horizontal and vertical angular-velocity components. \\
$K_i$, $|K_i|$ & Mesh cell $i$ and its area. \\
$\Delta t$ & Time step of the explicit SSP-RK2 scheme. \\
$\mathrm{CFL}$ & Courant--Friedrichs--Lewy number used for explicit time stepping {, defined through $\Delta t=\mathrm{CFL}\cdot\min_i\sqrt{|K_i|}$ with unit particle speed (\Cref{sec_numerics})}. \\
\bottomrule
\end{tabularx}
\end{table}

\begin{table}[htbp]
\centering
\caption{Notation for the generic surrogate-model formulation and training loss.}
\label{tab:notation-surrogates}
\renewcommand{\arraystretch}{1.15}
\begin{tabularx}{\textwidth}{p{0.22\textwidth} X}
\toprule
Symbol & Meaning \\
\midrule
$\mathcal{F}_\theta$ & Learned surrogate operator mapping cell-wise inputs to final-time particle concentration. \\
$\theta$ & Trainable model parameters. \\
$\bar{\bx}_i$ & Normalized cell-centroid coordinate used as a model input. \\
$\bp_i$ & Physical input vector for spatial cell $i$, $\bp_i=(\sigma_{\textup{a},i},\sigma_{\textup{s},i},\sigma_{\textup{t},i},Q_i)$.  $Q_i$ is excluded when the source is everywhere zero. \\
$\bff_i$ & Fourier features of cell $i$, $\bff_i=\bff(\bar{\bx}_i)$, \eqref{eq_fourier_feature}; optional input, ablation factor FF. \\
$\bff(\bar{\bx})$ & Fourier feature map \eqref{eq_fourier_feature}, applied to a normalized coordinate $\bar{\bx}\in[-1,1]^2$.  \\
$K_{\mathrm{ff}}$ & Number of Fourier frequencies, $K_{\mathrm{ff}}=3$. \\
$f_0$ & Base Fourier frequency, $f_0=2$ (\Cref{sec_shared_surrogate_setup}). \\
$f_k$ & Fourier frequency $f_k=f_0 2^k$, $k=0,\ldots,K_{\mathrm{ff}}-1$, i.e., $f_k\in\{2,4,8\}$. \\
$\hat{\Phi}_i$ & Surrogate-predicted particle concentration at cell $i$ and final time. \\
$\mathcal{L}(\theta)$ & Total training loss. \\
$\mathcal{L}_{\mathrm{region}}$ & Region-weighted particle-concentration reconstruction loss. \\
$\mathcal{L}_{\mathrm{QoI}}$ & Loss term penalizing relative absorption-QoI error. \\
$\lambda$ & Weight on the QoI loss. \\
$w_i$ & Per-cell region weight in $\mathcal{L}_{\mathrm{region}}$. \\
$W$ & Sum of region weights, $W=\sum_i w_i$. \\
$\epsilon$ & Small positive constant for numerical stability. \\
$\mathrm{RMSE}\,\Phi$ & Root-mean-square error in the final-time particle-concentration field  (\Cref{eqn:rmse}). \\
$E_{\textup{QoI},\textup{rel}}$ & Relative absorption-QoI error in percent. \\
\bottomrule
\end{tabularx}
\end{table}

\begin{table}[htbp]
\centering
\caption{Notation for the Transolver architecture. {Multi-indices are comma-separated ($i$ cell, $a$ head, $j$ token, $\ell$ layer) and every description names the indices it depends on.}}
\label{tab:notation-transolver}
\renewcommand{\arraystretch}{1.15}
\begin{tabularx}{\textwidth}{p{0.22\textwidth} X}
\toprule
Symbol & Meaning  \\
\midrule
$E_{\mathrm{in}}$ & Pointwise input embedding from raw cell features to hidden states. \\
$E_{\mathrm{out}}$ & Pointwise output embedding from hidden state to particle-concentration prediction. \\
$\bh_i^\ell$ & Hidden feature of cell $i$ after Transolver layer $\ell$. \\
$\ssbold{H}^\ell$ & Matrix of the $N_h$ hidden features after layer $\ell$, $\ssbold{H}^\ell=[\bh_1^\ell,\ldots,\bh_{N_h}^\ell]^T$. \\
$L$ & Number of Transolver layers. \\
$C$ & Hidden channel dimension. \\
$H_{\mathrm{attn}}$ & Number of attention heads. \\
$a$ & Attention-head index. \\
$M$ & Number of physics tokens per attention head. \\
$j$ & Physics-token index. \\
$d_h$ & Per-head feature dimension. \\
$\ssbold{u}_i^\ell$ & Layer-normalized state of cell $i$ entering layer $\ell$, used to form slice assignments and value features. \\
$P_{s,a}^{\ell}$ & Projection of head $a$ in layer $\ell$ that forms the slice-assignment features. \\
$P_{v,a}^{\ell}$ & Value projection of head $a$ in layer $\ell$. \\
$P_{\mathrm{slice},a}^{\ell}$ & Projection of head $a$ in layer $\ell$ from slice-assignment features to the $M$ slice logits. \\
$\tau_a^\ell$ & Learnable temperature for slice assignment in head $a$ and layer $\ell$. \\
$\ssbold{s}_{i,a}^{\ell}$ & Slice-assignment feature of cell $i$ for head $a$ in layer $\ell$. \\
$\bv_{i,a}^{\ell}$ & Value feature of cell $i$ for head $a$ in layer $\ell$. \\
$w_{i,a,j}^{\ell}$ & Soft assignment weight of cell $i$ to slice $j$ in head $a$ and layer $\ell$. \\
$\ssbold{z}_{a,j}^{\ell}$ & Physics token for head $a$, token index $j$, and layer $\ell$. \\
$\ssbold{Z}_a^\ell$ & Matrix of the $M$ physics tokens of head $a$ in layer $\ell$. \\
$\widetilde{\ssbold{Z}}_a^\ell$ & Token matrix of head $a$ in layer $\ell$ after self-attention over the tokens. \\
$\widetilde{\ssbold{r}}_{i,a}^{\ell}$ & Dispatched representation of cell $i$ from head $a$ in layer $\ell$, the combination of the updated tokens weighted by $w_{i,a,j}^{\ell}$. \\
$\operatorname{SA}_a$ & Self-attention operator applied to the $M$ physics tokens of head $a$. \\
$\operatorname{PhysAttn}$ & Physics-Attention operator. \\
$\mathcal{T}_{\ell}$ & The $\ell$-th Transolver layer. \\
\bottomrule
\end{tabularx}
\end{table}

 
\begin{table}[htbp]
\centering
\caption{Notation for the Bi-Stride Multi-Scale MeshGraphNet (BSMS-MGN) architecture.}
\label{tab:notation-bsms-mgn}
\renewcommand{\arraystretch}{1.15}
\begin{tabularx}{\textwidth}{p{0.22\textwidth} X}
\toprule
Symbol & Meaning \\
\midrule
$l$ & Mesh-level index in the bi-stride hierarchy, $l \in \{0,\ldots,L_g\}$, with $l=0$ the finest and $l=L_g$ the coarsest. \\
$L_g$ & Number of bi-stride coarsening levels, set to $L_g=5$ in the final models (distinct from the Transolver layer count $L$). \\
$\mathcal{G}^{(l)}=(\mathcal{V}^{(l)},\mathcal{E}^{(l)})$ & Directed graph at level $l$ of the bi-stride hierarchy; $\mathcal{G}^{(0)}$ is the cell-dual graph of the finite-volume mesh. \\
$\mathcal{V}^{(l)}$ & Set of graph nodes at level $l$. \\
$\mathcal{E}^{(l)}$ & Set of directed graph edges at level $l$. \\
$(i,j)$ & Directed edge from source node $i$ to destination node $j$. \\
$\mathcal{I}^{(l)} \subset \mathcal{V}^{(l)}$ & Node indices kept by bi-stride pooling when going from level $l$ to level $l+1$. \\
$\bx_{l,i}$ & Cell-centroid position of node $i$ at level $l$; for $l\ge 1$ obtained from the finer positions with the contribution table $C^{(l-1)}$ of \eqref{eq_bsms_contribution}, see \eqref{eq_bsms_down}. \\
$\Delta\bx_{l,ij}$ & Cell-centroid positional offset $\bx_{l,i}-\bx_{l,j}$ on level $l$. \\
$w_i^{(l)}$, $\widehat{w}_{ij}^{(l)}$ & Non-parametric (i.e. non-trainable) pooling weights used to form the BSMS contribution table, initialized with $w_i^{(0)}=1$. \\
$C_{ij}^{(l)}$ & Contribution-table entry for edge $(i,j)$ at level $l$. \\
$\widetilde{\ssbold{n}}_i$ & Raw input node feature at node $i$ on the finest level, before node encoding. \\
$\bfe_{ij}^{0}$ & Raw input edge feature for directed edge $(i,j)$ on the finest level. \\
$\ssbold{n}_i^k$ & Latent node state after finest-level message-passing block $k$. \\
$\ssbold{g}_{ij}^k$ & Latent edge state after finest-level message-passing block $k$; this edge latent is only propagated at level $0$. \\
$\ssbold{a}_i^{k+1}$ & Aggregated incoming message at node $i$ after the finest-level edge update in block $k$. \\
$\bv_i^{(l)}$ & Latent node state of node $i$ on level $l$ during the U-Net pass; on level $0$ it is initialized with $\ssbold{n}_i^{K_{\mathrm{mp}}}$ and overwritten by the down block and again by the up block. \\
$\ssboldnKmp$ & All finest-level node states after the $K_{\mathrm{mp}}$ message-passing blocks, the input of the U-Net pass $\mathcal{U}_{L_g}$. \\
$\bfe_{l,ij}$ & On-the-fly edge message at level $l$ inside the U-Net GMP block. \\
$C_g$ & Shared latent dimension for encoder, finest-level processor, and U-Net GMP blocks. \\
$K_{\mathrm{mp}}$ & Number of finest-level MeshGraphNet message-passing blocks. \\
$\mathrm{Enc}_v,\,\mathrm{Enc}_e$ & Node and edge encoder MLPs at the finest level. \\
$\mathrm{MLP}_e^{k},\,\mathrm{MLP}_v^{k}$ & Edge- and node-update MLPs in finest-level message-passing block $k$, with disjoint weights across blocks. \\
$\mathrm{MLP}_e^{(l)},\,\mathrm{MLP}_v^{(l)}$ & Edge- and node-update MLPs in the level-$l$ U-Net GMP block, with disjoint weights across mesh levels and across the down, up, and bottom passes. \\
$\makecell{
\mathrm{GMP}^{(l)}_{\mathrm{down}},\,\mathrm{GMP}^{(l)}_{\mathrm{up}},\\
\mathrm{GMP}^{(L_g)}_{\mathrm{bottom}}
}$ & Per-level GMP blocks of the bi-stride U-Net pass; each contains its own edge- and node-update MLP pair. \\
$\mathrm{Dec}$ & Node decoder mapping the final latent node state to particle concentration. \\
$\mathcal{P}_k$ & The $k$-th finest-level MeshGraphNet message-passing block. \\
$\mathcal{U}_{L_g}$ & Bi-stride U-Net pass with $L_g$ coarsening levels, composed of down, bottom, and up passes through the GMP blocks and non-parametric contribution-table transitions. \\
$d_v$ & Raw finest-level node-feature dimension. \\
$d_e$ & Raw finest-level edge-feature dimension. \\
\bottomrule
\end{tabularx}
\end{table}

\clearpage
\clearpage
\section{Model Details and Hyperparameters}\label{app:hyperparams}

\subsection{Definition Transolver}\label{sec_transolver}
Transolver \cite{wu2024transolver} instantiates the surrogate operator
$\mathcal{F}_\theta$ in \eqref{eq_surrogate_model} as a transformer that
communicates through a small set of learned, physics-aware tokens rather than
through direct all-to-all attention over the $N_h$ mesh cells.

For each cell $i$, the raw input is assembled from the normalized cell centroid coordinates $\bar{\bx}_i$, the auxiliary features $\bff_i$, and the per-cell material/source
quantities
\begin{equation}
    \bp_i=(\sigma_{\textup{a},i},\sigma_{\textup{s},i},\sigma_{\textup{t},i},Q_i).
\end{equation}
The source term
$Q_i$ is omitted for Hohlraum, where $Q\equiv 0$.

The Fourier features $\bff_i$ are the engineered input
factor FF of the Hohlraum ablation (the other factor, RW,
is the region-weighted loss). The main-text ablation results are collected
in \Cref{sec_ablations} (\Cref{tab:Hohlraum-eval}), with the
corresponding Transolver field-prediction visualizations across all four
configurations in \Cref{sec_transolver_ablation_grid}.
The full per-tensor layout is tabulated in \Cref{tab:transolver_inputs}.
A pointwise input embedding maps these cell-wise quantities to the hidden
Transolver representation,
\begin{align}
    \bh_i^0
     =
    E_{\mathrm{in}}(\bar{\bx}_i,\bp_i,\bff_i)
    \in \mathbb{R}^{C}, \qquad
    \ssbold{H}^0
    =
    [\bh_1^0,\ldots,\bh_{N_h}^0]^T
    \in \mathbb{R}^{N_h\times C}.
\end{align}

\paragraph{Physics-Attention.}
The Physics-Attention block uses multi-head self-attention \cite{vaswani2017attention} together with a downsampling strategy that decouples the cost of the attention operator from the grid resolution.
For head $a=1,\dots,H_{\mathrm{attn}}$ in layer $\ell$, the Physics-Attention block first normalizes the
cell states,
$
    \ssbold{u}_i^\ell
    =
    \operatorname{LayerNorm}(\bh_i^{\ell-1})
    \in \mathbb{R}^{C}.
$
The normalized state is then projected to a head-specific feature space. We
write with $P_{s,a}^{\ell},  P_{v,a}^{\ell}\in\mathbb{R}^{d_h\times C}$
\begin{align}
    \ssbold{s}_{i,a}^{\ell}
    =
    P_{s,a}^{\ell}\ssbold{u}_i^\ell
    \in \mathbb{R}^{d_h}, \qquad
    \bv_{i,a}^{\ell}
    =
    P_{v,a}^{\ell}\ssbold{u}_i^\ell
    \in \mathbb{R}^{d_h},
\end{align}
where $\ssbold{s}_{i,a}^{\ell}$ is used to compute slice-assignment weights and
$\bv_{i,a}^{\ell}$ is the value feature to be aggregated into slice tokens.
The soft assignment of cell $i$ to the $M$ learned slices is
\begin{align}
    \ssbold{w}_{i,a}^{\ell}
    &=
    \operatorname{softmax}
    \left(
        P_{\mathrm{slice},a}^{\ell}\ssbold{s}_{i,a}^{\ell}/\tau_a^\ell
    \right)
    \in \mathbb{R}^{M},
    \qquad i=1,\ldots,N_h,
    \label{eq_physics_tokens}
\end{align}
where
$P_{\mathrm{slice},a}^{\ell}\in\mathbb{R}^{M\times d_h}$ and
$\tau_a^\ell>0$ is a learnable temperature parameter. The weights satisfy
$\sum_{j=1}^{M} w_{i,a,j}^{\ell}=1$.
The $j$-th physics-aware token is the normalized weighted average of the value
features assigned to that slice,
\begin{align}
    \ssbold{z}_{a,j}^{\ell}
    =
    \frac{
        \sum_{i=1}^{N_h} w_{i,a,j}^{\ell}\bv_{i,a}^{\ell}
    }{
        \sum_{i=1}^{N_h} w_{i,a,j}^{\ell}+\epsilon
    },
    \qquad
    j=1,\ldots,M .
\end{align}
Thus,
\[
    \ssbold{Z}_a^\ell
    =
    [\ssbold{z}_{a,1}^{\ell},\ldots,\ssbold{z}_{a,M}^{\ell}]^T
    \in \mathbb{R}^{M\times d_h}.
\]
The learned slices are feature-dependent rather than fixed geometric
partitions, so spatially separated cells may contribute to the same token if their
learned states are similar.

Standard multi-head self-attention  \cite{vaswani2017attention} is applied over the $M$ physics-aware tokens rather than over the
$N_h$ mesh cells, i.e., 
\begin{align}
    \widetilde{\ssbold{Z}}_a^\ell = \operatorname{SA}_a( \ssbold{Z}_a^\ell)   \in \mathbb{R}^{M\times d_h},
\end{align}
where {$\operatorname{SA}_a$ denotes the standard self-attention operator of \cite{vaswani2017attention} applied to the token matrix $\ssbold{Z}_a^\ell$ of head $a$; the subscript indicates which head's tokens are attended over. In the reference implementation of \cite{wu2024transolver}, the query, key, and value projections of this token attention are single $d_h\times d_h$ matrices shared across heads, so the subscript does not denote head-specific weights.}
The updated token information is then dispatched back to mesh cells using the
same slice weights,
\begin{align}
    \widetilde{\ssbold{r}}_{i,a}^{\ell}
    =
    \sum_{j=1}^{M}
    w_{i,a,j}^{\ell}\,\widetilde{\ssbold{z}}_{a,j}^{\ell}
    \in \mathbb{R}^{d_h}, \qquad i=1,\dots,N_h.
\end{align}
The outputs from all heads are concatenated and mapped back to
$\mathbb{R}^{C}$ by a learned linear projection. This defines the
Physics-Attention operator $\operatorname{PhysAttn}(\cdot)$.

Thus, each Transolver layer follows the standard pre-norm transformer residual
structure,
\begin{equation}
\begin{split}
\label{eq:H-layer-l-update}
    \widehat{\ssbold{H}}^{\ell}
    &=
    \ssbold{H}^{\ell-1}
    +
    \operatorname{PhysAttn}
    \left(
        \operatorname{LayerNorm}(\ssbold{H}^{\ell-1})
    \right), \\
    \ssbold{H}^{\ell}
    &=
    \widehat{\ssbold{H}}^{\ell}
    +
    \operatorname{MLP}
    \left(
        \operatorname{LayerNorm}(\widehat{\ssbold{H}}^{\ell})
    \right),
    \qquad
    \ell=1,\ldots,L ,
    \end{split}
\end{equation}
using the standard layer normalization definition
\cite{ba2016layernormalization}.
After the final Transolver layer, a pointwise output embedding maps each hidden cell state to the predicted particle concentration at final time,
\begin{align}
    \hat{\Phi}_i
    =
    E_{\mathrm{out}}(\bh_i^L)
    \in \mathbb{R}.
\end{align}

The Transolver surrogate factors the operator in \eqref{eq_surrogate_model} as 
\begin{align}
    \mathcal{F}_\theta = E_{\mathrm{out}} \circ \mathcal{T}_L \circ \cdots \circ \mathcal{T}_1 \circ E_{\mathrm{in}},
\end{align}
where $\mathcal{T}_\ell : \ssbold{H}^{\ell-1} \mapsto \ssbold{H}^{\ell}$ denotes the $\ell$-th Transolver layer defined in \eqref{eq:H-layer-l-update}, and the trainable parameters $\theta$ collect $E_{\mathrm{in}}$, $E_{\mathrm{out}}$, and the per-layer assignment projections $P_{s,a}^{\ell}$, token-logit projections $P_{\mathrm{slice},a}^{\ell}$, value projections $P_{v,a}^{\ell}$, learnable temperatures $\tau_a^{\ell}$, multi-head self-attention weights, head-mixing output projections, and MLPs.

We reiterate that $ \mathcal{F}_\theta$ acts on an arbitrary point set, so it can be evaluated on different meshes than it was trained on.

\begin{table}[t]
\centering
\caption{Transolver model inputs for the lattice benchmark. The model consumes two input tensors per mesh cell, a functional input $\texttt{fx} \in \mathbb{R}^{N \times 14}$ and an embedding $\texttt{emb} \in \mathbb{R}^{N \times 4}$. For the hohlraum benchmark, the source term $Q$ is omitted from the embedding ($Q = 0$ everywhere), giving $\texttt{emb} \in \mathbb{R}^{N \times 3}$.}
\label{tab:transolver_inputs}
\begin{tabular}{l l c l}
\toprule
\textbf{Tensor} & \textbf{Feature} & \textbf{Dim} & \textbf{Description} \\
\midrule
\multirow{3}{*}{\texttt{fx}}
 & $\bar{x}, \bar{y}$            & 2  & Normalized cell centroid coordinates \\
 & $\bff_i$ & 12 & Fourier features \eqref{eq_fourier_feature} ($\sin$ and $\cos$ of $K_{\mathrm{ff}}=3$ frequencies for each of the 2 coordinates) \\
\cmidrule{2-4}
 & \textbf{Total}    & \textbf{14} & \\
\midrule
\multirow{5}{*}{\texttt{emb}}
 & $\sigmaa$        & 1  & Absorption cross-section \\
 & $\sigmas$        & 1  & Scattering cross-section \\
 & $\sigmat$        & 1  & Total cross-section \\
 & $Q$               & 1  & Source term (lattice only) \\
\cmidrule{2-4}
 & \textbf{Total}    & \textbf{4} & (3 for hohlraum) \\
\midrule
\multicolumn{2}{l}{\textbf{Output}: $\hat{\Phi}$} & \textbf{1} & Predicted particle concentration \\
\bottomrule
\end{tabular}
\end{table}

\subsection{Implementation details of BSMS-MGN}\label{sec_bsms_mgn_definition}
BSMS-MGN \cite{cao2022bi} models the
 surrogate operator $\mathcal{F}_\theta$ from \eqref{eq_surrogate_model} as a hierarchy of graphs $\{\mathcal{G}^{(l)}\}_{l=0}^{L_g}$ built from the spatial discretization, where $l=0$ is the finest level and $l=L_g$ the coarsest in a multi-grid-like U-Net \cite{ronneberger2015unetconvolutionalnetworksbiomedical} approach in a latent space.

At the finest level, $\mathcal{G}^{(0)} = (\mathcal{V}^{(0)}, \mathcal{E}^{(0)})$ is the cell-dual graph of the finite-volume mesh. Each node $i \in \mathcal{V}^{(0)}$ corresponds to a mesh cell located at the cell centroid $\bx_i \in \mathbb{R}^2$, and a directed edge $(i,j) \in \mathcal{E}^{(0)}$ exists whenever cells $i$ and $j$ share a face. The graph is bidirectional by construction.

{On the finest level, each node $i\in\mathcal{V}^{(0)}$ carries the per-cell input of \eqref{eq_surrogate_model}, the same as for Transolver: the normalized centroid $\bar{\bx}_i$, the Fourier features $\bff_i$, and the material/source vector $\bp_i$, where $Q_i$ is omitted for Hohlraum ($Q\equiv 0$). Coarser levels carry latent states only; their node positions $\bx_{l,i}$ are derived by \eqref{eq_bsms_down}. Without Fourier features, the $\sin$ and $\cos$ channels of \eqref{eq_fourier_feature} are dropped, giving $d_v=6$ for Lattice and $d_v=5$ for Hohlraum.}
For each edge $(i,j) \in \mathcal{E}^{(0)}$ of the finest level, the edge feature contains the relative geometric offset and its norm in physical units,
\begin{align}\label{eq_bsms_edge_feature}
\bfe_{ij}^{0} = [\,\Delta x_{ij},\, \Delta y_{ij},\, \|\Delta\bx_{ij}\|\,] \in \mathbb{R}^{3}, \quad \Delta\bx_{ij} = \bx_i - \bx_j,
\end{align}

The graph hierarchy $\{\mathcal{G}^{(l)}\}_{l=1}^{L_g}$ is constructed as a one-time preprocessing step per simulation and cached with the dataset. Each level is obtained by the bi-stride pooling algorithm of \cite{cao2022bi}. For each connected component of $\mathcal{G}^{(l)}$, the \emph{CloseCenter} heuristic \cite{cao2022bi} selects the cell closest to the component centroid as the BFS seed, and every other BFS frontier is retained to form the coarse node set $\mathcal{I}^{(l)} \subset \mathcal{V}^{(l)}$.

The Encoder consists of two MLPs that map the node and edge inputs to a shared latent space of dimension $C_g$,
\begin{align}\label{eq_graph_encoder_equations}
\ssbold{n}_i^{0} = \mathrm{Enc}_v(\bar{\bx}_i,\bp_i,\bff_i) \in \mathbb{R}^{C_g}, \qquad \ssbold{g}_{ij}^{0} = \mathrm{Enc}_e(\bfe_{ij}^{0}) \in \mathbb{R}^{C_g},
\end{align}
each a two-hidden-layer ReLU MLP with LayerNorm on the output.

\paragraph{Message passing.}
The encoded latent graph first passes through $K_{\mathrm{mp}}$ message-passing blocks on the finest level $\mathcal{G}^{(0)}$. Each block $k $ is parameterized by two two-hidden-layer ReLU MLPs with LayerNorm on the output, $\mathrm{MLP}_e^{k}\colon \mathbb{R}^{3C_g} \to \mathbb{R}^{C_g}$ and $\mathrm{MLP}_v^{k}\colon \mathbb{R}^{2C_g} \to \mathbb{R}^{C_g}$, with disjoint weights across blocks. The block updates edges first, then nodes, with residual connections:
\begin{align}
\ssbold{g}_{ji}^{k+1}
&= \ssbold{g}_{ji}^{k} + \mathrm{MLP}_e^{k}\!\left([\ssbold{g}_{ji}^{k},\,
\ssbold{n}_{j}^{k},\, \ssbold{n}_{i}^{k}]\right),
& (j,i) &\in \mathcal{E}^{(0)},\\[2pt]
\ssbold{a}_{i}^{k+1}
&= \sum_{j:(j,i)\in\mathcal{E}^{(0)}} \ssbold{g}_{ji}^{k+1},
& i &\in \mathcal{V}^{(0)},\\[2pt]
\ssbold{n}_{i}^{k+1}
&= \ssbold{n}_{i}^{k} + \mathrm{MLP}_v^{k}\!\left([\ssbold{n}_{i}^{k},\, \ssbold{a}_{i}^{k+1}]\right),
& i &\in \mathcal{V}^{(0)},
\end{align}
where $\ssbold{a}_{i}^{k+1}$ is the aggregated incoming message at node $i$. After $K_{\mathrm{mp}}$ blocks the latent state $\ssbold{n}_i^{K_{\mathrm{mp}}}$ initializes the bi-stride U-Net pass.

\paragraph{Cross level information transfer.}
A single bi-stride U-Net pass exchanges information across the mesh levels. Let $\bv_i^{(l)}$ denote the latent state of node $i$ at level $l$ during the pass; on entry, $\bv_i^{(0)} = \ssbold{n}_i^{K_{\mathrm{mp}}}$, and the down and up blocks below overwrite $\bv^{(l)}$ in place. At each level, a per-level graph-message-passing (GMP) block $\mathrm{GMP}^{(l)}$ recomputes edge messages on the fly from positional offsets prepended to the sender/receiver latents \cite{cao2022bi}:
\begin{align}\label{eq_bsms_gmp}
\bfe_{l,ij} &= \mathrm{MLP}_e^{(l)}\!\left([\Delta\bx_{l,ij},\, \|\Delta\bx_{l,ij}\|,\, \bv_{l,i},\, \bv_{l,j}]\right), & (i,j) &\in \mathcal{E}^{(l)}, \\
\bv_{l,i}' &= \bv_{l,i} + \mathrm{MLP}_v^{(l)}\!\left([\bv_{l,i},\, \sum_{j:(j,i)\in\mathcal{E}^{(l)}} \bfe_{l,ji}]\right), & i &\in \mathcal{V}^{(l)},
\end{align}
where $\Delta\bx_{l,ij} = \bx_{l,i} - \bx_{l,j}$ are the cell-centroid offsets at level $l$, and $\mathrm{MLP}_e^{(l)}$, $\mathrm{MLP}_v^{(l)}$ are two-hidden-layer ReLU MLPs with disjoint weights across mesh levels and across the down, up, and bottom passes. Parenthesized superscripts $(l)$ index the mesh level, distinguishing these from the finest-level $\mathrm{MLP}_e^{k}, \mathrm{MLP}_v^{k}$ which are indexed by the block number $k$. Except for the finest-level processor, each level's GMP block contains no separately propagated edge latent, so edge messages are recomputed from $(\Delta\bx_{l,ij}, \bv_{l,i}, \bv_{l,j})$ at every call.

Information transfer between adjacent levels uses a non-parametric weighted graph convolution governed by a contribution table with entries $C_{ij}^{(l)}$. Initialize node weights $w_i^{(0)} = 1$. At each downsampling step $l \to l+1$ we have
\begin{align}\label{eq_bsms_contribution}
\widehat{w}_{ij}^{(l)} = \frac{w_i^{(l)}}{\deg^{(l)}(i)}, \qquad
C_{ij}^{(l)} = \frac{\widehat{w}_{ij}^{(l)}}{\sum_{i': (i',j)\in\mathcal{E}^{(l)}} \widehat{w}_{i'j}^{(l)} + \epsilon},
\end{align}
where $\deg^{(l)}(i)$ is the out-degree of node $i$ at level $l$. The latent features, positions, and node weights are downsampled by
\begin{align}\label{eq_bsms_down}
\bv_j^{(l+1)} &= \sum_{i:(i,j)\in\mathcal{E}^{(l)}} C_{ij}^{(l)}\, \bv_i^{(l)}, &
\bx_{l+1,j} &= \sum_{i:(i,j)\in\mathcal{E}^{(l)}} C_{ij}^{(l)}\, \bx_{l,i}, &
j &\in \mathcal{I}^{(l)},
\end{align}
restricted to the kept indices. The updated weights $w_j^{(l+1)} = \sum_{i} \widehat{w}_{ij}^{(l)}$ are propagated to the next level.

The upsampling step $l+1 \to l$ first scatters (unpools) the level-$(l+1)$ latent into a zero-initialized array of length $|\mathcal{V}^{(l)}|$ at the positions of $\mathcal{I}^{(l)}$, then returns information from receivers back to senders using the cached contribution table in transposed orientation:
\begin{align}\label{eq_bsms_return}
\bv_i^{(l)} \;\mathrel{+}=\; \sum_{j:(i,j)\in\mathcal{E}^{(l)}} C_{ij}^{(l)}\, \bv_j^{(l+1)} \quad (j \in \mathcal{I}^{(l)}).
\end{align}
A per-level GMP block $\mathrm{GMP}_{\mathrm{up}}^{(l)}$ then refines the unpooled latent, and a residual connection adds the corresponding pre-downsampling  state saved during the down pass. The contribution tables carry no trainable parameters.

\paragraph{U-Net pass.}
The overall U-Net pass is
\begin{align}
\text{down:}\quad & \bv^{(l)} \xrightarrow{\mathrm{GMP}_{\mathrm{down}}^{(l)}} \bv^{(l)} \xrightarrow{\text{contribution }C^{(l)}} \bv^{(l+1)}, && l=0,\ldots,L_g-1, \\
\text{bottom:}\quad & \bv^{(L_g)} \xrightarrow{\mathrm{GMP}_{\mathrm{bottom}}^{(L_g)}} \bv^{(L_g)}, \\
\text{up:}\quad & \bv^{(l+1)} \xrightarrow{\text{transpose }C^{(l)}} \bv^{(l)} \xrightarrow{\mathrm{GMP}_{\mathrm{up}}^{(l)}} \bv^{(l)} + \bv_{\text{down}}^{(l)}, && l=L_g-1,\ldots,0.
\end{align}

The decoder consists of a two-hidden-layer ReLU MLP without LayerNorm, $\mathrm{Dec}\colon \mathbb{R}^{C_g} \to \mathbb{R}$, mapping the finest-level output of the U-Net pass to the predicted particle concentration at the final time,
\begin{align}
\hat{\Phi}_i = \mathrm{Dec}\Big(\big[\mathcal{U}_{L_g}(\ssboldnKmp)\big]_i\Big), \qquad i = 1,\ldots,N_h,
\end{align}

Overall, the BSMS-MGN models the operator in \eqref{eq_surrogate_model} as
\begin{align}
\mathcal{F}_\theta = \mathrm{Dec} \circ \mathcal{U}_{L_g} \circ \mathcal{P}_{K_{\mathrm{mp}}-1} \circ \cdots \circ \mathcal{P}_{0} \circ \mathrm{Enc},
\end{align}
where $\mathrm{Enc} = (\mathrm{Enc}_v, \mathrm{Enc}_e)$, $\mathcal{P}_k$ is the $k$-th finest-level message-passing block, and $\mathcal{U}_{L_g}$ denotes the bi-stride U-Net pass with $L_g$ coarsening levels. The trainable parameters are
\[
\theta =
\bigl\{
\mathrm{Enc},\,
\{\mathcal{P}_k\}_{k=0}^{K_{\mathrm{mp}}-1},\,
\{\mathrm{GMP}_{\mathrm{down}}^{(l)}\}_{l=0}^{L_g-1},\,
\mathrm{GMP}_{\mathrm{bottom}}^{(L_g)},\,
\{\mathrm{GMP}_{\mathrm{up}}^{(l)}\}_{l=0}^{L_g-1},\,
\mathrm{Dec}
\bigr\}.
\]
Each $\mathrm{GMP}$ block has its own edge and node MLP parameters; the down, bottom, and up blocks do not share weights. As for Transolver, $\mathcal{F}_\theta$ is defined for any mesh: the graph, its hierarchy, and the node positions are inputs, not part of the model, so the same weights can be evaluated on the coarser mesh of \Cref{app:low_res}.

\paragraph{Receptive field.}
The $K_{\mathrm{mp}}$ finest-level message-passing blocks reach a $K_{\mathrm{mp}}$-hop graph neighborhood, corresponding to a spatial reach of approximately $K_{\mathrm{mp}}\, r_0$ on the cell-dual graph, where $r_0$ is the mean cell-to-cell distance. The bi-stride U-Net pass extends this reach additively. One hop on level $l$ spans $2^{l}$ hops of the finest graph, and each level contributes one GMP hop plus one pooling or unpooling hop on the way down and on the way up, so the pass reaches about $2\sum_{l=0}^{L_g-1}2\cdot 2^{l}+2^{L_g}\approx 5\cdot 2^{L_g}$ finest-level hops. With $K_{\mathrm{mp}}=16$, $L_g=5$, and a mean cell-to-cell distance $r_0\approx 0.019$ on the Lattice cell-dual graph, the receptive field of one forward pass is about $(K_{\mathrm{mp}}+5\cdot 2^{L_g})\,r_0\approx 170\,r_0\approx 3.3$ length units, comparable to the distance from the central source to the domain boundary ($3.0$ to $3.5$) but short of the domain diagonal ($7\sqrt{2}\approx 9.9$). On the Hohlraum mesh ($r_0\approx 0.004$) it is about $0.7$, the distance from either inflow boundary to the capsule. Since the Euclidean distance between two cells is at most the length of the graph path connecting them, these values are upper bounds on the Euclidean reach.

\begin{table}[t]
\centering
\caption{BSMS-MGN finest-level model inputs. Node features collect normalized coordinates, Fourier features, and per-cell material properties: $\widetilde{\ssbold{n}}_i \in \mathbb{R}^{18}$ for the Lattice benchmark (including the source term $Q$), and $\widetilde{\ssbold{n}}_i \in \mathbb{R}^{17}$ for the Hohlraum benchmark (where $Q\equiv 0$ everywhere and is dropped from the input). Edge features $\bfe_{ij}^{0} \in \mathbb{R}^{3}$ are the relative geometric offset and its norm in physical units, z-score normalized using statistics computed over the training split. Edges are the cell-dual adjacency from the finite-volume mesh: $(i,j)$ exists iff cells $i$ and $j$ share a face.}
\label{tab:mgn_inputs}
\begin{tabular}{l l c l}
\toprule
\textbf{Tensor} & \textbf{Feature} & \textbf{Dim} & \textbf{Description} \\
\midrule
\multirow{4}{*}{Node $\widetilde{\ssbold{n}}_i$}
 & $x, y$                 & 2  & Normalized cell centroid coordinates \\
 & $\bff_i$ & 12 & Fourier features \eqref{eq_fourier_feature} ($\sin$ and $\cos$ of $K_{\mathrm{ff}}=3$ frequencies for each of the 2 coordinates) \\
 & $\sigmaa, \sigmas, \sigmat$ & 3 & Material cross-sections \\
 & $Q$                    & 1  & Source term (Lattice only; dropped for Hohlraum) \\
\cmidrule{2-4}
 & \textbf{Total}         & \textbf{18} & Lattice (17 for Hohlraum: $Q$ dropped) \\
\midrule
\multirow{2}{*}{Edge $\bfe_{ij}^{0}$}
 & $\Delta x_{ij},\; \Delta y_{ij}$ & 2 & Cell-centroid positional offset (z-score normalized) \\
 & $\|\Delta\bx_{ij}\|$         & 1 & Offset norm in physical units (z-score normalized) \\
\cmidrule{2-4}
 & \textbf{Total}         & \textbf{3} & Per-endpoint cross-sections enter via node features \\
\midrule
\multicolumn{2}{l}{\textbf{Output}: $\hat{\Phi}$} & \textbf{1} & Predicted particle concentration \\
\bottomrule
\end{tabular}
\end{table}

\begin{table}[ht]
  \centering
  \caption{Transolver hyperparameters for each case. The QoI loss weight is $\lambda=5\times10^{-3}$ for Lattice and $\lambda=10^{-2}$ for Hohlraum (matching \Cref{sec_transolver_training}). The void region weight of $10$ applies only to the region-weighted (+RW, +FF+RW) configurations; the Lattice model is trained in the Base configuration, so its void weight is $1$.}
  \label{tab:transolver-hparams}
  \begin{tabular}{lcc}
    \toprule
    Hyperparameter & Lattice & Hohlraum \\
    \midrule
    \multicolumn{3}{l}{\textbf{Architecture}} \\
    Layers                   & 8   & 8   \\
    Hidden dim               & 256 & 256 \\
    Attention heads           & 16  & 16  \\
    Functional dim           & 14  & 14  \\
    Embedding dim            & 4   & 3   \\
    Physics tokens           & 128 & 128 \\
    MLP ratio                & 4   & 4   \\
    Dropout                  & 0.0 & 0.0 \\
    Include $Q$ in embedding & Yes & No  \\
    Transformer Engine       & No  & No  \\
    \midrule
    \multicolumn{3}{l}{\textbf{Data \& features}} \\
    Fourier frequencies      & 3   & 3   \\
    Fourier base frequency   & 2.0 & 2.0 \\
    \midrule
    \multicolumn{3}{l}{\textbf{Optimization}} \\
    Epochs                   & 500 & 500 \\
    Effective batch size     & 8   & 8   \\
    Optimizer                & Muon & Muon \\
    Muon LR (matrix params)  & $3\!\times\!10^{-5}$ & $3\!\times\!10^{-5}$ \\
    Adam LR (1-D params)     & $3\!\times\!10^{-5}$ & $3\!\times\!10^{-5}$ \\
    Grad-norm clip           & $10$ & $10$ \\
    Min LR                   & $10^{-6}$ & $10^{-6}$ \\
    LR schedule              & Cosine & Cosine \\
    Warmup epochs            & 10  & 10  \\
    Weight decay             & 0.0 & 0.0 \\
    AMP dtype                & BF16 & BF16 \\
    QoI loss weight $\lambda$ & $5\!\times\!10^{-3}$ & $10^{-2}$ \\
    Void region weight ($+$RW) & $1$   & $10$  \\
    \bottomrule
  \end{tabular}
\end{table}

\begin{table}[ht]
  \centering
  \caption{BSMS-MGN hyperparameters for each case. Architecture and graph-construction values are those of the released training configuration (code repository linked in the introduction). The hidden dimension $C_g=156$ is chosen so that the trainable parameter count matches the Transolver baseline (\Cref{tab:training-cost}).  The void region weight $10$ applies to the +RW configurations only; the Lattice model is trained in the Base configuration, i.e., with weight $1$.}
  \label{tab:mgn-hparams}
  \begin{tabular}{lcc}
    \toprule
    Hyperparameter & Lattice & Hohlraum \\
    \midrule
    \multicolumn{3}{l}{\textbf{Architecture (finest-level processor)}} \\
    Message-passing blocks $K_{\mathrm{mp}}$ & 16  & 16  \\
    Hidden dim $C_g$           & 156 & 156 \\
    Node/edge MLP layers       & 2   & 2   \\
    Activation                 & ReLU & ReLU \\
    Aggregation                & Sum & Sum  \\
    Normalization              & LayerNorm & LayerNorm \\
    Concat trick               & Yes & Yes \\
    Input node features $d_v$  & 18  & 17  \\
    Input edge features $d_e$  & 3   & 3   \\
    \midrule
    \multicolumn{3}{l}{\textbf{Architecture (bi-stride U-Net)}} \\
    Coarsening levels $L_g$    & 5   & 5   \\
    Position dimension         & 2   & 2   \\
    GMP MLP layers per level   & 2   & 2   \\
    U-Net passes               & 1   & 1   \\
    Seeding heuristic          & CloseCenter & CloseCenter \\
    \midrule
    \multicolumn{3}{l}{\textbf{Graph construction}} \\
    Method                   & Cell-dual adjacency & Cell-dual adjacency \\
    Median directed node degree & $\approx 6$ & $\approx 6$ \\
    \multicolumn{3}{l}{\footnotesize Directed degree: an interior triangular cell has $3$ face neighbours, i.e.\ $6$ directed edges (both directions).} \\
    \midrule
    \multicolumn{3}{l}{\textbf{Data \& features}} \\
    Include $Q$ in nodes     & Yes & No  \\
    Include material props   & Yes & Yes \\
    Fourier frequencies      & 3   & 3   \\
    Fourier base frequency   & 2.0 & 2.0 \\
    \midrule
    \multicolumn{3}{l}{\textbf{Optimization}} \\
    Epochs                   & 500 & 500 \\
    Effective batch size     & 8   & 8   \\
    Gradient accumulation    & 1   & 1   \\
    Optimizer (matrix params)& Muon & Muon \\
    Optimizer (1-D params)   & Adam & Adam \\
    Shared learning rate     & $3\!\times\!10^{-5}$ & $3\!\times\!10^{-5}$ \\
    Min LR                   & $10^{-6}$ & $10^{-6}$ \\
    LR schedule              & Cosine & Cosine \\
    Warmup epochs            & 10  & 10  \\
    Grad-norm clip           & $10$ & $10$ \\
    Weight decay             & 0.0  & 0.0  \\
    AMP dtype                & BF16 & BF16 \\
    QoI loss weight $\lambda$& $5\!\times\!10^{-3}$ & $1\!\times\!10^{-2}$ \\
    QoI warmup epochs        & 0    & 0    \\
    Void region weight ($+$RW) & $1$   & $10$  \\
    \bottomrule
  \end{tabular}
\end{table}

\clearpage
\section{Hohlraum ablation visualizations}\label{app:ablation_visualizations}
This appendix collects the field-prediction panels for every (architecture, configuration) combination of the Hohlraum ablation reported in \Cref{tab:Hohlraum-eval}. Each subsection below contains one figure that stacks the four configurations of one architecture, Base, +RW, +FF, and +FF+RW, as four sub-panels in that order. The contrasting per-architecture sensitivities discussed in \Cref{sec_ablations} are visible directly in the absolute-error panels. For Transolver the spatial error structure shrinks substantially in the Fourier-equipped rows, while for BSMS-MGN the central-capsule absorption region is closer to the KiT-RT reference in the non-Fourier rows. Each panel shows the KiT-RT reference (left), the surrogate prediction (middle), and the absolute error (right). Per-figure seed selection follows the main-text convention (lowest field RMSE $\Phi$ across three seeds per cell). The sample design parameters are $\mathrm{ulr}=0.3$, $\mathrm{llr}=-0.4$, $\mathrm{urr}=0.5$, $\mathrm{lrr}=-0.4$, $\mathrm{hlr}=0.5$, $\mathrm{hrr}=0.6$, $c_x=0.1$, $c_y=0.0$ throughout.

\subsection{Transolver ablation grid}\label{sec_transolver_ablation_grid}
\begin{figure}[h!]
    \centering
    \begin{subfigure}{1\linewidth}
        \centering
        \includegraphics[width=1\linewidth]{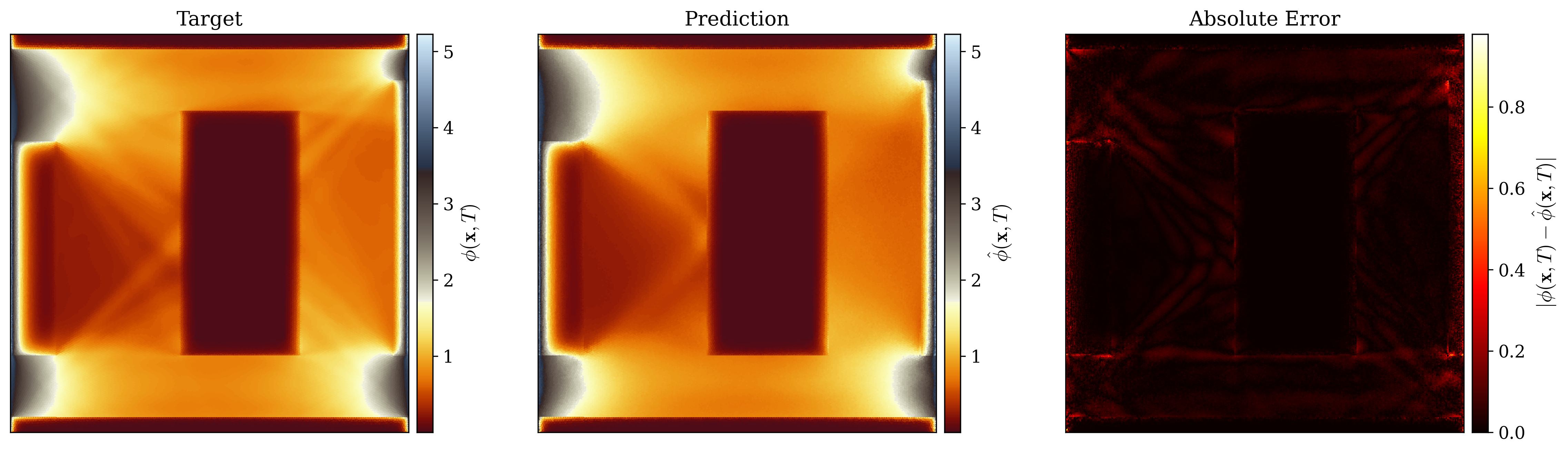}
        \caption{Transolver, Base.}
        \label{fig_transolver_hohlraum_baseline}
    \end{subfigure}

    \vspace{0.5em}

    \begin{subfigure}{1\linewidth}
        \centering
        \includegraphics[width=1\linewidth]{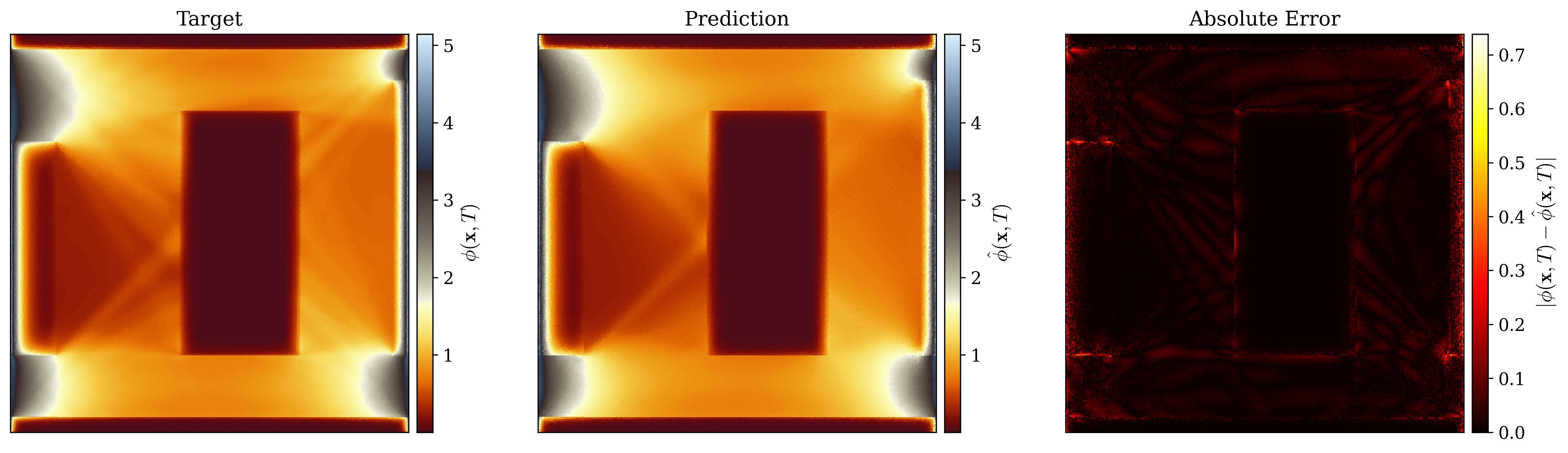}
        \caption{Transolver, +RW.}
        \label{fig_transolver_hohlraum_weights}
    \end{subfigure}

    \vspace{0.5em}

    \begin{subfigure}{1\linewidth}
        \centering
        \includegraphics[width=1\linewidth]{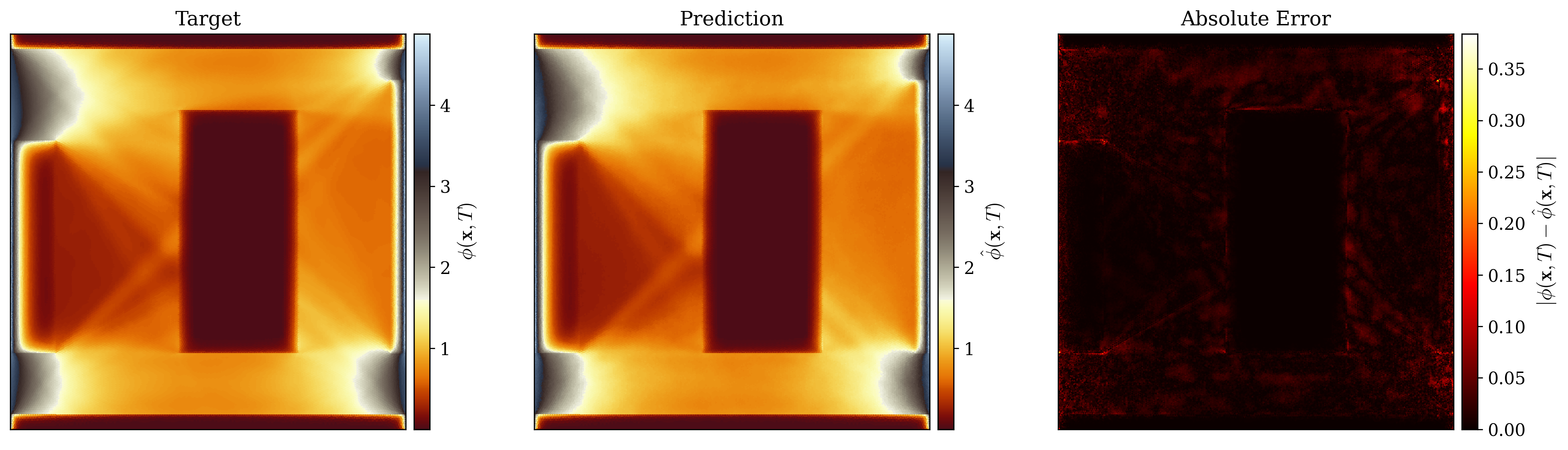}
        \caption{Transolver, +FF.}
        \label{fig_transolver_hohlraum_ff}
    \end{subfigure}

    \vspace{0.5em}

    \begin{subfigure}{1\linewidth}
        \centering
        \includegraphics[width=1\linewidth]{assets/transolver_hohlraum_high_filled.png}
        \caption{Transolver, +FF+RW. (Same panel as \Cref{fig:transolver_hohlraum}.)}
        \label{fig_transolver_hohlraum_full}
    \end{subfigure}

    \caption{Transolver Hohlraum ablation grid (high-resolution). Each row is one ablation configuration, and the three columns within each panel are the KiT-RT reference (left), surrogate prediction (middle), and absolute error (right). Quantitative comparison across configurations is in \Cref{tab:Hohlraum-eval}.}
    \label{fig_transolver_hohlraum_ablation_grid}
\end{figure}

\Cref{fig_transolver_hohlraum_ablation_grid} stacks the four Transolver cells. Moving from Base (top) to +FF+RW (bottom), the visible error magnitude near material interfaces and in the central capsule region drops noticeably once Fourier features are introduced (third and fourth rows), and is essentially unchanged by region weighting alone (second row).

\subsection{BSMS-MGN ablation grid}\label{sec_bsmsmgn_ablation_grid}
\begin{figure}[h!]
    \centering
    \begin{subfigure}{1\linewidth}
        \centering
        \includegraphics[width=1\linewidth]{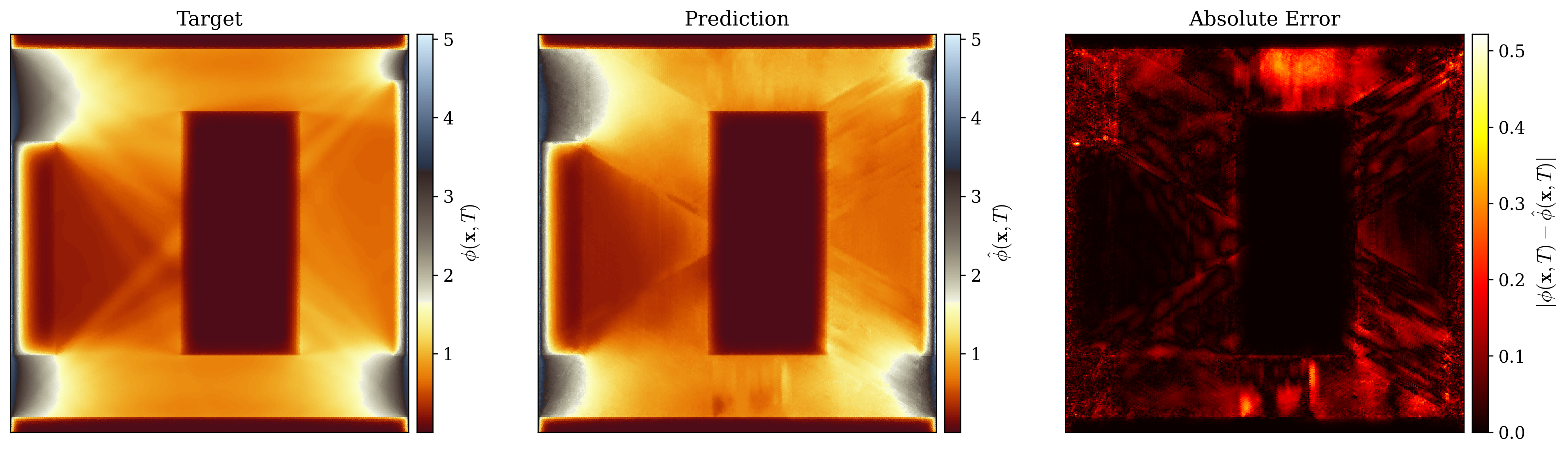}
        \caption{BSMS-MGN, Base.}
        \label{fig_bsmsmgn_hohlraum_baseline}
    \end{subfigure}

    \vspace{0.5em}

    \begin{subfigure}{1\linewidth}
        \centering
        \includegraphics[width=1\linewidth]{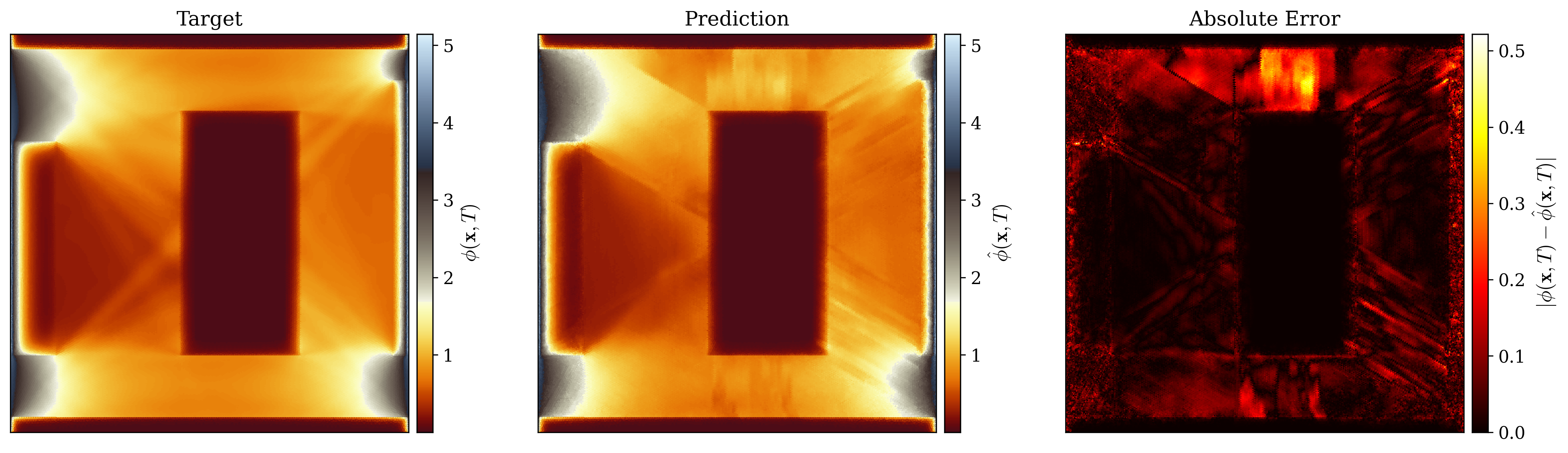}
        \caption{BSMS-MGN, +RW.}
        \label{fig_bsmsmgn_hohlraum_weights}
    \end{subfigure}

    \vspace{0.5em}

    \begin{subfigure}{1\linewidth}
        \centering
        \includegraphics[width=1\linewidth]{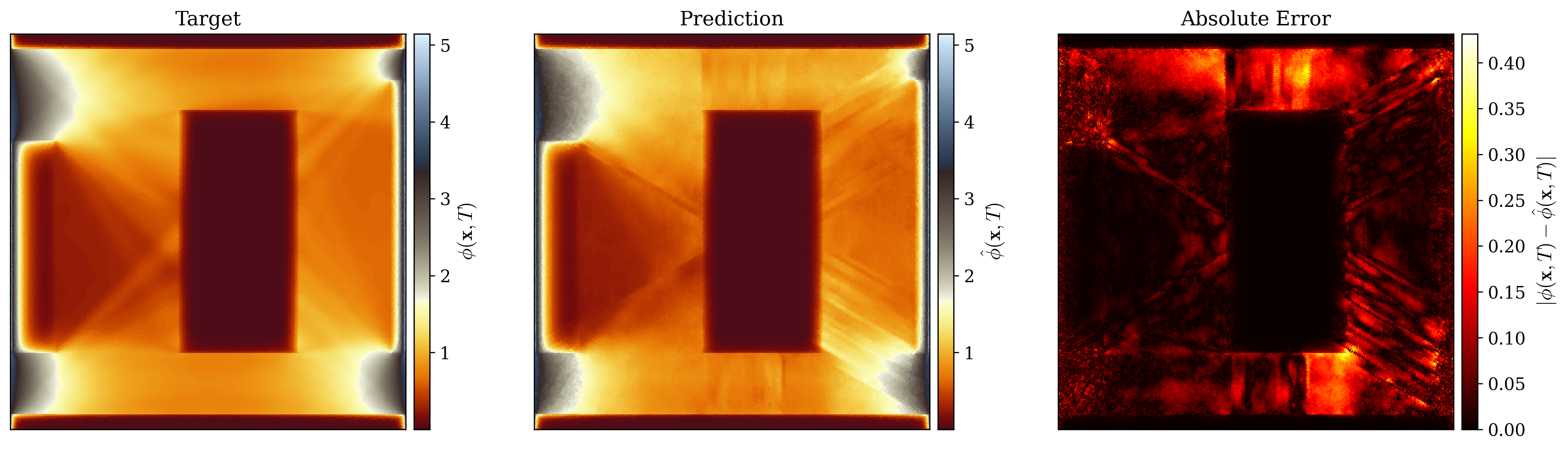}
        \caption{BSMS-MGN, +FF.}
        \label{fig_bsmsmgn_hohlraum_ff}
    \end{subfigure}

    \vspace{0.5em}

    \begin{subfigure}{1\linewidth}
        \centering
        \includegraphics[width=1\linewidth]{assets/bsmsmgn_hohlraum_high_filled.png}
        \caption{BSMS-MGN, +FF+RW. (Same panel as \Cref{fig:mgn_hohlraum}.)}
        \label{fig_bsmsmgn_hohlraum_full}
    \end{subfigure}

    \caption{BSMS-MGN Hohlraum ablation grid (high-resolution). Each row is one ablation configuration, and the three columns within each panel are the KiT-RT reference (left), surrogate prediction (middle), and absolute error (right). Quantitative comparison across configurations is in \Cref{tab:Hohlraum-eval}.}
    \label{fig_bsmsmgn_hohlraum_ablation_grid}
\end{figure}

\Cref{fig_bsmsmgn_hohlraum_ablation_grid} stacks the four BSMS-MGN cells in the same order. Moving from Base to +FF+RW, the absolute-error panels are visibly similar between the Base and +RW rows (first two), and the central-capsule absorption region degrades when Fourier features are added (third and fourth rows).

\section{Cross-mesh (low-resolution) surrogate evaluation}\label{app:low_res}
The main text evaluates the surrogates on the high-resolution mesh they were trained on (\Cref{sec_high_res}), and reports the low-resolution KiT-RT solver only as a cheap conventional-solver baseline. This appendix reports the complementary experiment: applying the same trained checkpoints, without retraining, to the coarser low-resolution mesh (cross-mesh transfer). This is possible because both surrogates are maps on point sets and graphs, \eqref{eq_surrogate_model}: the cell centroids, material data, Fourier features, and, for BSMS-MGN, the graph hierarchy are inputs, so the trained weights can be evaluated at the centroids of any mesh. The full high- and low-resolution results are given in \Cref{tab:Lattice-eval-full} (Lattice) and \Cref{tab:Hohlraum-eval-full} (Hohlraum); the high-resolution surrogate rows of these tables are the ones reproduced in the main-text \Cref{tab:Lattice-eval,tab:Hohlraum-eval}.

\begin{table}[t]
  \centering
  \caption{Lattice benchmark with high- and low-resolution evaluation. Surrogate models use the baseline configuration trained at high-resolution, with no Fourier features and no region-weighted loss, and are evaluated on either the high-resolution or low-resolution mesh. Columns report timings, speedup over the high-res KiT-RT solver, prediction accuracy of the final-time particle concentration (field RMSE, \Cref{eqn:rmse}), and relative error in the absorption QoI ($E_{\textup{QoI},\textup{rel}}$, \Cref{eqn:relative_error_pct}) evaluated at $t = T$. Surrogate rows report mean $\pm$ 1 standard deviation across $n=3$ random seeds, and KiT-RT rows are deterministic references. The best mean per column across the two surrogate architectures within each evaluation resolution is shown in bold. }
  \label{tab:Lattice-eval-full}
  \setlength{\tabcolsep}{4.5pt}
  \begin{tabular}{ll|rrrr}
    \toprule
   Resolution & Method & \makecell{Speedup vs\\high-res \\ KiT-RT}  & Time (s) & RMSE $\Phi$ & $E_{\textup{QoI},\textup{rel}}$ (\%) \\
    \midrule\midrule
    High & KiT-RT &$1.00\times$& $13.62\pm0.455$   & $-$ & $0.00$ \\
    \midrule
    High & Transolver & $197\times$&  $0.069\pm0.0007$  & $\bm{0.0828{\pm}5.20{\times}10^{-3}}$ & $1.06{\pm}0.0620$ \\
    High & BSMS-MGN & $90.4{\times}$ & $0.1506\pm0.0045$  & $0.0835{\pm}2.90{\times}10^{-5}$ & $\bm{0.276{\pm}0.210}$ \\
    \midrule
    Low & KiT-RT & $59.2\times$&  $0.23\pm0.007$ & $-$ & $6.60$ \\
    \midrule
    Low & Transolver & $852{\times}$ & $0.0160\pm0.0012$ & $\bm{0.0885{\pm}3.70{\times}10^{-3}}$ & $\bm{28.6{\pm}0.500}$ \\
    Low & BSMS-MGN & $376{\times}$ & $0.0363\pm0.0029$ &  $0.482{\pm}0.170$ & $193{\pm}11.0$ \\
    \bottomrule
  \end{tabular}
\end{table}

\begin{table}[t]
  \centering
  \caption{Hohlraum benchmark with high- and low-resolution evaluation, full ablation per architecture (Base, +RW, +FF, +FF+RW; defined in \Cref{tab:ablation-legend}).  Configurations are trained at high-resolution and evaluated on either the high-resolution or low-resolution mesh. Columns report timings, speedup over the high-res KiT-RT solver, prediction accuracy of the final-time particle concentration (field RMSE, \Cref{eqn:rmse}), and relative errors in the three per-region absorption QoIs relative to the high-resolution KiT-RT reference (\Cref{eqn:relative_error_pct}): center $S = G \cup B$, vertical-strip shield $S = R$, and horizontal-strip shield $S = K$. Surrogate rows report mean and standard deviation across $n=3$ random seeds per configuration. KiT-RT rows are deterministic references. The best mean per column within each architecture sub-block and evaluation resolution is shown in bold.
  }
  \label{tab:Hohlraum-eval-full}
  \setlength{\tabcolsep}{4pt}
  \resizebox{\linewidth}{!}{%
  \begin{tabular}{llrrrrrr}
    \toprule
   Resolution & Method & \makecell{Speedup vs\\high-res \\ KiT-RT} & Time (s) & RMSE $\Phi$ & $E_{\textup{QoI},\textup{rel}}$  Center (\%) & $E_{\textup{QoI},\textup{rel}}$  Vert. (\%) & $E_{\textup{QoI},\textup{rel}}$  Horiz. (\%) \\
 \midrule\midrule
    High & KiT-RT &$1{\times}$ & $35.95 \pm 0.495$ & $-$ & $0.00$ & $0.00$ & $0.00$ \\
    \midrule
    High & Transolver Base & $637{\times}$  & $0.05645 \pm 0.0067$ &  $0.0559{\pm}1.20{\times}10^{-3}$ & $0.860{\pm}0.0660$ & $0.551{\pm}0.0420$ & $1.17{\pm}0.190$ \\
    High & Transolver +RW & $618{\times}$  & $0.05815 \pm 0.0055$ & $0.0517{\pm}2.20{\times}10^{-4}$ & $0.959{\pm}0.0570$ & $0.684{\pm}0.0720$ & $1.37{\pm}0.130$ \\
    High & Transolver +FF & $639{\times}$  & $0.05625 \pm 0.0085$ & $\bm{0.0244{\pm}2.80{\times}10^{-4}}$ & $\bm{0.624{\pm}0.0380}$ & $\bm{0.280{\pm}0.0130}$ & $\bm{0.422{\pm}5.60{\times}10^{-3}}$ \\
    High & Transolver +FF+RW & $637{\times}$  & $0.05645 \pm 0.0058$& $0.0252{\pm}3.70{\times}10^{-4}$ & $0.649{\pm}0.0660$ & $0.335{\pm}9.10{\times}10^{-3}$ & $0.672{\pm}0.0450$ \\
    \midrule
    High & BSMS-MGN Base & $264{\times}$  & $0.13605 \pm 0.0936$ & $0.107{\pm}1.90{\times}10^{-3}$ & $10.6{\pm}0.120$ & $0.602{\pm}0.0960$ & $\bm{4.66{\pm}0.0520}$ \\
    High & BSMS-MGN +RW & $296{\times}$  & $0.12135 \pm 0.0623$ & $\bm{0.106{\pm}9.20{\times}10^{-4}}$ & $\bm{10.5{\pm}0.0960}$ & $\bm{0.508{\pm}0.0640}$ & $4.83{\pm}0.180$ \\
    High & BSMS-MGN +FF & $260{\times}$  & $0.13855 \pm 0.1033$ & $0.115{\pm}1.50{\times}10^{-3}$ & $15.6{\pm}0.850$ & $0.549{\pm}0.190$ & $4.81{\pm}0.0550$ \\
    High & BSMS-MGN +FF+RW &  $270{\times}$  & $0.13295 \pm 0.0959$ & $0.111{\pm}1.80{\times}10^{-3}$ & $13.8{\pm}0.710$ & $0.585{\pm}0.0990$ & $5.12{\pm}0.150$ \\
    \midrule
    Low & KiT-RT & $18.5{\times}$ & $1.94 \pm 0.043$ & $-$ & $10.6$ & $8.05$ & $4.00$ \\
    \midrule
    Low & Transolver Base & $4411{\times}$ & $0.00815 \pm 0.0009$ & $0.278{\pm}0.0480$ & $\bm{8.58{\pm}0.750}$ & $19.9{\pm}3.90$ & $25.4{\pm}2.90$ \\
    Low & Transolver +RW & $4411{\times}$ & $0.00815 \pm 0.0010$& $0.272{\pm}0.0290$ & $8.60{\pm}0.810$ & $19.1{\pm}2.30$ & $24.8{\pm}3.30$ \\
    Low & Transolver +FF & $4438{\times}$ & $0.00810 \pm 0.0011$ & $\bm{0.172{\pm}3.30{\times}10^{-3}}$ & $16.2{\pm}0.790$ & $7.46{\pm}0.0740$ & $\bm{15.0{\pm}1.20}$ \\
    Low & Transolver +FF+RW & $4438{\times}$ & $0.00810 \pm 0.0008$ & $0.193{\pm}5.30{\times}10^{-3}$ & $19.2{\pm}0.380$ & $\bm{7.05{\pm}0.110}$ & $18.1{\pm}0.660$ \\
    \midrule
    Low & BSMS-MGN Base & $704{\times}$  & $0.05105 \pm 0.0078$ & $0.385{\pm}0.0160$ & $41.3{\pm}2.80$ & $33.7{\pm}1.20$ & $66.2{\pm}2.30$ \\
    Low & BSMS-MGN +RW & $708{\times}$  & $0.05075 \pm 0.0067$ & $0.360{\pm}0.0200$ & $\bm{30.5{\pm}4.90}$ & $30.7{\pm}1.60$ & $65.4{\pm}12.0$ \\
    Low & BSMS-MGN +FF & $719{\times}$  & $0.05000 \pm 0.0067$ & $0.303{\pm}0.0130$ & $40.7{\pm}4.60$ & $\bm{6.23{\pm}0.340}$ & $\bm{14.3{\pm}3.50}$ \\
    Low & BSMS-MGN +FF+RW & $709{\times}$  & $0.05070 \pm 0.0067$ & $\bm{0.271{\pm}0.0400}$ & $32.0{\pm}10.0$ & $6.87{\pm}0.440$ & $18.9{\pm}8.50$ \\
    \bottomrule
  \end{tabular}
  }
\end{table}

\begin{figure}[t]
    \centering
    \begin{subfigure}{1\linewidth}
        \centering
        \includegraphics[width=1\linewidth]{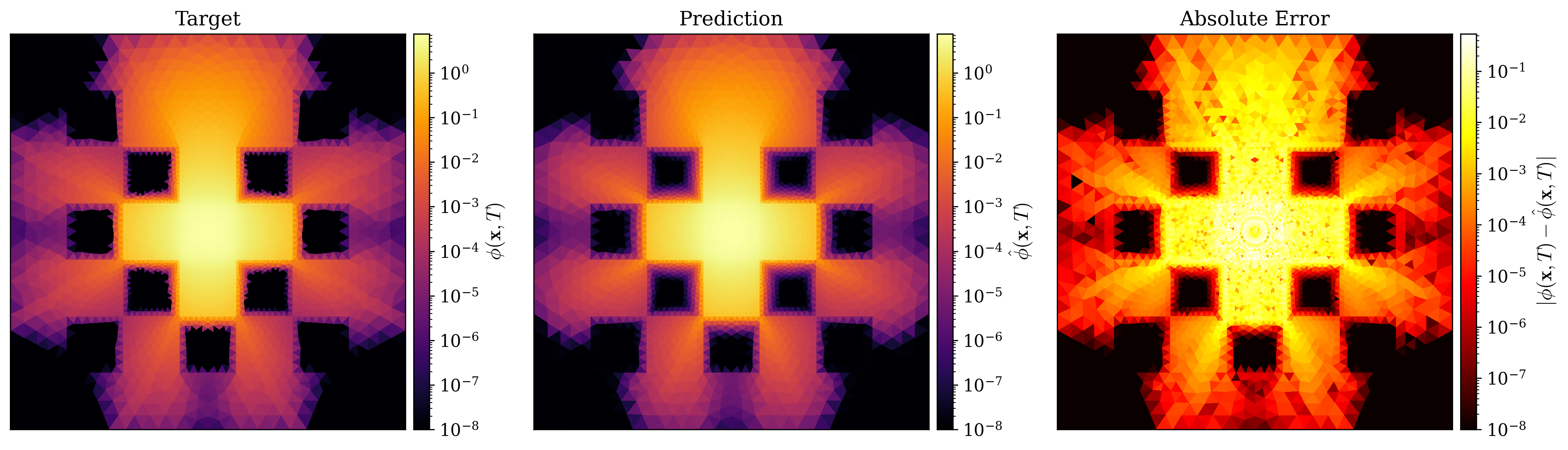}
        \caption{Transolver, bare baseline.}
        \label{fig_transolver_lattice_low}
    \end{subfigure}

    \vspace{0.75em}

    \begin{subfigure}{1\linewidth}
        \centering
        \includegraphics[width=1\linewidth]{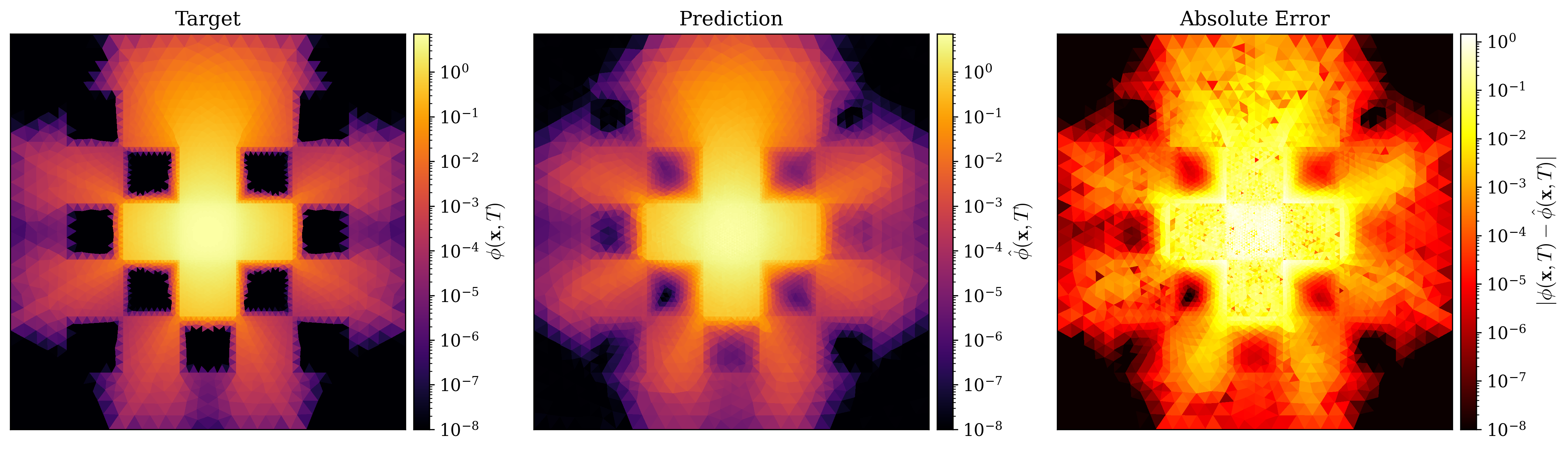}
        \caption{BSMS-MGN, bare baseline.}
        \label{fig_mgn_lattice_low}
    \end{subfigure}

    \caption{Lattice benchmark with low-resolution evaluation, baseline configuration (no Fourier features, no region-weighted loss). Per row: KiT-RT reference (left), surrogate prediction (middle), absolute error (right). Sample design parameters $\sigmaa^B=40.0$, $\sigmas^W=1.1$. The errors of both surrogates increase significantly, compared to the high-res reference. Transolver captures the high-absorbing regions, whereas BSMS-MGN induces too much diffusion to accurately model the benchmark.}
    \label{fig_lattice_low_comparison}
\end{figure}

\begin{figure}[t]
    \centering
    \begin{subfigure}{1\linewidth}
        \centering
        \includegraphics[width=1\linewidth]{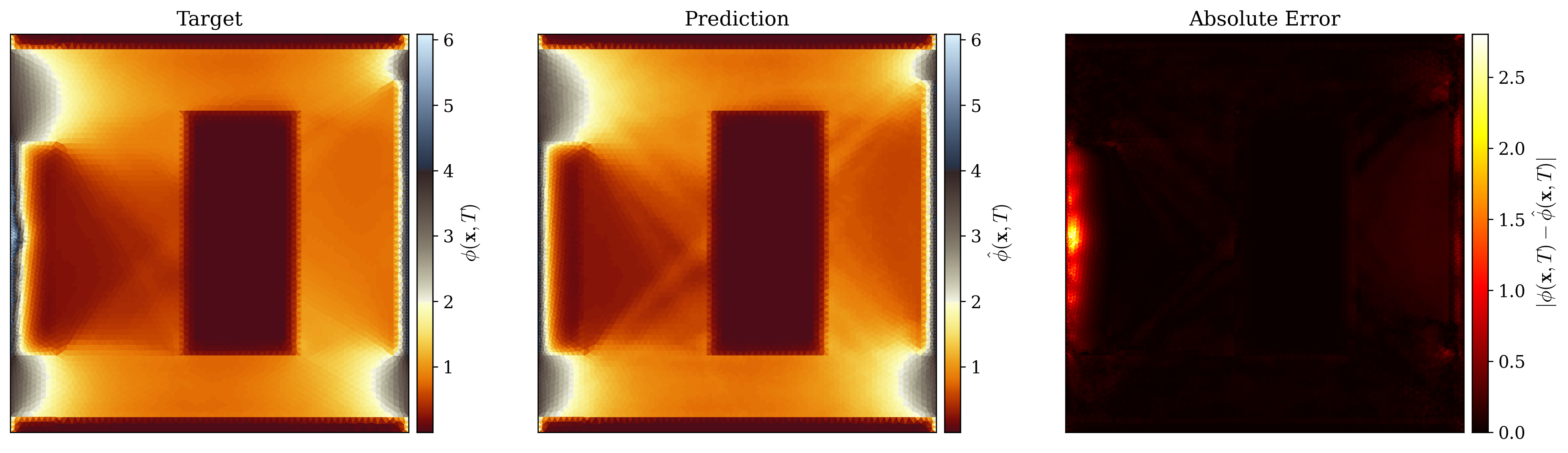}
        \caption{Transolver, full (+FF+RW) configuration.}
        \label{fig_transolver_hohlraum_low}
    \end{subfigure}

    \vspace{0.75em}

    \begin{subfigure}{1\linewidth}
        \centering
        \includegraphics[width=1\linewidth]{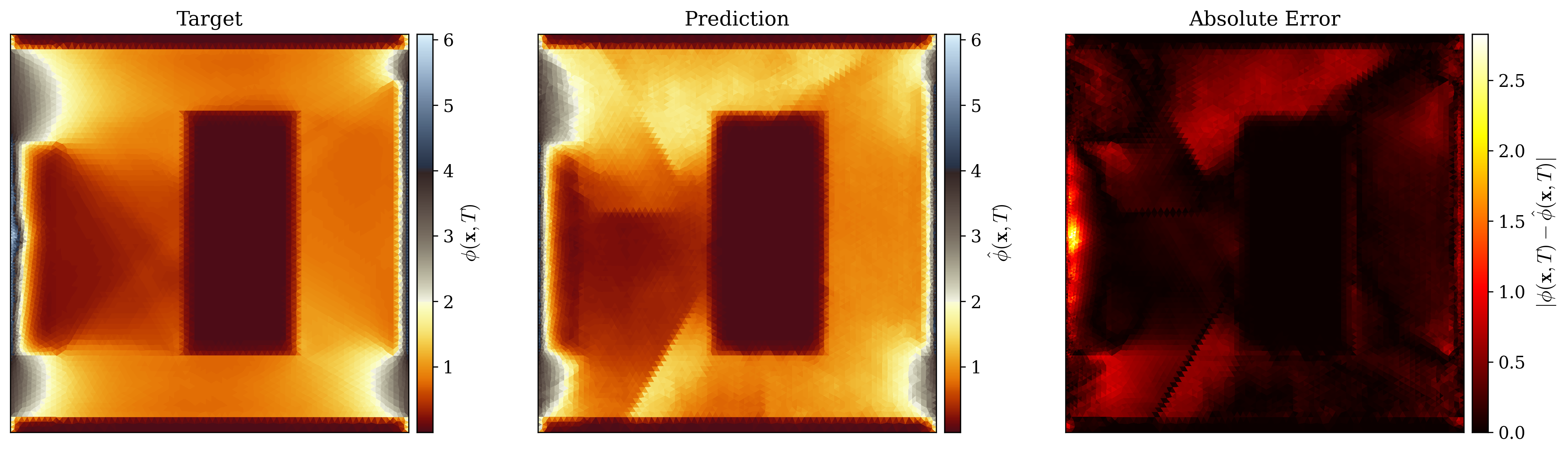}
        \caption{BSMS-MGN, full (+FF+RW) configuration.}
        \label{fig_mgn_hohlraum_low}
    \end{subfigure}

    \caption{Hohlraum benchmark, low-resolution evaluation, full (+FF+RW) configuration for both architectures. Per row: KiT-RT reference (left), surrogate prediction (middle), absolute error (right). Sample design parameters $\mathrm{ulr}=0.3$, $\mathrm{llr}=-0.4$, $\mathrm{urr}=0.5$, $\mathrm{lrr}=-0.4$, $\mathrm{hlr}=0.5$, $\mathrm{hrr}=0.6$, $c_x=0.1$, $c_y=0.0$. Both surrogates exhibit larger absolute errors compared to their high-res evaluations, especially near the inflow boundary. The Transolver captures the qualitative features of the benchmark, whereas BSMS-MGN exhibits nonphysical behavior.}
    \label{fig_hohlraum_low_comparison}
\end{figure}

\paragraph{Cross-mesh transfer degrades accuracy, asymmetrically.}
On the low-resolution mesh both surrogates degrade markedly relative to their high-resolution evaluation, and the degradation is asymmetric across architectures. Transolver largely retains its field RMSE but its absorption-QoI error rises well above the low-resolution KiT-RT baseline (\Cref{tab:Lattice-eval-full,tab:Hohlraum-eval-full}); on Hohlraum the +FF configuration remains best on field RMSE and the horizontal-shield QoI, though the center-capsule QoI worsens in every Fourier-equipped configuration, indicating that part of the Fourier advantage is mesh-specific. BSMS-MGN degrades much more: its field RMSE roughly triples to quadruples and most QoIs worsen by an order of magnitude or more (\Cref{fig_lattice_low_comparison,fig_hohlraum_low_comparison} show the corresponding nonphysical, over-diffused fields), consistent with the cell-dual graph hierarchy trained at high resolution not transferring cleanly to a coarser mesh. Cross-mesh surrogate evaluation is therefore competitive on raw inference time but not on accuracy; we recommend keeping the inference mesh resolution close to the training resolution, especially for BSMS-MGN.

%% file: refs.bib
@article{olson2000diffusion,
  title={Diffusion, P1, and other approximate forms of radiation transport},
  author={Olson, Gordon L and Auer, Lawrence H and Hall, Michael L},
  journal={Journal of Quantitative Spectroscopy and Radiative Transfer},
  volume={64},
  number={6},
  pages={619--634},
  year={2000},
  publisher={Elsevier}
}

@article{larsen1980diffusion,
  title={Diffusion theory as an asymptotic limit of transport theory for nearly critical systems with small mean free paths},
  author={Larsen, Edward W},
  journal={Annals of Nuclear Energy},
  volume={7},
  number={4-5},
  pages={249--255},
  year={1980},
  publisher={Pergamon}
}

@article{habetler1975uniform,
  title={Uniform asymptotic expansions in transport theory with small mean free paths, and the diffusion approximation},
  author={Habetler, GJ and Matkowsky, BJ},
  journal={Journal of Mathematical Physics},
  volume={16},
  number={4},
  pages={846--854},
  year={1975},
  publisher={American Institute of Physics}
}

@article{levermore1981flux,
  title={A flux-limited diffusion theory},
  author={Levermore, CD and Pomraning, GC},
  journal={Astrophysical Journal, Part 1, vol. 248, Aug. 15, 1981, p. 321-334.},
  volume={248},
  pages={321--334},
  year={1981}
}

@article{dubroca1999etude,
  title={Etude th{\'e}orique et num{\'e}rique d'une hi{\'e}rarchie de mod{\`e}les aux moments pour le transfert radiatif},
  author={Dubroca, Bruno and Feugeas, Jean-Luc},
  journal={Comptes Rendus de l'Acad{\'e}mie des Sciences-Series I-Mathematics},
  volume={329},
  number={10},
  pages={915--920},
  year={1999},
  publisher={Elsevier}
}

@article{brunner2005two,
  title={Two-dimensional time dependent Riemann solvers for neutron transport},
  author={Brunner, Thomas A and Holloway, James Paul},
  journal={Journal of Computational Physics},
  volume={210},
  number={1},
  pages={386--399},
  year={2005},
  publisher={Elsevier}
}

@article{frank2006partial,
  title={Partial moment entropy approximation to radiative heat transfer},
  author={Frank, Martin and Dubroca, Bruno and Klar, Axel},
  journal={Journal of Computational Physics},
  volume={218},
  number={1},
  pages={1--18},
  year={2006},
  publisher={Elsevier}
}

@article{hauck2011high,
  title={High-order entropy-based closures for linear transport in slab geometry},
  author={Hauck, Cory D},
  journal={Commun. Math. Sci},
  volume={9},
  number={1},
  pages={187--205},
  year={2011}
}

@article{yamamoto2005generalized,
  title={Generalized coarse-mesh rebalance method for acceleration of neutron transport calculations},
  author={Yamamoto, Akio},
  journal={Nuclear science and engineering},
  volume={151},
  number={3},
  pages={274--282},
  year={2005},
  publisher={Taylor \& Francis}
}

@article{mcclarren2010theoretical,
  title={Theoretical aspects of the simplified Pn equations},
  author={McClarren, Ryan G},
  journal={Transport Theory and Statistical Physics},
  volume={39},
  number={2-4},
  pages={73--109},
  year={2010},
  publisher={Taylor \& Francis}
}

@article{gelbard1960application,
  title={Application of spherical harmonics method to reactor problems},
  author={Gelbard, Ely M},
  journal={Bettis Atomic Power Laboratory, West Mifflin, PA, Technical Report No. WAPD-BT-20},
  year={1960}
}

@article{hamilton2015efficient,
  title={Efficient solution of the simplified PN equations},
  author={Hamilton, Steven P and Evans, Thomas M},
  journal={Journal of Computational Physics},
  volume={284},
  pages={155--170},
  year={2015},
  publisher={Elsevier}
}

@incollection{boulle2024mathematical,
  title={A mathematical guide to operator learning},
  author={Boull{\'e}, Nicolas and Townsend, Alex},
  booktitle={Handbook of Numerical Analysis},
  volume={25},
  pages={83--125},
  year={2024},
  publisher={Elsevier}
}

@article{kovachki2024operator,
  title={Operator learning: Algorithms and analysis},
  author={Kovachki, Nikola B and Lanthaler, Samuel and Stuart, Andrew M},
  journal={Handbook of Numerical Analysis},
  volume={25},
  pages={419--467},
  year={2024},
  publisher={Elsevier}
}

@article{kovachki2023neural,
  title={Neural operator: Learning maps between function spaces with applications to pdes},
  author={Kovachki, Nikola and Li, Zongyi and Liu, Burigede and Azizzadenesheli, Kamyar and Bhattacharya, Kaushik and Stuart, Andrew and Anandkumar, Anima},
  journal={Journal of Machine Learning Research},
  volume={24},
  number={89},
  pages={1--97},
  year={2023}
}

@article{subedi2025operator,
  title={Operator learning: A statistical perspective},
  author={Subedi, Unique and Tewari, Ambuj},
  journal={Annual Review of Statistics and Its Application},
  volume={13},
  year={2025},
  publisher={Annual Reviews}
}

@book{dupree2012monte,
  title={A Monte Carlo primer: A Practical approach to radiation transport},
  author={Dupree, Stephen A and Fraley, Stanley K},
  year={2012},
  publisher={Springer Science \& Business Media}
}

@article{gol1964quasi,
  title={A quasi-diffusion method of solving the kinetic equation},
  author={Gol'Din, V Ya},
  journal={USSR Computational Mathematics and Mathematical Physics},
  volume={4},
  number={6},
  pages={136--149},
  year={1964},
  publisher={Elsevier}
}

@article{warsa2004krylov,
  title={Krylov iterative methods and the degraded effectiveness of diffusion synthetic acceleration for multidimensional SN calculations in problems with material discontinuities},
  author={Warsa, James S and Wareing, Todd A and Morel, Jim E},
  journal={Nuclear science and engineering},
  volume={147},
  number={3},
  pages={218--248},
  year={2004},
  publisher={Taylor \& Francis}
}

@article{chacon2017multiscale,
  title={Multiscale high-order/low-order (HOLO) algorithms and applications},
  author={Chacon, Luis and Chen, Guangye and Knoll, Dana A and Newman, C and Park, H and Taitano, William and Willert, Jeff A and Womeldorff, Geoffrey},
  journal={Journal of Computational Physics},
  volume={330},
  pages={21--45},
  year={2017},
  publisher={Elsevier}
}

@article{adams2002fast,
  title={Fast iterative methods for discrete-ordinates particle transport calculations},
  author={Adams, Marvin L and Larsen, Edward W},
  journal={Progress in nuclear energy},
  volume={40},
  number={1},
  pages={3--159},
  year={2002},
  publisher={Elsevier}
}

@article{jin2010asymptotic,
  title={Asymptotic preserving (AP) schemes for multiscale kinetic and hyperbolic equations: a review},
  author={Jin, Shi},
  journal={Lecture notes for summer school on methods and models of kinetic theory (M\&MKT), Porto Ercole (Grosseto, Italy)},
  pages={177--216},
  year={2010},
  publisher={Citeseer}
}

@article{larsen1989asymptotic,
  title={Asymptotic solutions of numerical transport problems in optically thick, diffusive regimes II},
  author={Larsen, Edward W and Morel, Jim E},
  year={1989},
  publisher={Elsevier}
}

@article{larsen1987asymptotic,
  title={Asymptotic solutions of numerical transport problems in optically thick, diffusive regimes},
  author={Larsen, Edward W and Morel, Jim E and Miller Jr, Warren F},
  journal={Journal of Computational Physics},
  volume={69},
  number={2},
  pages={283--324},
  year={1987},
  publisher={Elsevier}
}

@article{larsen2023asymptotic,
  title={Asymptotic Diffusion Limits of the Multigroup Neutron Transport Equations},
  author={Larsen, Edward W},
  journal={Nuclear Science and Engineering},
  volume={197},
  number={2},
  pages={145--163},
  year={2023},
  publisher={Taylor \& Francis}
}

@article{bensoussan1979boundary,
  title={Boundary layers and homogenization of transport processes},
  author={Bensoussan, Alain and Lions, Jacques L and Papanicolaou, George C},
  journal={Publications of the Research Institute for Mathematical Sciences},
  volume={15},
  number={1},
  pages={53--157},
  year={1979},
  publisher={Research Institute forMathematical Sciences}
}

@article{mihalas1984radiation,
  title={Radiation hydrodynamics},
  author={Mihalas, Dimitri and Mihalas, BW},
  journal={Computational methods for astrophysical fluid flow. Saas-Fee advanced course},
  volume={27},
  pages={161--261},
  year={1984},
  publisher={Springer}
}

@book{lewis1984computational,
  title={Computational methods of neutron transport},
  author={Lewis, Elmer Eugene and Miller, Warren F},
  year={1984},
  publisher={John Wiley and Sons, Inc., New York, NY}
}

@book{suetens2017fundamentals,
  title={Fundamentals of medical imaging},
  author={Suetens, Paul},
  year={2017},
  publisher={Cambridge university press}
}

@article{baskar2012cancer,
  title={Cancer and radiation therapy: current advances and future directions},
  author={Baskar, Rajamanickam and Lee, Kuo Ann and Yeo, Richard and Yeoh, Kheng-Wei},
  journal={International journal of medical sciences},
  volume={9},
  number={3},
  pages={193},
  year={2012},
  publisher={Ivyspring International Publisher}
}

@book{mihalis2013foundations,
  title={Foundations of Radiation Hydrodynamics},
  author={Mihalis, B. and Mihalis, D.},
  year={2013},
  publisher={Dover Publications}
}

@book{davison1957neutron,
  title={Neutron transport theory},
  author={Davison, Boris},
  year={1957},
  publisher={Oxford: Clarendon Press}
}

@book{lewis2008fundamentals,
  title={Fundamentals of nuclear reactor physics},
  author={Lewis, Elmer E},
  year={2008},
  publisher={Elsevier}
}

@article{hurricane2023physics,
  title={Physics principles of inertial confinement fusion and US program overview},
  author={Hurricane, OA and Patel, PK and Betti, R and Froula, DH and Regan, SP and Slutz, SA and Gomez, MR and Sweeney, MA},
  journal={Reviews of Modern Physics},
  volume={95},
  number={2},
  pages={025005},
  year={2023},
  publisher={APS}
}

@article{carlson1968transport,
  title={Transport theory-the method of discrete ordinates},
  author={Carlson, Bengt G},
  journal={Computing methods in reactor physics},
  year={1968},
  publisher={Gordon and Breach Science Pub.}
}

@article{azmy2010advances,
  title={Advances in discrete-ordinates methodology},
  author={Azmy, Yousry and Sartori, Enrico and Larsen, Edward W and Morel, Jim E},
  journal={Nuclear computational science: A century in review},
  pages={1--84},
  year={2010},
  publisher={Springer}
}

@article{hauck2013collision,
author = {Hauck, Cory D. and McClarren, Ryan G.},
title = {A Collision-Based Hybrid Method for Time-Dependent, Linear, Kinetic Transport Equations},
journal = {Multiscale Modeling \& Simulation},
volume = {11},
number = {4},
pages = {1197-1227},
year = {2013},
doi = {10.1137/110846610},

URL = { 
        https://doi.org/10.1137/110846610
    
},
eprint = { 
        https://doi.org/10.1137/110846610
    
}
}

@article{CROCKATT2019455,
title = {Hybrid methods for radiation transport using diagonally implicit Runge–Kutta and space–time discontinuous Galerkin time integration},
journal = {Journal of Computational Physics},
volume = {376},
pages = {455-477},
year = {2019},
issn = {0021-9991},
doi = {https://doi.org/10.1016/j.jcp.2018.09.041},
url = {https://www.sciencedirect.com/science/article/pii/S0021999118306430},
author = {Michael M. Crockatt and Andrew J. Christlieb and C. Kristopher Garrett and Cory D. Hauck}
}

@inproceedings{JarrellAdams2011,
  author    = {Jarrell, Joshua J. and Adams, Marvin L.},
  title     = {Discrete-ordinates quadrature sets based on linear discontinuous finite elements},
  booktitle = {M\&C 2011: International Conference on Mathematics and Computational Methods Applied to Nuclear Science and Engineering},
  year      = {2011},
  address   = {Brazil},
}

@article{osti_1550352,
title = {An arbitrary-order, fully implicit, hybrid kinetic solver for linear radiative transport using integral deferred correction},
author = {Crockatt, Michael M. and Christlieb, Andrew J. and Garrett, C. Kristopher and Hauck, Cory D.},
abstractNote = {},
doi = {10.1016/j.jcp.2017.06.017},
url = {https://www.osti.gov/biblio/1550352}, journal = {Journal of Computational Physics},
issn = {0021-9991},
volume = 346,
place = {United States},
year = {2017},
month = {10}
}

@techreport{brunner2002forms,
  title={Forms of approximate radiation transport},
  author={Brunner, Thomas A},
  year={2002},
  institution={Sandia National Laboratories, Albuquerque, NM (United States)},
number = {SAND2002-1778}
}

@article{kusch2023kit,
  title={KiT-RT: An extendable framework for radiative transfer and therapy},
  author={Kusch, Jonas and Schotth{\"o}fer, Steffen and Stammer, Pia and Wolters, Jannick and Xiao, Tianbai},
  journal={ACM Transactions on Mathematical Software},
  volume={49},
  number={4},
  pages={1--24},
  year={2023},
  publisher={ACM New York, NY}
}

@article{CROCKATT2020109765,
title = {Improvements to a class of hybrid methods for radiation transport: Nyström reconstruction and defect correction methods},
journal = {Journal of Computational Physics},
volume = {422},
pages = {109765},
year = {2020},
issn = {0021-9991},
doi = {https://doi.org/10.1016/j.jcp.2020.109765},
url = {https://www.sciencedirect.com/science/article/pii/S0021999120305398},
author = {Michael M. Crockatt and Andrew J. Christlieb and Cory D. Hauck}
}

@article{reference_paper,
author = {Steffen Schotthöfer and C. D. Hauck},
title = {Reference Solutions for Linear Radiation Transport: the Hohlraum and Lattice Benchmarks},
journal = {Journal of Computational and Theoretical Transport},
volume = {54},
number = {6-7},
pages = {409--435},
year = {2025},
publisher = {Taylor \& Francis},
doi = {10.1080/23324309.2025.2573932},


URL = { 
    
        https://doi.org/10.1080/23324309.2025.2573932
    
    

},
eprint = { 
    
        https://doi.org/10.1080/23324309.2025.2573932
    
    

}

}

@software{PhysicsNeMo_Contributors_NVIDIA_PhysicsNeMo_An_2023,
author = {{PhysicsNeMo Contributors}},
month = feb,
title = {{NVIDIA PhysicsNeMo: An open-source framework for physics-based deep learning in science and engineering}},
url = {https://github.com/NVIDIA/physicsnemo},
year = {2023}
}

@article{wu2024transolver,
  title={Transolver: A fast transformer solver for pdes on general geometries},
  author={Wu, Haixu and Luo, Huakun and Wang, Haowen and Wang, Jianmin and Long, Mingsheng},
  journal={arXiv preprint arXiv:2402.02366},
  year={2024}
}

@inproceedings{pfaff2020learning,
  title={Learning mesh-based simulation with graph networks},
  author={Pfaff, Tobias and Fortunato, Meire and Sanchez-Gonzalez, Alvaro and Battaglia, Peter},
  booktitle={International conference on learning representations},
  year={2020}
}

@article{vaswani2017attention,
  title={Attention is all you need},
  author={Vaswani, Ashish and Shazeer, Noam and Parmar, Niki and Uszkoreit, Jakob and Jones, Llion and Gomez, Aidan N and Kaiser, {\L}ukasz and Polosukhin, Illia},
  journal={Advances in neural information processing systems},
  volume={30},
  year={2017}
}

@misc{ronneberger2015unetconvolutionalnetworksbiomedical,
      title={U-Net: Convolutional Networks for Biomedical Image Segmentation}, 
      author={Olaf Ronneberger and Philipp Fischer and Thomas Brox},
      year={2015},
      eprint={1505.04597},
      archivePrefix={arXiv},
      primaryClass={cs.CV},
      url={https://arxiv.org/abs/1505.04597}, 
}

@misc{ba2016layernormalization,
      title={Layer Normalization}, 
      author={Jimmy Lei Ba and Jamie Ryan Kiros and Geoffrey E. Hinton},
      year={2016},
      eprint={1607.06450},
      archivePrefix={arXiv},
      primaryClass={stat.ML},
      url={https://arxiv.org/abs/1607.06450}, 
}

@misc{jordan2024muon,
  author = {Jordan, Keller},
  title = {Muon: An optimizer for hidden layers in neural networks},
  year = {2024},
  note = {Blog post},
  url = {https://kellerjordan.github.io/posts/muon/}
}

@inproceedings{kingma2015adam,
  title     = {Adam: A Method for Stochastic Optimization},
  author    = {Kingma, Diederik P. and Ba, Jimmy},
  booktitle = {International Conference on Learning Representations},
  year      = {2015},
  url       = {https://arxiv.org/abs/1412.6980}
}

@article{bronstein2017geometric,
  title={Geometric deep learning: going beyond euclidean data},
  author={Bronstein, Michael M and Bruna, Joan and LeCun, Yann and Szlam, Arthur and Vandergheynst, Pierre},
  journal={IEEE Signal Processing Magazine},
  volume={34},
  number={4},
  pages={18--42},
  year={2017},
  publisher={IEEE}
}

@article{battaglia2018relational,
  title={Relational inductive biases, deep learning, and graph networks. arXiv 2018},
  author={Battaglia, Peter W and Hamrick, Jessica B and Bapst, Victor and Sanchez-Gonzalez, Alvaro and Zambaldi, Vinicius and Malinowski, Mateusz and Tacchetti, Andrea and Raposo, David and Santoro, Adam and Faulkner, Ryan and others},
  journal={arXiv preprint arXiv:1806.01261},
  volume={10},
  year={2018}
}

@inproceedings{sanchez2020learning,
  title={Learning to simulate complex physics with graph networks},
  author={Sanchez-Gonzalez, Alvaro and Godwin, Jonathan and Pfaff, Tobias and Ying, Rex and Leskovec, Jure and Battaglia, Peter},
  booktitle={International conference on machine learning},
  pages={8459--8468},
  year={2020},
  organization={PMLR}
}

@article{tancik2020fourier,
  title={Fourier features let networks learn high frequency functions in low dimensional domains},
  author={Tancik, Matthew and Srinivasan, Pratul and Mildenhall, Ben and Fridovich-Keil, Sara and Raghavan, Nithin and Singhal, Utkarsh and Ramamoorthi, Ravi and Barron, Jonathan and Ng, Ren},
  journal={Advances in neural information processing systems},
  volume={33},
  pages={7537--7547},
  year={2020}
}

@article{xu2018powerful,
  title={How powerful are graph neural networks?},
  author={Xu, Keyulu and Hu, Weihua and Leskovec, Jure and Jegelka, Stefanie},
  journal={arXiv preprint arXiv:1810.00826},
  year={2018}
}

@article{lam2023learning,
  title={Learning skillful medium-range global weather forecasting},
  author={Lam, Remi and Sanchez-Gonzalez, Alvaro and Willson, Matthew and Wirnsberger, Peter and Fortunato, Meire and Alet, Ferran and Ravuri, Suman and Ewalds, Timo and Eaton-Rosen, Zach and Hu, Weihua and others},
  journal={Science},
  volume={382},
  number={6677},
  pages={1416--1421},
  year={2023},
  publisher={American Association for the Advancement of Science}
}

@article{MCCLENNY2023111722,
title = {Self-adaptive physics-informed neural networks},
journal = {Journal of Computational Physics},
volume = {474},
pages = {111722},
year = {2023},
issn = {0021-9991},
doi = {https://doi.org/10.1016/j.jcp.2022.111722},
url = {https://www.sciencedirect.com/science/article/pii/S0021999122007859},
author = {Levi D. McClenny and Ulisses M. Braga-Neto}
}

@article{cao2022bi,
  title={Bi-stride multi-scale graph neural network for mesh-based physical simulation},
  author={Cao, Yadi and Chai, Menglei and Li, Minchen and Jiang, Chenfanfu},
  year={2022}
}
